\documentclass[svgnames,dvipsnames]{article}
\usepackage{notes}
\usepackage{subfiles}
\usepackage{wrapfig}
\usepackage[natbib=true]{biblatex}
\newcommand{\platent}{p_{\text{latent}}}

\newcommand{\prior}{p_{\text{prior}}}
\newcommand{\dd}{\mathrm{d}}

\newcommand{\norm}[1]
{\left\Vert#1\right\Vert}

\newcommand{\parr}[1]{\left (#1\right )}
\newcommand{\brac}[1]{\left [#1\right ]}
\newcommand{\ip}[1]{\left \langle #1 \right \rangle }
\newcommand{\Real}{\mathbb R}
\newcommand{\uref}{u^{\text{target}}}
\newcommand{\Lcond}{\mathcal{L}_{\text{CFM}}}
\newcommand{\Lmarg}{\mathcal{L}_{\text{FM}}}
\newcommand{\uflow}{u^{\text{flow}}}

\newcommand{\uforw}{u^{\text{forw}}}
\newcommand{\sigforw}{\sigma^{\text{forw}}}
\newcommand{\psiref}{\psi^{\text{target}}}
\newcommand{\Nat}{\mathbb N}

\newcommand{\too}{\rightarrow}

\newcommand{\diag}{\textrm{diag}} %diagonal matrix
\newcommand{\divv}{\mathrm{div}} %divergence
\newcommand{\Unif}{\text{Unif}}
\newcommand{\dap}{z}

\newcommand{\defe}{\coloneqq}%{\stackrel{\text{def}}{=}}

\definecolor{mygray}{gray}{0.95}

\def\eqref#1{equation~\ref{#1}}
\def\1{\bm{1}}

\DeclareMathAlphabet{\mathsfit}{\encodingdefault}{\sfdefault}{m}{sl}
\SetMathAlphabet{\mathsfit}{bold}{\encodingdefault}{\sfdefault}{bx}{n}

\def\gN{{\mathcal{N}}}

\def\sP{{\mathbb{P}}}

\renewcommand{\eqref}[1]{\labelcref{#1}}

\newcommand{\pdata}{p_{\rm{data}}}
\newcommand{\pinit}{p_{\rm{init}}}
\newcommand{\E}{\mathbb{E}}

\newcommand{\R}{\mathbb{R}}

\newcolumntype{C}[1]{>{\Centering}m{#1}}
\newcolumntype{Z}[1]{>{\Left}m{#1}}

\usepackage{algorithm}
\usepackage{algorithmic}

\usepackage{caption}
\usepackage{subcaption}

\usepackage{epigraph}
\date{}
\begin{document}

%\vspace{1em} % Adjust vertical space as needed
 
%\begin{center}
%{MIT Class 6.S184: \textit{Generative AI With Stochastic Differential Equations}, 2026}
%\end{center}
\vspace{0.0em}

\hrule
\vspace{0.7em}
\begin{center}
{\Large \bfseries \sffamily An Introduction to Flow Matching and Diffusion Models} \vspace{0.7em}\\
{\normalsize \sffamily 
Peter Holderrieth and Ezra Erives\\
\vspace{0.7em}
Website: \url{https://diffusion.csail.mit.edu/}}
\end{center}

\vspace{0.6em}

\hrule

\doparttoc

\begin{adjustwidth}{1.0cm}{1.0cm}
\tableofcontents
\end{adjustwidth}

\vspace{1.5em}
\hrule

\newpage
\section{Introduction}
\label{sec:introduction}
\epigraph{Creating noise from data is easy; creating data from noise is generative modeling.}{Song et al. \cite{yangsong_sde}}

\subsection{Overview}
In recent years, we all have witnessed a tremendous revolution in artificial intelligence (AI). 
Image generators like \themeit{Nano Banana} or \themeit{Stable Diffusion 3} can generate photorealistic and artistic images across a diverse range of styles, video models like Meta's \themeit{VEO-3} can generate highly realistic movie clips, and large language models like \themeit{ChatGPT} can generate seemingly human-level responses to text prompts. At the heart of this revolution lies a new ability of AI systems: the ability to \themebf{generate} objects. While previous generations of AI systems were mainly used for \themebf{prediction}, these new AI system are creative: they dream or come up with new objects based on user-specified input. Such \themebf{generative AI} systems are at the core of this recent AI revolution.  

The goal of this class is to teach you two of the most widely used generative AI algorithms: \themebf{denoising diffusion models} \citep{song2020score} and \themebf{flow matching} \citep{lipman2022flow,liu2022flow, albergo2023stochastic, lipman2024flow}. These models are the backbone of the best image, audio, and video generation models (e.g., \themeit{Nano Banana}, \themeit{FLUX}, or \themeit{VEO-3}), and have most recently became the state-of-the-art in scientific applications such as protein structures (e.g., \themeit{AlphaFold3} is a diffusion model). Without a doubt, understanding these models is truly an extremely useful skill to have.

All of these generative models generate objects by iteratively converting  \themebf{noise} into \themebf{data}. This evolution from noise to data is facilitated by the simulation of \themebf{ordinary or stochastic differential equations (ODEs/SDEs)}. Flow matching and denoising diffusion models are a family of techniques that allow us to construct, train, and simulate, such ODEs/SDEs at large scale with deep neural networks. While these models are rather simple to implement, the technical nature of SDEs can make these models difficult to understand. In this course, we provide a self-contained introduction to the necessary mathematical toolbox regarding differential equations to enable you to systematically understand these models. We then explain step-by-step the modern stack of state-of-the-art image and video generators. Beyond being widely applicable, we believe that the theory behind flow and diffusion models is elegant in its own right. Therefore, most importantly, we hope that this course will be a lot of fun to you. 

\begin{remarkbox}[Additional Resources] 
While these lecture notes are self-contained, there are two additional resources that we encourage you to use:
\begin{enumerate}
    \item \textbf{Lecture recordings:} These guide you through each section in a lecture format.
    \item \textbf{Labs:} These guide you in implementing your own diffusion model from scratch. We highly recommend that you ``get your hands dirty'' and code.
\end{enumerate}
You can find these on our course website: \url{https://diffusion.csail.mit.edu/}.
\end{remarkbox}

\subsection{Course Structure}

We give a brief overview over of this document.
\begin{itemize}
\item \textbf{\sffamily Section \ref{sec:introduction}, Generative Modeling as Sampling:} We formalize what it means to ``generate'' an image, video, protein, etc. We will translate the problem of e.g., ``how to generate an image of a dog?'' into the more precise problem of sampling from a probability distribution.
\item \textbf{\sffamily Section \ref{sec:odes_sdes}, Flow and Diffusion Models:} We explain the machinery of generation. As you can guess by the name of this class, this machinery consists of simulating ordinary and stochastic differential equations. We provide an introduction to differential equations and explain how to use them to construct generative models.
\item \textbf{\sffamily Section \ref{sec:flow_matching}, Flow Matching:} Next, we  explain and derive \emph{flow matching}, a simple and scalable algorithm lying at the core of all afore-mentioned large-scale generative models such as Stable Diffusion, Nano Banana, or SORA.
\item \textbf{\sffamily Section \ref{sec:training_generative_models}, Score Matching:} We study \emph{score functions} and how they can be learnt via \emph{score matching}. Not only is this the training algorithm for diffusion models, but it unlocks SDE sampling and guidance.
\item \textbf{\sffamily Section \ref{sec:guidance}, Guidance:} We learn how to condition our samples on a prompt (e.g. ``an image of a cat'') and how we can enforce adherence to such a prompt via classifier-free guidance.
\item  \textbf{\sffamily Section \ref{sec:image_generation}, Latent Spaces, Neural Network architectures:}  We discuss how one builds large-scale image and video generators such as \emph{Nano Banana}. This includes common neural network architectures and how to build things in latent space. We also survey state-of-the-art models.
\item  \textbf{\sffamily Section \ref{sec:discrete_diffusion} (Optional), Discrete Diffusion Models:} We learn how to translate the principles of diffusion models from Euclidean space to discrete data such as language. This enables the construction of large language models using the principles of diffusion models.
\end{itemize}

\paragraph{Required background.} Due to the technical nature of this subject, we recommend some base level of mathematical maturity, and in particular some familiarity with probability theory. For this reason, we included a brief reminder section on probability theory in \cref{appendix:prob_theory_reminder}. Don't worry if some of the concepts there are unfamiliar to you.

\subsection{Generative Modeling As Sampling}
\label{subsec:gm_as_sampling}
Let's begin by thinking about various data types, or \themebf{data modalities}, that we might encounter, and how we will go about representing them numerically:
\begin{enumerate}
    \item \textbf{\sffamily Image: }Consider images with $H \times W$ pixels where $H$ describes the height and $W$ the width of the image, each with three color channels (RGB). For every pixel and every color channel, we are given an intensity value in $\mathbb{R}$. Therefore, an image can be represented by an element $\dap\in\mathbb{R}^{H \times W \times 3}$.
    \item \textbf{\sffamily Video: }A video is simply a series of images in time. If we have $T$ time points or \themebf{frames}, a video would therefore be represented by an element $\dap\in\mathbb{R}^{T\times H \times W \times 3}$.
    \item \textbf{\sffamily Molecular structure: }A naive way would be to represent the structure of a molecule by a matrix \\$z=(z^1,\dots,z^N)\in\mathbb{R}^{3\times N}$ where $N$ is the number of atoms in the molecule and each $z^i\in\mathbb{R}^3$ describes the location of that atom. Of course, there are other, more sophisticated ways of representing such a molecule.
\end{enumerate}
In all of the above examples, the object that we want to generate can be mathematically represented as a vector (potentially after flattening). Therefore, throughout this document, we will have:
\begin{ideabox}[Objects as Vectors]
    We identify the objects being generated as vectors $z \in \mathbb{R}^d$.
\end{ideabox}
A notable exception to the above is text data, which is typically modeled as a discrete object by language models (such as \emph{ChatGPT}). While continuous data $z\in \R^d$ is our main focus, we also study text generation in \cref{sec:discrete_diffusion}.

\paragraph{Generation as Sampling.}Let us define what it means to ``generate'' something. For example, let's say we want to generate an image of a dog. Naturally, there are \emph{many} possible images of dogs that we would be happy with. In particular, there is no one single ``best'' image of a dog. Rather, there is a spectrum of images that fit better or worse. In machine learning, it is common to realize this diversity of possible images as a \emph{probability distribution} over the \emph{space of images}. We call such a distribution a \themebf{data distribution} and denote it as $\pdata$. Mathematically, one can think of $\pdata$ as a \themebf{probability density}, i.e. a function $\pdata:\R^d\to\mathbb{R}_{\geq 0}$ that assigns each possible object $z\in \R^d$ a \emph{likelihood} $\pdata(z)\geq 0$. In the example of dog images, this distribution would therefore give higher likelihood $\pdata(z)$ to images $z$ that look more like a dog. Therefore, how "good" an image/video/molecule fits - a rather subjective statement - is replaced by how "likely" it is under the data distribution $\pdata$. With this, we can mathematically express the task of generation as sampling from the (unknown) distribution $\pdata$:
\begin{ideabox}[Generation as Sampling]
    Generating an object $z$ is modeled as sampling from the data distribution $z\sim \pdata$.
\end{ideabox}
A \themebf{generative model} is a machine learning model that allows us to generate samples from $\pdata$. In machine learning, we require data to train models. In generative modeling, we usually assume access to a finite number of examples sampled independently from $\pdata$, which together serve as a proxy for the true distribution.
\begin{ideabox}[Dataset]
    A dataset consists of a finite number of samples $z_1, \dots, z_N \sim \pdata$.
\end{ideabox}
For images, we might construct a dataset by compiling publicly available images from the internet. For videos, we might similarly look to use YouTube. For protein structures, sources like the RCSB Protein Data Bank (PDB) provide hundreds of thousands of experimentally resolved structures. As the size of our dataset grows very large, it becomes an increasingly better representation of the underlying distribution $\pdata$.

\paragraph{Guided/Conditional Generation.} In many cases, we want to generate an object \themebf{conditioned} on some data $y$. For example, we might want to generate an image conditioned on $y=$``a dog running down a hill covered with snow with mountains in the background''. We can rephrase this as sampling from a \themebf{conditional distribution}:
\begin{ideabox}[Guided Generation]
Guided generation involves sampling from $z\sim \pdata(\cdot | y)$, where $y$ is a conditioning variable.
\end{ideabox}
We call $\pdata(\cdot|y)$ the \themebf{guided data distribution}. The guided generative modeling task typically involves learning to condition on an arbitrary, rather than fixed, choice of $y$. Using our previous example, we might alternatively want to condition on a different text prompt, such as $y=$``a photorealistic image of a cat blowing out birthday candles''. We therefore seek a single model which may be conditioned on any such choice of $y$. It turns out that techniques for unconditional generation are readily generalized to the conditional case. Therefore, for the first 3 sections, we will focus almost exclusively on the unconditional case (keeping in mind that conditional generation is what we're building towards).

% \paragraph{Generative Models.} \ee{General note: this section makes a very large intuitive jump (with flow models) in just two sentences. The central point is therefore not ``generative modeling'' but specifically the idea of transforming samples from a simple Gaussian (a flow). I think the title should reflect this. See revised section below.} A \themebf{generative model} is a machine learning model that allows us to generate samples from $\pdata$. For this, we assume that we have access to some simple distribution $\pinit$ that we can easily sample from, e.g. $\pinit=\mathcal{N}(0,I_d)$ could be a Gaussian distribution. The goal of a generative model is then to transform samples from $X\sim \pinit$ into samples from $\pdata$. We note that $\pinit$ does not have to be simple or Gaussian at all. In fact, there are interesting usecases for leveraging this flexibility (see \ph{reference to where this is discussed}). We just call it $\pinit$ because in the majority of applications we think of it as a simple Gaussian.

% \ee{Revise previous section: 
\paragraph{Generative Models.} Abstractly speaking, a generative model is an algorithm that returns samples from $z\sim\pdata$ (or at least approximately). If $\pdata$ is the distribution of images of dogs, this algorithm would return random images of dogs. In this course, we will focus on the specific construction of generative models using flow or diffusion models as these represent the current state-of-the-art. However, it is important to keep in mind that many other generative models were developed (and maybe even more that will be discovered in the future).

\begin{summarybox}[Generation as Sampling] 
\label{par:summary}
We 
summarize the findings of this section:
\begin{enumerate}
\item In this work, we mainly consider the task of generating objects that are represented as vectors $z\in\mathbb{R}^d$ such as images, videos, and molecular structures.
\item Generation is the task of generating samples from a probability distribution $\pdata$ having access to a dataset of samples $z_1,\dots,z_N\sim \pdata$ during training. 
\item Guided generation assumes that we condition the distribution on a label $y$ and we want to sample from $\pdata(\cdot|y)$ having access to data set of pairs $(z_1,y)\dots,(z_N,y)$ during training.
\item Our goal is to construct a generative model, i.e. a model that returns samples from $\pdata$ after training.
\end{enumerate}
\end{summarybox}

\newpage
\section{Flow and Diffusion Models}
\label{sec:odes_sdes}
In the previous section, we formalized generative modeling as sampling from a data distribution $\pdata$. Further, we formalized our goal: To construct a generative model, i.e. an algorithm that returns samples $z\sim \pdata$. In this section, we describe how a generative model can be built as the simulation of a suitably constructed differential equation. For example, flow matching and diffusion models involve simulating \themebf{ordinary differential equations} (ODEs) and \themebf{stochastic differential equations} (SDEs), respectively. The goal of this section is therefore to define and construct these generative models as they will be used throughout the remainder of the notes. Specifically, we first define ODEs and SDEs, and discuss their simulation. Second, we describe how to parameterize an ODE/SDE using a deep neural network. This leads to the definition of a flow and diffusion model and the fundamental algorithms to sample from such models. In later sections, we then explore how to train these models.

% \paragraph{Note.} We remark that there are other ways to transform $\pinit$ into $\pdata$ other than ODEs/SDEs. We focus on ODEs/SDEs because they represent the current state-of-the-art but just keep in mind that many other ways have been proposed over the course of the history of machine learning offering unique advantages and disadvantages.

\subsection{Flow Models}

\begin{figure}
    \centering
    \begin{tabular}{ccc}
         \includegraphics[width=0.3\textwidth]{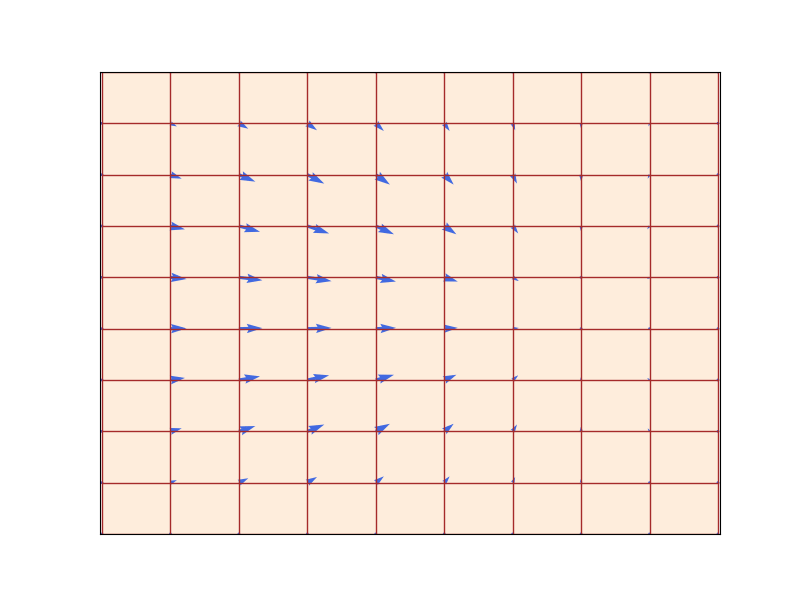} &
         \includegraphics[width=0.3\textwidth]{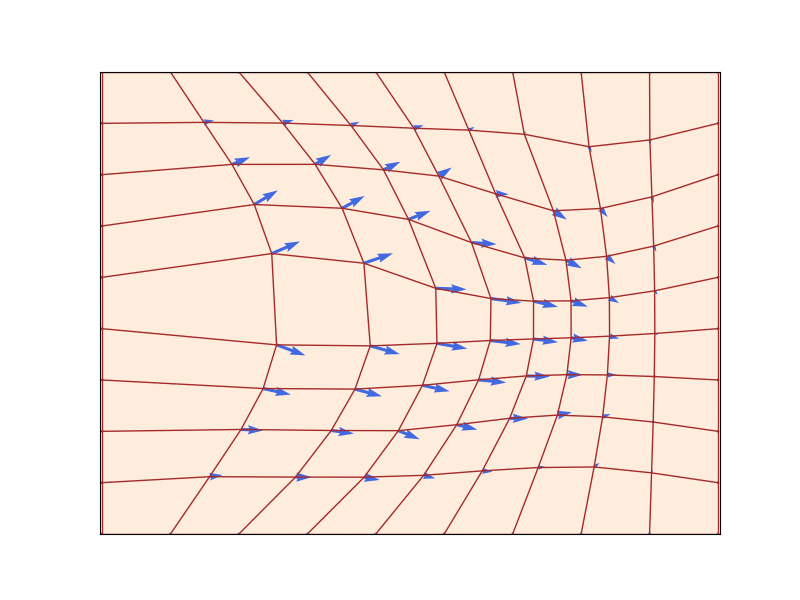} &
         \includegraphics[width=0.3\textwidth]{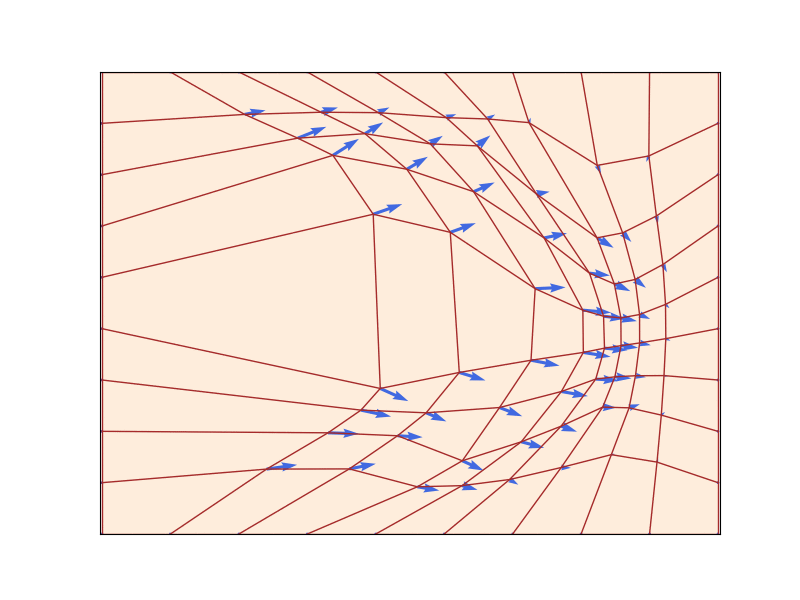} 
    \end{tabular}
    \caption{A flow $\psi_t:\Real^d\too \Real^d$ (red square grid) is defined by a velocity field $u_t :\Real^d\too\Real^d$ (visualized with blue arrows) that prescribes its instantaneous movements at all locations (here, $d=2$). We show three different times $t$. As one can see, a flow is a diffeomorphism that "warps" space. Figure from \citep{lipman2024flow}.}
    \label{fig:flow}
\end{figure}

We start by defining \themebf{ordinary differential equations (ODEs)}. A solution to an ODE is defined by a \themebf{trajectory}, i.e. a function of the form
\begin{align*}
X: [0,1] \to \R^d, \quad t \mapsto X_t,
\end{align*}
that maps from time $t$ to some location in space $\mathbb{R}^d$. Every ODE is defined by a \themebf{vector field} $u$, i.e. a function of the form
\begin{align*}
u:\mathbb{R}^d\times [0,1]\to \R^d,\quad (x,t)\mapsto u_t(x),
\end{align*}
i.e. for every time $t$ and location $x$ we get a vector $u_t(x)\in\R^d$ specifying a velocity in space (see  \cref{fig:flow}). An ODE imposes a condition on a trajectory: we want a trajectory $X$ that ``follows along the lines'' of the vector field $u_t$, starting at the point $x_0$. We may formalize such a trajectory as being the solution to the equation:
\begin{subequations}
    \begin{align} 
      \frac{\dd}{\dd t}X_{t} &= u_t(X_t) &&\blacktriangleright\,\,\text{ODE}\label{e:ode_ode}\\
      X_0&= x_0            &&\blacktriangleright\,\,\text{initial conditions}\label{e:ODE_boundary} 
    \end{align}
\end{subequations}
\Cref{e:ode_ode} requires that the derivative of $X_t$ is specified by the direction given by $u_t$. \Cref{e:ODE_boundary} requires that we start at $x_0$ at time $t=0$. We may now ask: if we start at $X_0 = x_0$ at $t=0$, where are we at time $t$ (what is $X_t$)? This question is answered by a function called the \themebf{flow}, which is a solution to the ODE
\begin{subequations}\label{e:flow}
    \begin{align}
    \psi:\R^d\times [0,1]\to& \R^d,\quad (x_0,t)\mapsto \psi_t(x_0)\\
      \frac{\dd}{\dd t}\psi_{t}(x_0) &= u_t(\psi_{t}(x_0)) &&\blacktriangleright\,\, \text{flow ODE}\label{e:flow_flow}\\
      \psi_{0}(x_0)             &= x_0                &&\blacktriangleright\,\,\text{flow initial conditions}\label{e:flow_boundary} 
    \end{align}
\end{subequations}
For a given initial condition $X_0=x_0$, a trajectory of the ODE is recovered via $X_t = \psi_t(X_0)$. Therefore, vector fields, ODEs, and flows are, intuitively, three descriptions of the same object: \textbf{vector fields define ODEs whose solutions are flows}. As with every equation, we should ask ourselves about an ODE: Does a solution exist and if so, is it unique? A fundamental result in mathematics is "yes!" to both, as long as we impose weak assumptions on $u_t$:

\begin{theorem}[Flow existence and uniqueness]
\label{thm:ode_existence_and_uniqueness}
If $u:\R^d\times[0,1]\to\R^d$ is continuously differentiable with a bounded derivative, then the ODE in \eqref{e:flow} has a unique solution given by a flow $\psi_t$. In this case, $\psi_t$ is a \themebf{diffeomorphism} for all $t$, i.e. $\psi_t$ is continuously differentiable with a continuously differentiable inverse $\psi_t^{-1}$.
\end{theorem}
Note that the assumptions required for the existence and uniqueness of a flow are almost always fulfilled in machine learning, as we use neural networks to parameterize $u_t(x)$ and they always have bounded derivatives. Therefore, \cref{thm:ode_existence_and_uniqueness} should not be a concern for you but rather good news: \textbf{flows exist and are unique solutions to ODEs in our cases of interest.} A proof can be found in \citep{perko2013differential,coddington1956theory}.

\begin{examplebox}[Linear Vector Fields]
Let us consider a simple example of a vector field $u_t(x)$ that is a simple linear function in $x$, i.e. $u_t(x)=-\theta x$ for $\theta>0$. Then the function
\begin{align}
    \label{e:flow_linear_vf}
    \psi_t(x_0) =  \exp\left(-\theta t\right)x_0
\end{align}
defines a flow $\psi$ solving the ODE in \cref{e:flow}. You can check this yourself by checking that $\psi_0(x_0)=x_0$ and computing
\begin{align*}
    \frac{\dd}{\dd t}\psi_t(x_0) \overset{\eqref{e:flow_linear_vf}}{=}\frac{\dd}{\dd t}\left(\exp\left(-\theta t\right)x_0\right)
    \overset{(i)}{=}-\theta\exp\left(-\theta t\right)x_0\overset{\eqref{e:flow_linear_vf}}{=}-\theta\psi_t(x_0)=u_t(\psi_t(x_0)),
\end{align*}
where in (i) we used the chain rule. In \cref{fig:bm_ou_process}, we visualize a flow of this form converging to $0$ exponentially.
\end{examplebox}

\paragraph{Simulating an ODE.}
In general, it is not possible to compute the flow $\psi_t$ explicitly if $u_t$ is not as simple as in the previous example. In these cases, one uses \themebf{numerical methods} to simulate ODEs. Fortunately, this is a classical and well researched topic in numerical analysis, and a myriad of powerful methods exist \citep{iserles2009first}. One of the simplest and most intuitive methods is the \themebf{Euler method}. In the Euler method, we initialize with $X_0=x_0$ and update via
\begin{equation}
\label{e:euler_method}
X_{t+h} = X_t + h u_t(X_t)\quad (t=0,h,2h,3h,\dots,1-h)
\end{equation}
where $h=n^{-1}>0$ is the \themebf{step size} and $n \in \Nat$ is the number of simulation steps. For this class, the Euler method will be good enough. To give you a taste of a more complex method, let us consider \themebf{Heun's method} defined via the update rule
\begin{alignat*}{3}
    X_{t+h}'&=X_t+hu_t(X_t)\quad &&\blacktriangleright\,\, \text{initial guess of new state (same as Euler step)}\\
    X_{t+h} &= X_{t} + \frac{h}{2}(u_t(X_t)+u_{t+h}(X_{t+h}'))\quad &&\blacktriangleright\,\,\text{update with average $u$ at current and guessed state}
\end{alignat*}
Intuitively, Heun's method is as follows: it takes a first guess $X_{t+h}'$ of what the next step could be but corrects the direction initially taken via an updated guess.

\paragraph{Flow models.} We can now construct a generative model via an ODE by making the vector field a \themebf{neural network vector field} $u_t^\theta$. For now, we simply mean that $u_t^\theta$ is a parameterized function  $u_t^\theta:\mathbb{R}^d\times [0,1]\to\mathbb{R}^d$ with parameters $\theta$. Later, we will discuss particular choices of neural network architectures. Remember that our goal was to generate samples $z\sim \pdata$ from a distribution $\pdata$. In particular, these samples must be \emph{random}. Note though that an ODE itself is not random but fully deterministic. To inject some randomness, we simple make the initial condition $X_0$ random. Specifically, we choose an \themebf{initial distribution} $\pinit$. In most cases, we set $\pinit=\mathcal{N}(0,I_d)$ to be a simple standard Gaussian. Most importantly, whatever distribution you choose, it must be one that we can easily sample from at inference-time. A \themebf{flow model} is then described by the ODE
\begin{align*}
    X_0 &\sim \pinit  &&\blacktriangleright\,\,\text{random initialization} \\
    \frac{\dd}{\dd t}X_t &= u_t^\theta(X_t) &&\blacktriangleright\,\,\text{ODE}
\end{align*}
Our goal is to make the endpoint  $X_1$ of the trajectory have distribution $\pdata$, i.e.
\begin{align*}
    X_1 \sim \pdata \quad \Leftrightarrow \quad \psi_1^\theta(X_0)\sim \pdata
\end{align*}
where $\psi_t^\theta$ describes the flow induced by $u_t^\theta$. Note however: although it is called \themeit{flow model}, \textbf{the neural network parameterizes the vector field, not the flow}. In order to compute the flow, we need to simulate the ODE. In \cref{alg:sampling_flow_model}, we summarize the procedure how to sample from a flow model.

\begin{algorithm}[h]
\caption{Sampling from a Flow Model with Euler method}
\label{alg:sampling_flow_model}
\begin{algorithmic}[1]
\REQUIRE Neural network vector field $u_t^\theta$, number of steps $n$
\STATE Set $t=0$
\STATE Set step size $h=\frac{1}{n}$
\STATE Draw a sample $X_0\sim \pinit$
\FOR{$i=1,\dots,n$}
    \STATE $X_{t+h} = X_{t} + h u_t^\theta(X_t)$
    \STATE Update $t\leftarrow t+h$
\ENDFOR
\RETURN $X_1$
\end{algorithmic}
\end{algorithm}

\subsection{Diffusion Models}
Stochastic differential equations (SDEs) extend the deterministic trajectories from ODEs with \themebf{stochastic} trajectories. A stochastic trajectory is commonly called a \themebf{stochastic process} $(X_t)_{0\leq t\leq 1}$ and is given by
\begin{align*}
    X_t \text{ is a random variable for every } 0\leq t\leq 1\\
    X: [0,1] \to \R^d, \quad t \mapsto X_t\text{ is a random trajectory for every draw of }X
\end{align*}
In particular, when we simulate the same stochastic process twice, we might get different outcomes because the dynamics are designed to be random.

\paragraph{Brownian Motion.} SDEs are constructed via a \themebf{Brownian motion} - a fundamental stochastic process that came out of the study physical diffusion processes. You can think of a Brownian motion as a continuous random walk.
\begin{wrapfigure}{r}{0.4\textwidth}
  %\begin{center}
\includegraphics[width=0.4\textwidth]{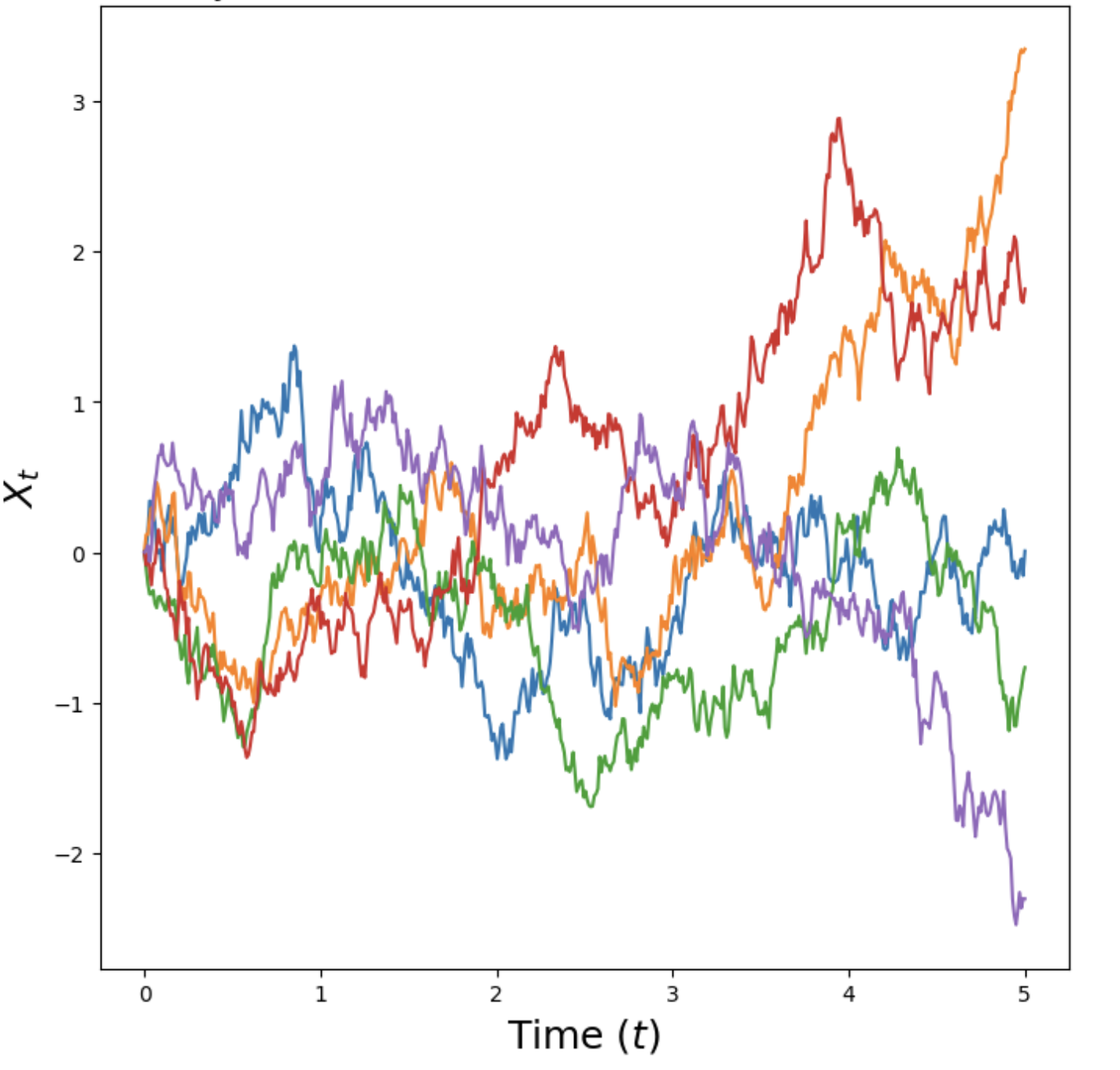}
  %\end{center}
  \vspace{-20pt}
  \caption{Sample trajectories of a Brownian motion $W_t$ in dimension $d=1$ simulated using \cref{eq:brownian_motion_simulation}.}
  \label{fig:brownian_motion_trajectories}
  \vspace{-10pt}
\end{wrapfigure}
Let us define it: A \themebf{Brownian motion} $W = (W_t)_{0\leq t\leq 1}$ is a stochastic process such that $W_0=0$, the trajectories $t\mapsto W_t$ are continuous, and the following two conditions hold:
\begin{enumerate}
    \item \textbf{Normal increments: }$W_{t}-W_{s}\sim \mathcal{N}(0,(t-s)I_d)$ for all $0\leq s<t$, i.e. increments have a Gaussian distribution with variance increasing linearly in time ($I_d$ is the identity matrix).
    \item \textbf{Independent increments: }For any $0\leq t_0<t_1<\dots <t_n=1$, the increments $W_{t_1}-W_{t_0},\dots,W_{t_n}-W_{t_{n-1}}$ are independent random variables.
\end{enumerate}

Brownian motion is also called a \themebf{Wiener process}, which is why we denote it with a "$W$".\footnote{Norbert Wiener was a famous mathematician who taught at MIT. You can still see his portraits hanging at the MIT math department.} We can easily simulate a Brownian motion approximately with step size $h>0$ by setting $W_0=0$ and updating
\begin{align}
    \label{eq:brownian_motion_simulation}
    W_{t+h} =& W_{t} + \sqrt{h}\epsilon_t,\quad \epsilon_t\sim\mathcal{N}(0,I_d)\quad (t=0,h,2h,\dots,1-h)
\end{align}
In \cref{fig:brownian_motion_trajectories}, we plot a few example trajectories of a Brownian motion.  Brownian motion is as central to the study of stochastic processes as the Gaussian distribution is to the study of probability distributions. From finance to statistical physics to epidemiology, the study of Brownian motion has far reaching applications beyond  machine learning. In finance, for example, Brownian motion is used to model the price of complex financial instruments. Also just as a mathematical construction, Brownian motion is fascinating: For example, while the paths of a Brownian motion are continuous (so that you could draw it without ever lifting a pen), they are infinitely long (so that you would never stop drawing).
%Furhter (2) Any continuous-time Markov process with zero mean and independent, stationary increments is automatically a Brownian motion.

\begin{figure}
    \centering
    \begin{tabular}{ccc}
         \includegraphics[width=\textwidth]{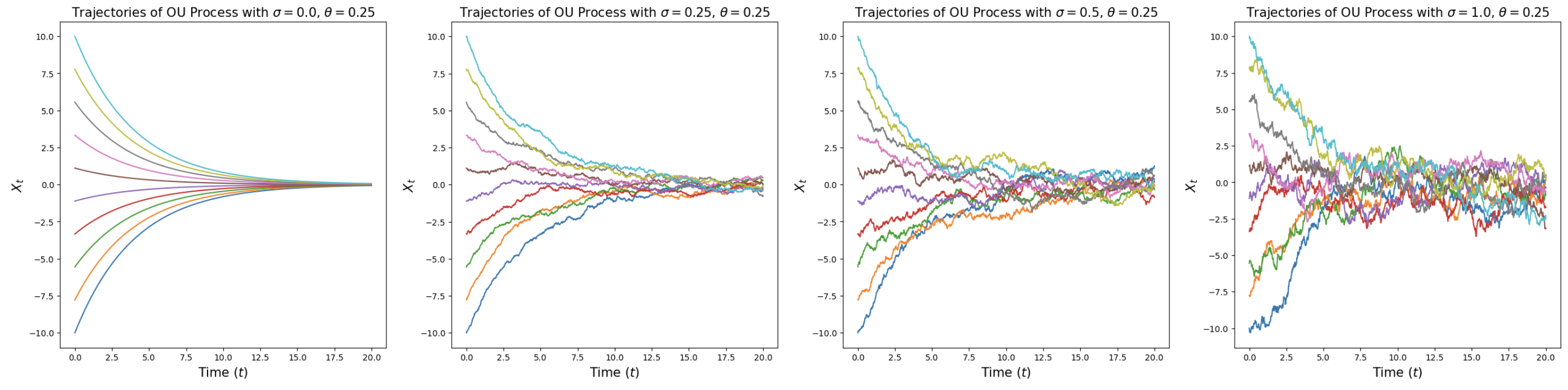} &
         % \includegraphics[width=0.22\textwidth]{assets/flow_velocity/flow_v_5.png} &
         % \includegraphics[width=0.3\textwidth]{fm_guide_assets/flow_10.png} &
         % % \includegraphics[width=0.22\textwidth]{assets/flow_velocity/flow_v_14.png} &
         % \includegraphics[width=0.3\textwidth]{fm_guide_assets/flow_16.png} 
    \end{tabular}
    \caption{\label{fig:bm_ou_process} Illustration of Ornstein-Uhlenbeck processes (\cref{eq:ohrstein_uhlenbeck_process}) in dimension $d=1$ for $\theta=0.25$ and various choices of $\sigma$ (increasing from left to right). For $\sigma=0$, we recover a flow (smooth, deterministic trajectories) that converges to the origin as $t \to \infty$. For $\sigma>0$ we have random paths which converge towards the Gaussian $\mathcal{N}(0,\frac{\sigma^2}{2\theta})$ as $t\to\infty$.}
\end{figure} 

\paragraph{From ODEs to SDEs.} The idea of an SDE is to extend the deterministic dynamics of an ODE by adding stochastic dynamics driven by a Brownian motion. Because everything is stochastic, we may no longer take the derivative as in \cref{e:ode_ode}. Hence, we need to find an \textbf{equivalent formulation of ODEs that does not use derivatives}. For this, let us therefore rewrite trajectories $(X_t)_{0\leq t\leq 1}$ of an ODE as follows:
\begin{alignat*}{3}
    \frac{\dd}{\dd t} X_t &=  u_t(X_t) \quad &&\blacktriangleright\,\,\text{expression via derivatives}\\
    \overset{(i)}{\Leftrightarrow} \quad  \frac{1}{h}\left(X_{t+h}-X_{t}\right)&=u_t(X_t) + R_t(h)&&\\
\Leftrightarrow \quad X_{t+h} &= X_{t}+hu_t(X_t) + hR_t(h)\quad &&\blacktriangleright\,\,\text{expression via infinitesimal updates}
\end{alignat*}
where  $R_t(h)$ describes a negligible function for small $h$, i.e. such that $\lim\limits_{h\to 0}R_t(h)=0$, and in $(i)$ we simply use the definition of derivatives. The derivation above simply restates what we already know: A trajectory $(X_t)_{0 \le t \le 1}$ of an ODE takes, at every timestep, a small step in the direction $u_t(X_t)$. We may now amend the last equation to make it stochastic: A trajectory $(X_t)_{0 \le t \le 1}$ of an SDE takes, at every timestep, a small step in the direction $u_t(X_t)$ \themeit{plus} some contribution from a Brownian motion:
\begin{align}
    \label{e:infinitesimal_updates_sdes}
    X_{t+h} = X_{t}+\underbrace{hu_t(X_t)}_{\text{deterministic}} + \sigma_t\underbrace{(W_{t+h}-W_{t})}_{\text{stochastic}}+\underbrace{hR_t(h)}_{\text{error term}}
\end{align}
where $\sigma_t\geq 0$ describes the \themebf{diffusion coefficient} and $R_t(h)$ describes a stochastic error term such that the standard deviation $\mathbb{E}[\|R_t(h)\|^2]^{1/2}\to 0$ goes to zero for $h\to 0$. The above describes a \themebf{stochastic differential equation (SDE)}. It is common to denote it in the following symbolic notation:
\begin{subequations}\label{e:sde_generic}
    \begin{align} 
      \dd X_t &= u_t(X_t)\dd t + \sigma_t\dd W_t &&\blacktriangleright\,\,\text{SDE}\\
      X_0 &= x_0               &&\blacktriangleright\,\,\text{initial condition}
    \end{align}
\end{subequations}
However, always keep in mind that the "$\dd X_t$"-notation above is a purely informal notation of \cref{e:infinitesimal_updates_sdes}. Unfortunately, SDEs do not have a flow map $\phi_t$ anymore. This is because the value $X_t$ is not fully determined by $X_0\sim \pinit$ anymore as the evolution itself is stochastic. Still, in the same way as for ODEs, we have:
\begin{theorem}[SDE Solution Existence and Uniqueness]
\label{thm:sde_existence_and_uniqueness}
If $u:\R^d\times[0,1]\to\R^d$ is continuously differentiable with a bounded derivative and $\sigma_t$ is continuous, then the SDE in \eqref{e:sde_generic} has a solution given by the unique stochastic process $(X_t)_{0\leq t\leq 1}$ satisfying \cref{e:infinitesimal_updates_sdes}.
\end{theorem}
If this was a stochastic calculus class, we would spend several lectures proving this theorem and constructing SDEs with full mathematical rigor, i.e. constructing a Brownian motion from first principles and constructing the process $X_{t}$ via \themebf{stochastic integration}. As we focus on machine learning in this class, we refer to \citep{mao2007stochastic} for a more technical treatment. Finally, note that every ODE is also an SDE - simply with a vanishing diffusion coefficient $\sigma_t=0$. Therefore, for the remainder of this class, \textbf{when we speak about SDEs, we consider ODEs as a special case}.
% \begin{remarkbox}[A Note on Formality]
% If this was a stochastic calculus class, we would spend more time constructing SDEs with full mathematical rigor, i.e. constructing a Brownian motion from first principles and constructing the process $X_{t}$ via \emph{stochastic integrals}. \ee{Revise: Those of you familiar with stochastic calculus will note that our discussion of SDEs has thus far been quite informal. For example, we have technically not actually shown that a Brownian motion even exists. If this were a class on stochastic calculus, we might proceed more formally, by e.g., constructing a Brownian motion and then arriving at SDEs via the notion of a \themeit{stochastic integral}. Nevertheless, we feel that this extra machinery is overly cumbersome for the need at hand, and have therefore opted to take a more direct approach.}
% \end{remarkbox}

\begin{examplebox}[Ornstein-Uhlenbeck Process]
Let us consider a constant diffusion coefficient $\sigma_t=\sigma\geq 0$ and a constant linear drift $u_t(x)=-\theta x$ for $\theta>0$, yielding the SDE
\begin{align}
\label{eq:ohrstein_uhlenbeck_process}
\dd X_t = -\theta X_t\dd t + \sigma \dd W_t.
\end{align} 
A solution $(X_t)_{0 \le t \le 1}$ to the above SDE is known as an \themebf{Ornstein-Uhlenbeck (OU) process}. We visualize it in \cref{fig:bm_ou_process}. The vector field $-\theta x$ pushes the process back to its center $0$ (since the drift always points in the direction opposite to the current position), while the diffusion coefficient $\sigma$ always adds more noise. This process converges towards a Gaussian distribution $\mathcal{N}(0,\sigma^2/(2\theta))$ if we simulate it for $t\to \infty$. Note that for $\sigma=0$, we have a flow with linear vector field that we have studied in \cref{e:flow_linear_vf}.
\end{examplebox}

\paragraph{Simulating an SDE.} If you struggle with the abstract definition of an SDE so far, then don't worry about it. A more intuitive way of thinking about SDEs is given by answering the question: How might we simulate an SDE? The simplest such scheme is known as the \themebf{Euler-Maruyama method}, and is essentially to SDEs what the Euler method is to ODEs. Using the Euler-Maruyama method, we initialize $X_0=x_0$ and update iteratively via
\begin{align}
\label{e:euler_method_sdes}
    X_{t+h} = X_{t}+hu_t(X_t) + \sqrt{h}\sigma_t\epsilon_t,\quad \quad \epsilon_t \sim \mathcal{N}(0,I_d)
\end{align}
where $h=n^{-1}>0$ is a step size hyperparameter for $n \in \Nat$. In other words, to simulate using the Euler-Maruyama method, we take a small step in the direction of $u_t(X_t)$ as well as add a little bit of Gaussian noise scaled by $\sqrt{h}\sigma_t$. When simulating SDEs in this class (such as in the accompanying labs), we will usually stick to the Euler-Maruyama method.

\begin{algorithm}[ht]
\caption{Sampling from a Diffusion Model (Euler-Maruyama  method)}
\label{alg:sampling_diffusion_model}
\begin{algorithmic}[1]
\REQUIRE Neural network $u_t^\theta$, number of steps $n$, diffusion coefficient $\sigma_t$
\STATE Set $t=0$
\STATE Set step size $h=\frac{1}{n}$
\STATE Draw a sample $X_0\sim \pinit$
\FOR{$i=1,\dots,n$}
    \STATE Draw a sample $\epsilon\sim \mathcal{N}(0,I_d)$
    \STATE $X_{t+h} = X_{t} + h u_t^\theta(X_t)+\sigma_t\sqrt{h}\epsilon$
    \STATE Update $t\leftarrow t+h$
\ENDFOR
\RETURN $X_1$
\end{algorithmic}
\end{algorithm}

\paragraph{Diffusion Models.} We can now construct a generative model via an SDE in the same way as we did for ODEs. Remember that our goal was to convert a simple distribution $\pinit$ into a complex distribution $\pdata$. Like for ODEs, the simulation of an SDE randomly initialized with $X_0\sim \pinit$ is a natural choice for this transformation. To parameterize this SDE, we can simply parameterize its central ingredient - the vector field $u_t$ - via a neural network $u_t^\theta$. A \themebf{diffusion model} is thus given by
\begin{align*}
    X_0 &\sim \pinit  &&\blacktriangleright\,\,\text{random initialization}\\
    \dd X_t &= u_t^\theta(X_t)\dd t + \sigma_t \dd W_t &&\blacktriangleright\,\,\text{SDE}
    \end{align*}
In \cref{alg:sampling_diffusion_model}, we describe the procedure by which to sample from a diffusion model with the Euler-Maruyama method. We summarize the results of this section as follows.
\begin{summarybox}[SDE generative model] Throughout this document, a \themebf{diffusion model} consists of a neural network $u_t^\theta$ with parameters $\theta$ that parameterize a vector field and a fixed  diffusion coefficient $\sigma_t$:
\begin{align*}
    \textbf{\sffamily Neural network: }&u^\theta:\R^d\times [0,1]\to \R^d,\,\, (x,t)\mapsto u_t^\theta(x)\text{  with parameters }\theta\\
    \textbf{\sffamily Fixed: }&\sigma_t:[0,1]\to [0,\infty),\,\, t\mapsto \sigma_t
\end{align*}
To obtain samples from our SDE model (i.e. generate objects), the procedure is as follows:
\begin{align*}
\textbf{\sffamily Initialization:}\quad X_0&\sim\pinit \quad  &&\blacktriangleright\,\,\text{Initialize with simple distribution, e.g. a Gaussian}\\
    \textbf{\sffamily Simulation:}\quad \dd X_t &= u_t^\theta(X_t)\dd t + \sigma_t\dd W_t\quad &&\blacktriangleright\,\,\text{Simulate SDE from 0 to 1}\\
    \textbf{\sffamily Goal:}\quad X_1 &\sim  \pdata \quad &&\blacktriangleright\,\,\text{Goal is to make $X_1$ have distribution $\pdata$}
\end{align*}
A diffusion model with $\sigma_t=0$ is a \themebf{flow model}.
\label{summary:diffusion_model}
\end{summarybox}
\newpage
\section{Flow Matching}
\label{sec:flow_matching}

In the previous section, we constructed flow and diffusion models as generative models parameterized by a neural network vector field $u_t^\theta$. However, we have not yet discussed how to \emph{train} them, i.e. how to optimize the parameters $\theta$ such that generative model returns something sensible, e.g. a nice-looking image or exciting video. Next, we discuss \themebf{flow matching} \citep{lipman2022flow,albergo2023stochastic, liu2022flow}, a  algorithm to train $u_t^\theta$ that is simple, scalable, and represents the current state-of-the-art.

In this section, we restrict ourselves to flow models, i.e. we have a neural network $u_t^\theta$ and obtain samples from the generative model by simulating the ODE 
\begin{align}
\label{e:sde_generative_model_restated_section_3}
 X_0\sim&\pinit,\quad \dd X_t = u_t^\theta(X_t)\dd t& \text{(Flow model)}
\end{align}
and using the endpoints $X_1$ fro $t=1$ as samples. As we discussed, our goal is that $X_1$ is distributed according to the data distribution $\pdata$, i.e. $X_1\sim \pdata$. Therefore, the question ``how to train'' the neural network is really the following question: \textbf{How do we optimize $\theta$ such that simulating the flow model in \cref{e:sde_generative_model_restated_section_3} results in samples from the data distribution $X_1\sim \pdata$?}

\begin{figure}[h!]
    \centering
    \begin{tabular}{ccc}
         \includegraphics[width=\textwidth]{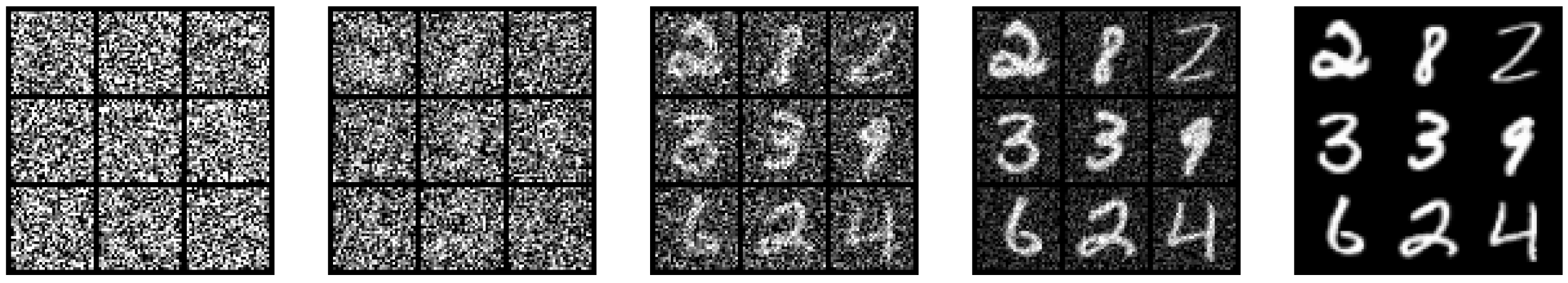} &
    \end{tabular}
    \caption{\label{fig:noising_image} Gradual interpolation from noise to data via  a Gaussian conditional probability path for a collection of images. Note that each image is a data point of dimension $d=32\times 32$, so we are plotting individual samples from the probability path, while in \cref{fig:cond_marginal_path_histograms} we plot the distribution as a 2d histogram.}
\end{figure}
\vspace{-1em}
\subsection{Conditional and Marginal Probability Path}

The first step of flow matching is to specify a \themebf{probability path}. Intuitively, a probability path specifies a gradual interpolation between noise $\pinit$ and data $\pdata$ (see \cref{fig:noising_image}). But why would we want that? Remember that our desired ODE trajectory fulfills $X_0\sim \pinit$ for $t=0$ and $X_1\sim \pdata$ for $t=1$. But what about times $0<t<1$ in between start and end? It turns out that we have some freedom to choose what should happen in between and this is what is mathematically formalized in a probability path.

In the following, for a data point $z\in\mathbb{R}^d$, we denote with $\delta_{z}$ the \themebf{Dirac delta} ``distribution''. This is the simplest distribution that one can imagine: sampling from $\delta_{z}$ always returns $z$ (i.e. it is deterministic). A \themebf{conditional (interpolating) probability path} is a set of distribution $p_t(x|z)$ over $\mathbb{R}^d$ such that:
\begin{align}
    % p_t(\cdot|z) \text{ distribution over }\R^d \quad (0\leq t\leq 1)\\
\label{eq:interpolating_condition_conditional_path}
    p_0(\cdot|z)=\pinit, \quad p_1(\cdot|z)=\delta_{z}\quad \text{ for all }z\in\R^d.
\end{align}
In other words, a conditional probability path gradually converts the initial distribution $\pinit$ into a \themeit{single} data point (see e.g. \cref{fig:noising_image}). You can think of a probability path as a trajectory in the space of distributions.

Every conditional probability path $p_t(x|z)$ induces a \themebf{marginal probability path} $p_t(x)$ defined as the distribution that we obtain by first sampling a data point $z\sim \pdata$ from the data distribution and then sampling from $p_t(\cdot|z)$:
\begin{align}
    \label{eq:marginal_prob_path}
    z&\sim\pdata, \quad x\sim p_t(\cdot|z)\quad \Rightarrow x\sim p_t &&\blacktriangleright\,\,\text{sampling from marginal path}\\
\label{eq:marginal_prob_path_integral}
    p_t(x) &= \int p_t(x|\dap) \pdata (z) \dd z &&\blacktriangleright\,\,\text{density of marginal path}
\end{align}
Note that we know how to sample from $p_t$ but we don't know the density values $p_t(x)$ as the integral is intractable (i.e. we can actually compute \cref{eq:marginal_prob_path} but not \cref{eq:marginal_prob_path_integral}). Check for yourself that because of the conditions on $p_t(\cdot|z)$ in \cref{eq:interpolating_condition_conditional_path}, the marginal probability path $p_t$ interpolates between $\pinit$ and $\pdata$:
\begin{align}
\label{eq:noise_interpolation}
p_0 =\pinit\quad\text{and}\quad p_1=\pdata.\quad\quad\quad\quad \blacktriangleright\,\,\text{noise-data interpolation}
\end{align}
The - by far - most important example of a probability path is the Gaussian probability path - hence, we strongly recommend reading the next example thoroughly.
\begin{figure}[!t]
    \centering
    \begin{tabular}{ccc}
\includegraphics[width=\textwidth]{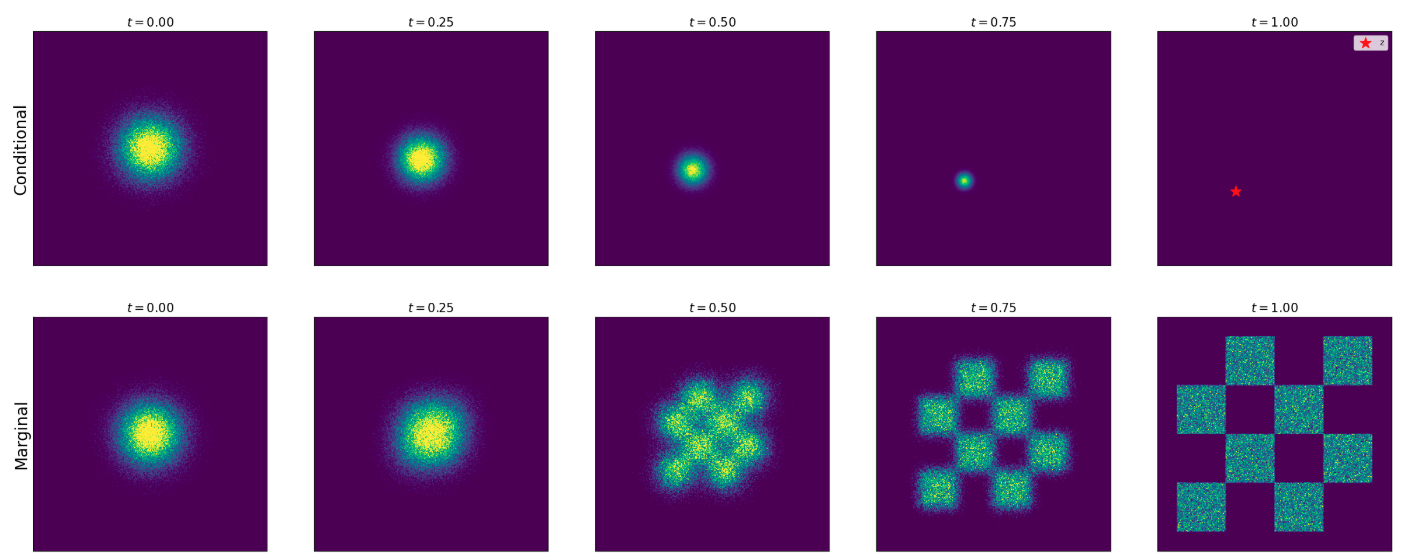} &
    \end{tabular}
\caption{\label{fig:cond_marginal_path_histograms}Illustration of a conditional (top) and marginal (bottom) probability path. Here, we plot a Gaussian probability path with $\alpha_t=t,\beta_t=1-t$. The conditional probability path interpolates a Gaussian $\pinit=\mathcal{N}(0,I_d)$ and $\pdata=\delta_{z}$ for single data point $z$. The marginal probability path interpolates a Gaussian and a data distribution $\pdata$ (Here, $\pdata$ is a toy distribution in dimension $d=2$ represented by a chess board pattern.)}
\end{figure}

\begin{examplebox}[Gaussian Conditional Probability Path]
\label{example:gaussian_path}
    One particularly popular probability path is the \themebf{Gaussian probability path}. This is the \textbf{probability path used by most state-of-the-art models}. Let $\alpha_t,\beta_t$ be \themebf{noise schedulers}: two continuously differentiable, monotonic functions with $\alpha_0=\beta_1=0$ and $\alpha_1=\beta_0=1$. We then define the conditional probability path
\begin{align}
\label{eq:gaussian_conditional_probability_paths}
p_t(\cdot|\dap) &= \mathcal{N}(\alpha_t \dap,\beta_t^2 I_d)
& \blacktriangleright \,\, \text{Gaussian conditional path}
\end{align}
which, by the conditions we imposed on $\alpha_t$ and $\beta_t$, fulfills
\begin{align*}
    p_0(\cdot|\dap) &= \mathcal{N}(\alpha_0 \dap,\beta_0^2 I_d) = \mathcal{N}(0,I_d),\quad \text{and}\quad
p_1(\cdot|\dap) = \mathcal{N}(\alpha_1 \dap,\beta_1^2 I_d) = \delta_{\dap},
\end{align*}
where we have used the fact that a normal distribution with zero variance and mean $\dap$ is just $\delta_{\dap}$. Therefore, this choice of $p_t(x|\dap)$ fulfills \cref{eq:interpolating_condition_conditional_path} for $\pinit=\mathcal{N}(0,I_d)$ and is therefore a valid conditional interpolating path.
% The Gaussian conditional probability path has several useful properties which makes it especially amenable to our goals, and because of this we will use it as our prototypical example of a conditional probability path for the rest of the section. 
In \cref{fig:noising_image}, we illustrate its application to an image. We can express sampling from the marginal path $p_t$ as: 
\begin{align}
    \label{eq:gaussian_prob_path_sampling}
    z\sim&\pdata,\,\epsilon\sim\pinit = \mathcal{N}(0,I_d) \,\Rightarrow\, x=\alpha_tz+\beta_t \epsilon\sim p_t \quad &\blacktriangleright \,\,\text{sampling from marginal Gaussian path}
\end{align}
Intuitively, the above procedure adds more noise for lower $t$ until time $t=0$, at which point there is only noise. In \cref{fig:cond_marginal_path_histograms}, we plot an example of such an interpolating path.
\end{examplebox}

\subsection{Conditional and Marginal Vector Fields}

A probability path $(p_t)_{0\leq t\leq 1}$ specifies what distributions $X_t\sim p_t$ the points $X_t$ along a trajectory \emph{should} have. At this point, this is just what we ``wish'' to be the case. But how can we find a vector field such that the trajectories $X_t$ follow the probability path? Flow matching explicitly constructs such a vector field - the ``marginal vector field'' - which we explain in this section.

For every data point $z\in \mathbb{R}^d$, let $\uref_t(\cdot|z)$ denote a \themebf{conditional vector field}. This can be any vector field such that corresponding ODE yields the conditional probability path $p_t(\cdot|z)$, i.e. such that it holds
\begin{align}
\label{eq:cond_ref_ode_conds}
X_0&\sim\pinit,\quad \frac{\dd}{\dd t}X_t =\uref_t(X_t|z)\quad \Rightarrow \quad X_t\sim p_t(\cdot|z)\quad (0\leq t\leq 1).
\end{align}
We can often find a conditional vector field $\uref_t(\cdot|z)$  analytically by hand (i.e. by just doing some algebra ourselves). We illustrate this by deriving a conditional vector field $u_t(x|z)$ for our running example of a Gaussian probability path in Example \ref{example:cond_vf_Gaussian_prob_path}.

At first sight, a conditional vector field seems useless because all endpoints of the ODE $X_1$ will collapse to $X_1=z$, i.e. we are just re-generating known data points $z$. However, the conditional vector field serves as a building block for a vector field that generates actual samples from $\pdata$:
\begin{theorem}[Marginalization trick]
\label{thm:marginalization_trick}
Let $\uref_t(x|z)$ be a conditional vector field (\cref{eq:cond_ref_ode_conds}). Then the \themebf{marginal vector field} $\uref_t(x)$ defined as
\begin{align}
    \label{eq:marginal_vector_field}
    \uref_t(x) = \int \uref_t(x|z)\frac{p_t(x|z)\pdata(z)}{p_t(x)}\dd z,
\end{align}
follows the marginal probability path, i.e. 
\begin{align}
\label{eq:marginal_ode_follows_marginal_path}
    X_0&\sim\pinit,\quad \frac{\dd}{\dd t}X_t =\uref_t(X_t)\quad \Rightarrow \quad X_t\sim p_t\quad (0\leq t\leq 1).
\end{align}
\textbf{In particular, $X_1\sim \pdata$ for this ODE, so that we might say "$\uref_t$ converts noise $\pinit$ into data $\pdata$"}.
\end{theorem}

% The marginalization trick from \cref{thm:marginalization_trick} then allows us to construct the marginal vector field from that. All that is left to do is to learn the marginal vector field, which is what flow matching does - 

% while the marginal vector field $\uref_t(\cdot)$ is generally intractable, we can often find a conditional vector field $\uref_t(\cdot|z)$ satisfying \cref{eq:cond_ref_ode_conds} analytically by hand (i.e. by just doing some algebra ourselves). The marginalization trick from \cref{thm:marginalization_trick} then allows us to relate the conditional and marginal vector fields. As it turns out, one may therefore substitute the conditional vector field $\uref_t(\cdot | z)$ in place of the marginal vector field $\uref_t(\cdot)$. We will make this intuition precise in the next section, when we discuss \themebf{flow matching}. 
\begin{figure}[t!]
\centering
   \begin{subfigure}[b]{\textwidth}
   \centering
   \includegraphics[width=\textwidth]{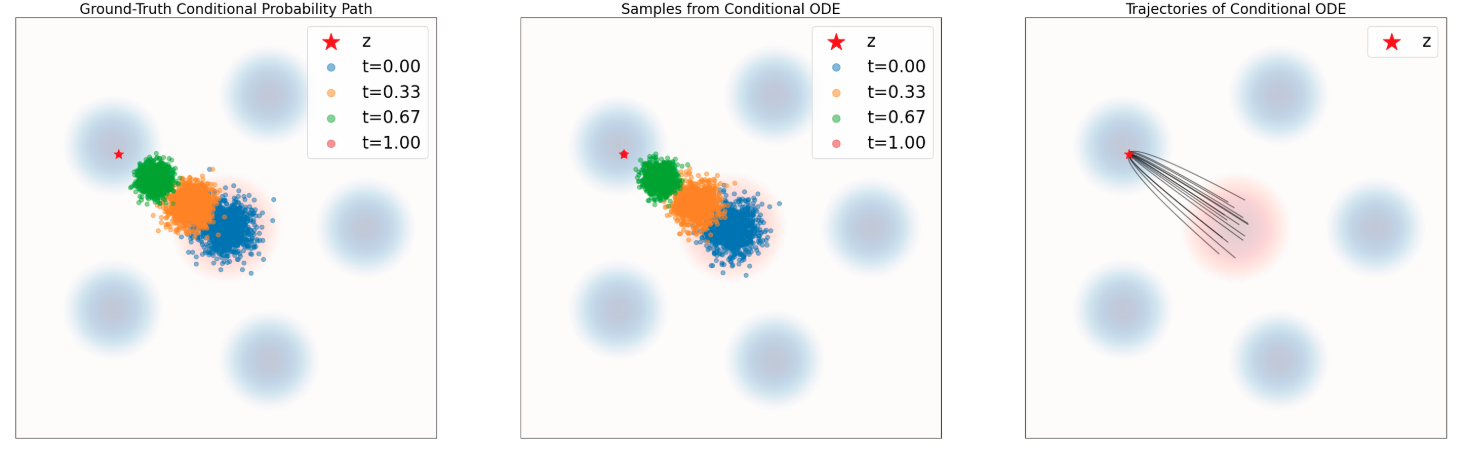}
\end{subfigure}
\begin{subfigure}[b]{\textwidth}
    \centering
   \includegraphics[width=\textwidth]{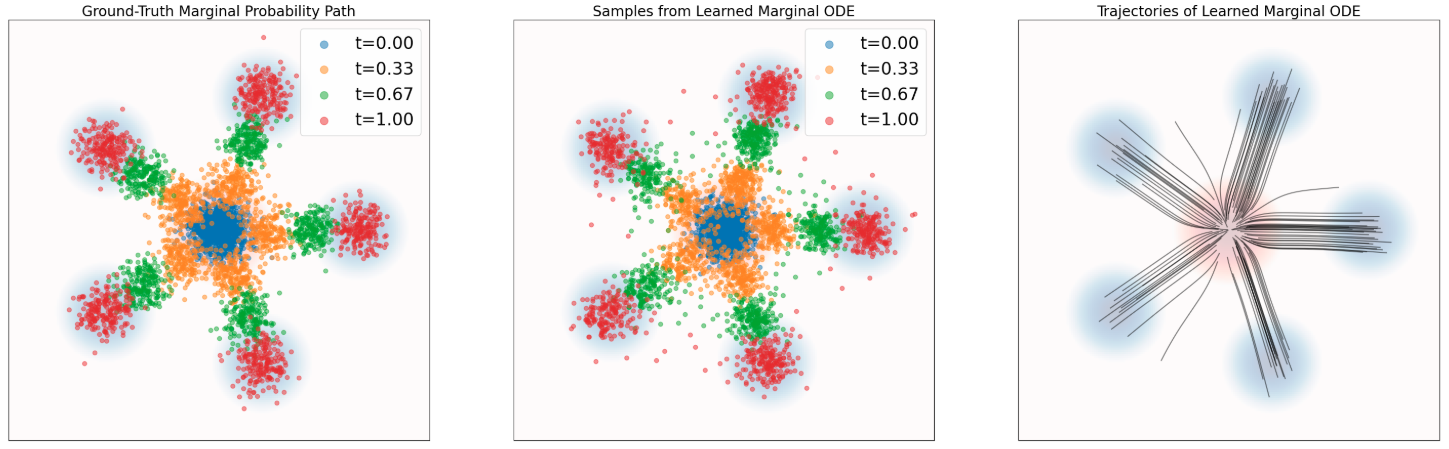}
\end{subfigure}
\caption{\label{fig:cond_marginal_path_simulation}Illustration of \cref{thm:marginalization_trick}. Simulating a  probability path with ODEs. Data distribution $\pdata$ in blue background. Gaussian $\pinit$ in red background. Top row: Conditional probability path. Left: Ground truth samples from conditional path $p_t(\cdot|z)$. Middle: ODE samples over time. Right: Trajectories by simulating ODE with $\uref_t(x|z)$ in \cref{eq:conditional_gaussian_vf}. Bottom row: Simulating a marginal probability path. Left: Ground truth samples from $p_t$. Middle: ODE samples over time. Right: Trajectories by simulating ODE with marginal vector field $\uflow_t(x)$. As one can see, the conditional vector field follows the conditional probability path and the marginal vector field follows the marginal probability path.}
\end{figure}
\begin{examplebox}
[Target ODE for Gaussian probability paths]
\label{example:cond_vf_Gaussian_prob_path}
As before, let $p_t(\cdot|\dap) = \mathcal{N}(\alpha_t \dap,\beta_t^2 I_d)$ for noise schedulers $\alpha_t,\beta_t$ (see \cref{eq:gaussian_conditional_probability_paths}). Let $\dot{\alpha}_t=\partial_t\alpha_t$ and $\dot{\beta}_t=\partial_t\beta_t$ denote respective time derivatives of $\alpha_t$ and $\beta_t$. Here, we want to show that the \themebf{conditional Gaussian vector field} given by
\begin{align}
\label{eq:conditional_gaussian_vf}
    \uref_t(x|z) = \left(\dot{\alpha}_t-\frac{\dot{\beta}_t}{\beta_t}\alpha_t\right)z+\frac{\dot{\beta}_t}{\beta_t}x
\end{align}
is a valid conditional vector field model in the sense of \cref{thm:marginalization_trick}: its ODE trajectories $X_t$ satisfy $X_t\sim p_t(\cdot|z)=\mathcal{N}(\alpha_t z,\beta_t^2I_d)$ if $X_0\sim \mathcal{N}(0,I_d)$. In \cref{fig:cond_marginal_path_simulation}, we confirm this visually by comparing samples from the conditional probability path (ground truth) to samples from simulated ODE trajectories of this flow. As you can see, the distribution match. We will now prove this.

\begin{proof}
Let us construct a conditional flow model $\psiref_t(x|z)$ first by defining
\begin{align}
    \label{eq:cond_flow_gaussian}
    \psiref_t(x|z) = \alpha_t z + \beta_t x.
\end{align}
If $X_t$ is the ODE trajectory of $\psiref_t(\cdot|z)$ with $X_0\sim\pinit = \mathcal{N}(0,I_d)$, then by definition
\begin{align*}
    X_t = \psiref_t(X_0|z) = \alpha_t z + \beta_t X_0 \sim \mathcal{N}(\alpha_t z,\beta^2 I_d) = p_t(\cdot|z).
\end{align*}
We conclude that the trajectories are distributed like the conditional probability path (i.e, \cref{eq:cond_ref_ode_conds} is fulfilled). It remains to extract the vector field $\uref_t(x|z)$ from $\psiref_t(x|z)$. By the definition of a flow (\cref{e:flow_flow}), it holds
\begin{align*}
      \frac{\dd}{\dd t}\psiref_{t}(x|z) &= \uref_t(\psiref_t(x|z)|z)\quad \text{ for all }x,z\in\mathbb{R}^d\\
    \overset{(i)}{\Leftrightarrow} \quad \dot{\alpha}_tz+\dot{\beta}_tx &= \uref_t(\alpha_tz+\beta_t x|z)\quad \text{ for all }x,z\in\mathbb{R}^d\\
    \overset{(ii)}{\Leftrightarrow} \quad \dot{\alpha}_tz+\dot{\beta}_t\left(\frac{x-\alpha_tz}{\beta_t}\right)&= \uref_t(x|z)\quad \text{ for all }x,z\in\mathbb{R}^d\\
    \overset{(iii)}{\Leftrightarrow} \quad\left(\dot{\alpha}_t-\frac{\dot{\beta}_t}{\beta_t}\alpha_t\right)z+\frac{\dot{\beta}_t}{\beta_t}x &= \uref_t(x|z)\quad \text{ for all }x,z\in\mathbb{R}^d
\end{align*}
where in $(i)$ we used the definition of $\psiref_t(x|z)$ (\cref{eq:cond_flow_gaussian}), in $(ii)$ we reparameterized $x\rightarrow (x-\alpha_t z)/\beta_t$, and in $(iii)$ we just did some algebra. Note that the last equation is the conditional Gaussian vector field as we defined in \cref{eq:conditional_gaussian_vf}. This proves the statement.\footnote{One can also double check this by plugging it into the continuity equation introduced later in this section.}
\end{proof}
\end{examplebox}

See \cref{fig:cond_marginal_path_simulation} for an illustration of  \cref{thm:marginalization_trick}. Let's gain some intuition for the marginal vector field. \textbf{Bayes' rule} from statistics says that the following term describes a posterior distribution
\begin{align*}
\frac{p_t(x|z)\pdata(z)}{p_t(x)} = \text{"posterior over data points }z\text{ given noisy data }x\text{"}
\end{align*}
where $\pdata(z)$ is the prior distribution. The marginal vector field then is simply a average: for every \emph{possible} data point $z$ it takes the velocity $u_t(x|z)$ - i.e. the direction that would bring us to $z$ - and then weighs this velocity by how much we believe that $x$ comes from $z$. Averaging over all data points, we obtain the marginal vector field. 

The remainder of this section will make this intuition rigorous and prove \cref{thm:marginalization_trick}. As the main mathematical tool, we will use the \themebf{continuity equation}, a fundamental equation in mathematics and physics. Define the \themebf{divergence} operator $\divv$ as
\begin{align}
\label{eq:divergence_laplacian_definition}
    \divv(v_t)(x)=&\sum\limits_{i=1}^{d}\frac{\partial}{\partial x_i}v_t^i(x)
\end{align}
where $v_t^i$ is the $i$-th coordinate of $v_t$.
\begin{theorem}[Continuity Equation]
\label{thm:continuity_equation}
Let us consider an flow model with vector field $\uref_t$ with $X_0\sim\pinit=p_0$. Then $X_t\sim p_t$ for all $0\leq t\leq 1$ if and only if
\begin{align}
\label{e:continuity_equation}
\partial_t p_t(x)=-\divv (p_t\uref_t)(x)\quad \text{ for all }x\in\R^d, 0\leq t\leq 1,
\end{align}
where $\partial_tp_t(x)= \frac{\dd}{\dd t}p_t(x)$ denotes the time-derivative of $p_t(x)$. Equation \ref{e:continuity_equation} is known as the \themebf{continuity equation}.
\end{theorem}
For the mathematically-inclined reader, we present a self-contained proof of the Continuity Equation in \cref{subsec:proof_fokker_planck}. Before we move on, let us try and understand intuitively the continuity equation. The left-hand side $\partial_tp_t(x)$ describes how much the probability $p_t(x)$ at $x$ changes over time. Intuitively, the change should correspond to the net inflow of probability mass. For a flow model, a particle $X_t$ follows along the vector field $\uref_t$. As you might recall from physics, the divergence measures a sort of net outflow from the vector field. Therefore, the negative divergence measures the net inflow. Scaling this by the total probability mass currently residing at $x$, we get that the net $-\divv (p_tu_t)$ measures the total inflow of probability mass. Since probability mass is conserved (always integrates to 1), the left-hand and right-hand side of the equation should be the same! We now proceed with a proof of the marginalization trick from \cref{thm:marginalization_trick}.
\begin{proof}[Proof of \cref{thm:marginalization_trick}.]
By \cref{thm:continuity_equation}, we have to show that the marginal vector field $\uref_t$, as defined as in \cref{eq:marginal_vector_field}, satisfies the continuity equation. We can do this by direct calculation:
\begin{align*}
\partial_t p_t(x) \overset{(i)}
{=} \partial_t\int p_t(x|\dap) \pdata (z) \dd z
&= \int \partial_t p_t(x|\dap) \pdata (z) \dd z\\
&\overset{(ii)}{=} \int -\divv (p_t(\cdot|z)\uref_t(\cdot|z))(x) \pdata (z) \dd z\\
&\overset{(iii)}{=} -\divv \left(\int p_t(x|z) \uref_t(x|z)\pdata(z) \dd z\right)\\
&\overset{(iv)}{=} -\divv \left(p_t(x)\int \uref_t(x|z) \frac{p_t(x|z)\pdata(z)}{p_t(x)}\dd z\right)(x)\\
&\overset{(v)}{=} -\divv \left(p_t\uref_t\right)(x),
\end{align*}
where in $(i)$ we used the definition of $p_t(x)$ in \cref{eq:marginal_prob_path}, in $(ii)$ we used the continuity equation for the conditional probability path $p_t(\cdot|z)$, in $(iii)$ we swapped the integral and divergence operator using \cref{eq:divergence_laplacian_definition}, in $(iv)$ we multiplied and divided by $p_t(x)$, and in $(v)$ we used \cref{eq:marginal_vector_field}. The beginning and end of the above chain of equations show that the continuity equation is fulfilled for $\uref_t$. By \cref{thm:continuity_equation}, this is enough to imply \cref{eq:marginal_ode_follows_marginal_path}, and we are done.
\end{proof}

\subsection{Learning the Marginal Vector Field}

Now, we are ready to describe the training algorithm. The goal of flow matching is to train the neural network $u_t^\theta$ such that it equals the marginal vector field $\uref_t$. If this holds, we know that the endpoints $X_1\sim \pdata$ have the desired distribution by \cref{thm:marginalization_trick}. In the following, we denote by $\text{Unif}=\text{Unif}_{[0,1]}$ the uniform distribution on the interval $[0,1]$, and by $\mathbb{E}$ the expected value of a random variable. An intuitive way of obtaining $u_t^\theta\approx \uref_t$ is to use a mean-squared error, i.e. to use the \themebf{flow matching loss} defined as
\label{subsec:training_algorithm}
\begin{align}
    \label{eq:marginal_loss_function}
\Lmarg(\theta)&=\mathbb{E}_{t\sim\text{Unif},x\sim p_t}[\|u_t^\theta(x) - \uref_t(x)\|^2]\\&\overset{(i)}{=} \mathbb{E}_{t\sim\text{Unif},z\sim \pdata, x\sim p_t(\cdot|z)}[\|u_t^\theta(x) - \uref_t(x)\|^2],
\end{align}
where $p_t(x)=\int p_t(x|z)\pdata(z) \dd z$ is the marginal probability path and in $(i)$ we used the sampling procedure given by \cref{eq:marginal_prob_path}. Intuitively, this loss says: First, draw a random time $t \in [0,1]$. Second, draw a random point $z$ from our data set, sample from $p_t(\cdot|z)$ (e.g., by adding some noise), and compute $u_t^\theta(x)$. Finally, compute the mean-squared error between the output of our neural network and the marginal vector field $\uref_t(x)$. Unfortunately, we are \textit{not} done here. While we do know the formula for $\uref_t$ by \cref{thm:marginalization_trick}, we cannot compute it efficiently as the integral is intractable. Instead, we will exploit the fact that the \themebf{conditional} velocity field $\uref_t(x|z)$ is tractable. % We could try to approximate the above integral by screening over the whole dataset. However, even for a dataset of 1 million images (which is very small in modern days), we would need to loop over 1 million images every time we evaluate $\uref_t(x)$ - an extremely computationally expensive approach. However, it turns out that we can solve this. 
To do so, let us define the \themebf{conditional flow matching loss} 
\begin{align}
\label{eq:cfm}
\Lcond(\theta) =     \mathbb{E}_{t\sim \Unif, z\sim \pdata, x\sim p_t(\cdot|\dap)}[\|u_t^\theta(x) - \uref_t(x|\dap)\|^2].
\end{align}
Note the difference to     \cref{eq:marginal_loss_function}: we use the conditional vector field $\uref_t(x|z)$ instead of the marginal vector $\uref_t(x)$. As we have an analytical formula for $\uref_t(x|z)$, we can minimize the above loss easily. But wait, what sense does it make to regress against the conditional vector field if it's the marginal vector field we care about? As it turns out, by \themeit{explicitly} regressing against the tractable, conditional vector field, we are \themeit{implicitly} regressing against the intractable, marginal vector field. The next result makes this intuition precise.

\begin{figure}[!t]
    \centering
    \begin{tabular}{ccc}
         \includegraphics[width=\textwidth]{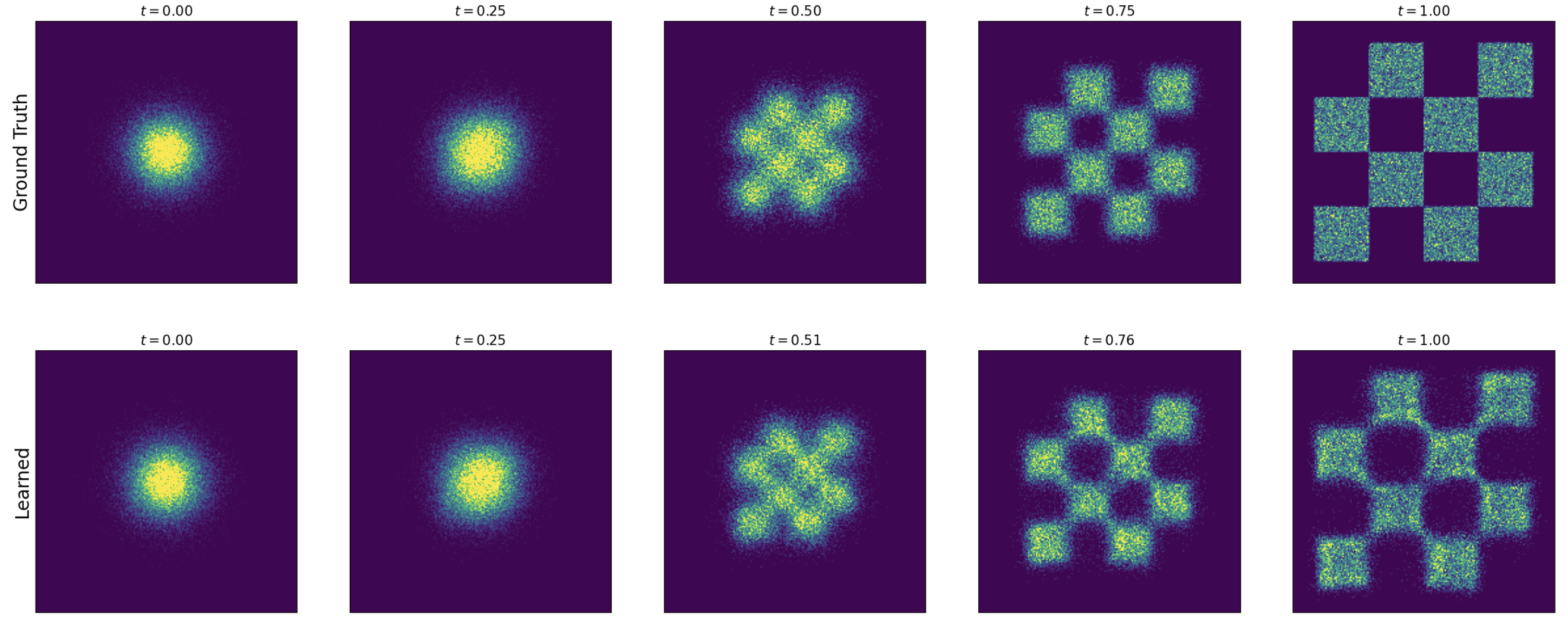} &
         % \includegraphics[width=0.22\textwidth]{assets/flow_velocity/flow_v_5.png} &
         % \includegraphics[width=0.3\textwidth]{fm_guide_assets/flow_10.png} &
         % % \includegraphics[width=0.22\textwidth]{assets/flow_velocity/flow_v_14.png} &
         % \includegraphics[width=0.3\textwidth]{fm_guide_assets/flow_16.png} 
    \end{tabular}
    \caption{\label{fig:fm_illustration_checkerboard}Illustration of \cref{thm:fm_loss} with a Gaussian CondOT probability path: simulating an ODE from a trained flow matching model. The data distribution is the chess board pattern (top right). Top row: Histogram from ground truth marginal probability path $p_t(x)$. Bottom row: Histogram of samples from flow matching model. As one can see, the top row and bottom row match after training (up to training error). The model was trained using \cref{alg:training_fm_basic}.}
\end{figure}

\begin{theorem}
\label{thm:fm_loss} 
The marginal flow matching loss equals the conditional flow matching loss up to a constant. That is,
\begin{align*}
\Lmarg(\theta) = \Lcond(\theta) + C,
\end{align*}
where $C$ is independent of $\theta$. Therefore, their gradients coincide:
\begin{align*}
\nabla_\theta \Lmarg(\theta) = \nabla_\theta \Lcond(\theta).
\end{align*}
Hence, minimizing $\Lcond(\theta)$ with e.g., stochastic gradient descent (SGD) is equivalent to minimizing $\Lmarg(\theta)$ in the same fashion. In particular, \textbf{for the minimizer $\theta^*$ of $\Lcond(\theta)$, it will hold that $u_t^{\theta^*}=\uref_t$, i.e. the neural network will equal the marginal vector field} (assuming an infinitely expressive parameterization). 
\end{theorem}
\begin{proof}[Direct Proof]
The proof works by expanding the mean-squared error into three components and removing constants:
\begin{align*}
    \Lmarg(\theta)&\overset{(i)}{=}\mathbb{E}_{t\sim\text{Unif},x\sim p_t}[\|u_t^\theta(x) - \uref_t(x)\|^2]\\
    &\overset{(ii)}{=}\mathbb{E}_{t\sim\text{Unif},x\sim p_t}[\|u_t^\theta(x)\|^2 - 2u_t^\theta(x)^T\uref_t(x) + \|\uref_t(x)\|^2]\\
    &\overset{(iii)}{=}\mathbb{E}_{t\sim\text{Unif},x\sim p_t}\left[\|u_t^\theta(x)\|^2\right] - 2\mathbb{E}_{t\sim\text{Unif},x\sim p_t}[u_t^\theta(x)^T\uref_t(x)] + \underbrace{\mathbb{E}_{t\sim\text{Unif}_{[0,1]}, x\sim p_t}[\|\uref_t(x)\|^2]}_{=:C_1}\\
    &\overset{(iv)}{=}\mathbb{E}_{t\sim\text{Unif},z\sim\pdata, x\sim p_t(\cdot|z)}[\|u_t^\theta(x)\|^2] - 2\mathbb{E}_{t\sim\text{Unif},x\sim p_t}[u_t^\theta(x)^T\uref_t(x)] + C_1
\end{align*}
where $(i)$ holds by definition, in $(ii)$ we used the formula $\|a-b\|^2=\|a\|^2-2a^Tb+\|b\|^2$, in $(iii)$ we define a constant $C_1$ and in $(iv)$ we used the sampling procedure of $p_t$ given by \cref{eq:marginal_prob_path}. Let us reexpress the second summand:
\begin{align*}
\mathbb{E}_{t\sim\text{Unif},x\sim p_t}[u_t^\theta(x)^T\uref_t(x)]&\overset{(i)}{=}\int\limits_{0}^{1}\int p_t(x)u_t^\theta(x)^T\uref_t(x)\,\dd x\, \dd t\\
&\overset{(ii)}{=}\int\limits_{0}^{1}\int p_t(x)u_t^\theta(x)^T\left[\int \uref_t(x|z)\frac{ p_t(x|z)\pdata(z)}{p_t(x)}\dd z\right]\dd x\, \dd t\\
&\overset{(iii)}{=}\int\limits_{0}^{1}\int\int u_t^\theta(x)^T\uref_t(x|z) p_t(x|z)\pdata(z)\,\dd z\,\dd x\, \dd t\\
&\overset{(iv)}{=}\mathbb{E}_{t\sim\text{Unif},z\sim \pdata, x\sim p_t(\cdot|z)}[u_t^\theta(x)^T\uref_t(x|z)]
\end{align*}
where in $(i)$ we expressed the expected value as an integral, in $(ii)$ we use \cref{eq:marginal_vector_field}, in $(iii)$ we use the fact that integrals are linear, in $(iv)$ we express the integral as an expected value. Note that this was really the crucial step of the proof: The beginning of the equality used the marginal vector field $\uref_t(x)$, while the end uses the conditional vector field $\uref_t(x|z)$. We plug is into the equation for $\Lmarg$ to get:
\begin{align*}
\Lmarg(\theta)&\overset{(i)}{=}\mathbb{E}_{t\sim\text{Unif},z\sim \pdata, x\sim p_t(\cdot|z)}[\|u_t^\theta(x)\|^2]-2\mathbb{E}_{t\sim\text{Unif},z\sim \pdata, x\sim p_t(\cdot|z)}[u_t^\theta(x)^T\uref_t(x|z)] +C_1\\
&\overset{(ii)}{=}\mathbb{E}_{t\sim\text{Unif},z\sim \pdata, x\sim p_t(\cdot|z)}[\|u_t^\theta(x)\|^2-2u_t^\theta(x)^T\uref_t(x|z)+\|\uref_t(x|z)\|^2-\|\uref_t(x|z)\|^2] +C_1\\
&\overset{(iii)}{=}\mathbb{E}_{t\sim\text{Unif},z\sim \pdata, x\sim p_t(\cdot|z)}[\|u_t^\theta(x)-\uref_t(x|z)\|^2]+\underbrace{\mathbb{E}_{t\sim\text{Unif},z\sim \pdata, x\sim p_t(\cdot|z)}[-\|\uref_t(x|z)\|^2]}_{C_2} +C_1\\
&\overset{(iv)}{=}\Lcond(\theta) + \underbrace{C_2+C_1}_{=:C}
\end{align*}
where in $(i)$ we plugged in the derived equation, in $(ii)$ we added and subtracted the same value,  in $(iii)$ we used the formula $\|a-b\|^2=\|a\|^2-2a^Tb+\|b\|^2$ again, and in $(iv)$ we defined a constant in $\theta$. This finishes the proof.
\end{proof}
\begin{algorithm}[ht]
\caption{Flow Matching Training Procedure (for Gaussian CondOT path $p_t(x|z)=\mathcal{N}(tz,(1-t)^2)$)}
\label{alg:training_fm_basic}
\begin{algorithmic}[1]
\REQUIRE A dataset of samples $z\sim \pdata$, neural network $u_t^\theta$
\FOR{each mini-batch of data}
    \STATE Sample a data example $\dap$ from the dataset.
    \STATE Sample a random time $t \sim \text{Unif}_{[0,1]}$.
    \STATE Sample noise $\epsilon\sim\mathcal{N}(0,I_d)$
    \STATE Set 
    \begin{align*}
        x&=t z + (1-t)\epsilon \quad &(\text{General case: }x\sim p_t(\cdot\mid z))
    \end{align*}
    
    %\IF{Flow matching}
    \STATE Compute loss
    \begin{align*}
        \mathcal{L}(\theta) =& \|u_t^\theta(x)-(z-\epsilon)\|^2 \quad &(\text{General case: }=\|u_t^\theta(x)-\uref_t(x|z)\|^2)
    \end{align*}
    %\ENDIF
    % \IF{Score matching}
    % \STATE
    % \begin{align*}
    %     \mathcal{L}(\theta) =& \|s_t^\theta(x_t)+\frac{\epsilon}{\beta_t}\|^2 \quad &(\text{General case: }=\|s_t^\theta(x_t)+\nabla\log p_t(x_t|z)\|^2)
    % \end{align*}
    % \ENDIF
    \STATE Update $\theta \gets \text{grad\_update}(\mathcal{L}(\theta))$.
    % \STATE Update the model parameters $\theta$ via gradient descent on $\mathcal{L}(\theta)$.
\ENDFOR
\end{algorithmic}
\end{algorithm}
Therefore, \textbf{flow matching training consists of minimizing the conditional flow matching loss.} The training procedure is summarized in \cref{alg:training_fm_basic} and visualized in \cref{fig:fm_illustration_checkerboard}. Note that there are several striking features about this algorithm: First, we never actually simulate any ODE during training. People call this feature of the algorithm \textbf{simulation-free}. This makes training extremely cheap as you don't have to roll out trajectories of the ODE during training (which takes a lot of steps). Second, the training is a simple regression objective - we are just regressing against $\uref_t(x|z)$. So it is not too different from supervised learning after all. Finally, the algorithm is extremely simple - it is hard to think of a much simpler training objective. All of this makes flow matching an extremely appealing method for large-scale machine learning models. Once $u_t^{\theta}$ has been trained, we may simulate the flow model
\begin{align}
    \dd X_t = u_t^\theta(X_t)\, \dd t,\quad\quad X_0\sim\pinit
\end{align}
via e.g., \cref{alg:sampling_flow_model} to obtain samples $X_1\sim \pdata$. This whole pipeline is called \themebf{flow matching} in the literature \citep{lipman2022flow, liu2022flow, albergo2023stochastic, lipman2024flow}.
Let us now instantiate the conditional flow matching loss for Gaussian probability paths:
\begin{examplebox}[Flow Matching for Gaussian Conditional Probability Paths]
Let us return to the example of Gaussian probability paths $p_t(\cdot|z)=\mathcal{N}(\alpha_t z; \beta_t^2 I_d)$, where we may sample from the conditional path via
\begin{align}
\label{eq:sampling_procedure_gaussian_path_restated_for_fm}
    \epsilon\sim\mathcal{N}(0,I_d)\quad 
    \Rightarrow\quad x_t = \alpha_t z + \beta_t \epsilon \sim \mathcal{N}(\alpha_tz,\beta_t^2I_d)=p_t(\cdot|z).
\end{align}
As we derived in \cref{eq:conditional_gaussian_vf}, the conditional vector field $\uref_t(x|z)$ is given by
\begin{align}
\label{e:marginal_vf_cond_score}
    \uref_t(x|z) =& \left(\dot{\alpha}_t-\frac{\dot{\beta}_t}{\beta_t}\alpha_t\right)z+\frac{\dot{\beta}_t}{\beta_t}x,
\end{align}
where $\dot{\alpha}_t=\partial_t\alpha_t$ and $\dot{\beta}_t=\partial_t\beta_t$ are the respective time derivatives. Plugging in this formula, the conditional flow matching loss reads
\begin{align}
\Lcond(\theta) &= \mathbb{E}_{t\sim \text{Unif},z\sim \pdata, x\sim \mathcal{N}(\alpha_tz,\beta_t^2I_d)}[\lVert u_t^\theta(x)-\left(\dot{\alpha}_t-\frac{\dot{\beta}_t}{\beta_t}\alpha_t\right)z-\frac{\dot{\beta}_t}{\beta_t}x\rVert^2]\\&\overset{(i)}{=}\mathbb{E}_{t\sim\Unif,z\sim \pdata, \epsilon\sim \mathcal{N}(0,I_d)}[\|u_t^\theta(\alpha_tz+\beta_t\epsilon)-(\dot{\alpha}_tz+\dot{\beta}_t\epsilon)\|^2]
\end{align}
where in $(i)$ we plugged in \cref{eq:sampling_procedure_gaussian_path_restated_for_fm} and replaced $x$ by $\alpha_tz+\beta_t\epsilon$. Note the simplicity of $\Lcond$ : We sample a data point $z$, sample some noise $\epsilon$ and then we take a mean squared error. Let us make this even more concrete for the special case of $\alpha_t=t$, and $\beta_t=1-t$. The corresponding probability $p_t(x|z)=\mathcal{N}(tz,(1-t)^2)$ is sometimes referred to as the (Gaussian) \themebf{CondOT probability path}. Then we have $\dot{\alpha}_t=1,\dot{\beta}_t=-1$, so that
\begin{align*}
        \mathcal{L}_{\text{cfm}}(\theta)=&\mathbb{E}_{t\sim\Unif,z\sim \pdata, \epsilon\sim \mathcal{N}(0,I_d)}[\|u_t^\theta(tz+(1-t)\epsilon)-(z-\epsilon)\|^2]
\end{align*}
Many famous state-of-the-art models have been trained using this simple yet effective procedure, e.g. \themeit{Stable Diffusion 3}, Meta's \themeit{Movie Gen Video}, and probably many more proprietary models. In \cref{fig:fm_illustration_checkerboard}, we visualize it in a simple example and in \cref{alg:training_fm_basic} we summarize the training procedure.
\end{examplebox}

% When I first saw \cref{thm:fm_loss}, I found it a big magical: Without seeing $\uref_t(x)$ explicitly ever, we can approximate it implicitly with a neural network. Therefore, \cref{thm:fm_loss} gives us a powerful of training our ODE generative model by minimizing $\Lcond(\theta)$. All we have to do is minimize a simple mean squared error. This is scalable to very large datasets in high dimensions and can be implemented in a few lines of code. When tracking the training, you should keep in mind that \textbf{the absolute value of the loss is meaningless} (you will never achieve zero loss because $C<0$). This makes it sometimes a little tricky to realize when the training is converged.

Let us summarize the results of this section.
\begin{summarybox}[Flow Matching]
\textbf{Flow matching training consists of learning the marginal vector field $\uref_t$.} To construct it, we choose a \themebf{conditional probability path} $p_t(x|\dap)$ that fulfils $p_0(\cdot|z)=\pinit$, $p_1(\cdot|z)=\delta_{z}$. Next, we find a \themebf{conditional vector field} $\uref_t(x|z)$ such that its corresponding flow $\psiref_t(x|z)$ fulfills
\begin{align*}
    X_0\sim \pinit \quad \Rightarrow \quad X_t = \psiref_t(X_0|z) \sim p_t(\cdot|z),
\end{align*}
or, equivalently, that $\uref_t$ satisfies the continuity equation.  Then the \themebf{marginal vector field} defined by
\begin{align}
    \label{eq:marginal_vector_field_restated}
    \uref_t(x) = \int \uref_t(x|z)\frac{p_t(x|z)\pdata(z)}{p_t(x)}\dd z,
\end{align}
follows the marginal probability path, i.e.,
\begin{align}
    \label{eq:marginal_ode_follows_marginal_path_restated}
    X_0&\sim\pinit,\quad \dd X_t =\uref_t(X_t)\dd t\Rightarrow X_t\sim p_t\quad (0\leq t\leq 1).
\end{align}
In particular, $X_1\sim \pdata$ for this ODE, so that $\uref_t$ "converts noise into data", as desired. To learn it, we minimize the \themebf{conditional flow matching loss}
\begin{align}
\Lcond(\theta) =     \mathbb{E}_{t\sim \Unif, z\sim \pdata, x\sim p_t(\cdot|\dap)}[\|u_t^\theta(x) - \uref_t(x|\dap)\|^2].
\end{align}

The most widely used example is the \themebf{Gaussian probability path}. For this case, the formulas become:
\begin{align}
p_t(x|z) =& \mathcal{N}(x;\alpha_t z,\beta_t^2 I_d)\\
\uflow_t(x|z)=&\left(\dot{\alpha}_t-\frac{\dot{\beta}_t}{\beta_t}\alpha_t\right)z+\frac{\dot{\beta}_t}{\beta_t}x\\
\Lcond(\theta) =&\mathbb{E}_{t\sim\Unif,z\sim \pdata, \epsilon\sim \mathcal{N}(0,I_d)}[\|u_t^\theta(\alpha_tz+\beta_t\epsilon)-(\dot{\alpha}_tz+\dot{\beta}_t\epsilon)\|^2]
\end{align}
for \themebf{noise schedulers} $\alpha_t,\beta_t\in\mathbb{R}$, i.e. continuously differentiable, monotonic functions that we choose such that $\alpha_0=\beta_1=0$ $\alpha_1=\beta_0=1$ (e.g. $\alpha_t=t,\beta_t=1-t$).
\end{summarybox}

\newpage
\section{Score Functions and Score Matching}
\label{sec:training_generative_models}
In the last section, we showed how to train a flow model with flow matching. In this section, we discuss diffusion models and demonstrate how to train them using \textbf{score matching}.

\subsection{Conditional and Marginal Score Functions}

\begin{wrapfigure}{r}{0.6\textwidth}
  %\begin{center}
\includegraphics[width=0.6\textwidth]{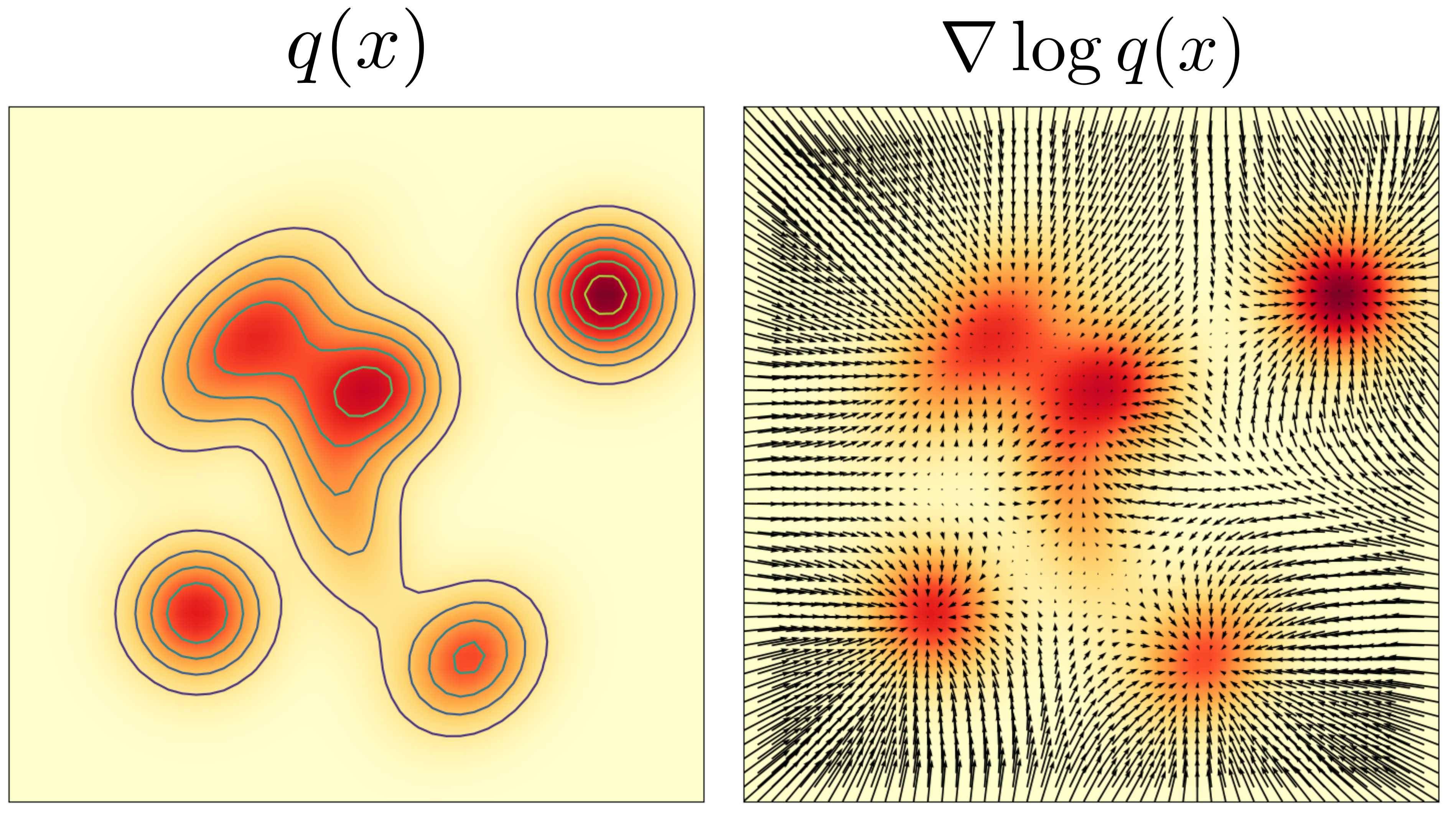}
  %\end{center}
\caption{\label{fig:score_function}Illustration of score function $\nabla\log q(x)$ plotted as black rows (right) of a general probability distribution $q(x)$ (left).}
\end{wrapfigure}
So far, the central object of interest for our investigation was a vector field $u_t(x)$. Diffusion models \citep{song2020score, song2021sde} take a different perspective focused on \textbf{score functions}. Therefore, in this section, we will rephrase what we have learned here in the language of score functions - providing a novel perspective. Let $q(x)$ be an arbitrary probability distribution. Then the \themebf{score function} of $q$ is defined as $\nabla\log q(x)$, i.e. as the gradient of the log-likelihood of $q$ with respect to $x$. The score has an intuitive meaning: $\nabla\log q(x)$ is the direction of steepest ascent with respect to log-likelihood. This is illustrated in \cref{fig:score_function}.

Let us return to the setting of conditional probability paths $p_t(x|z)$ and marginal probability paths $p_t(x)$ as in \cref{sec:flow_matching}. Then we can equivalently define the \themebf{conditional score function} as $\nabla\log p_t(x|z)$ and the \themebf{marginal score function} as $\nabla\log p_t(x)$. Similar to \cref{eq:marginal_vector_field}, the marginal score can be expressed via the conditional score function $\nabla \log p_t(x|z)$ via
\begin{align}
\label{e:marginal_score}
\nabla\log p_t(x)
=\int \nabla \log p_t(x|z)\frac{ p_t(x|z)\pdata(z)}{p_t(x)}\dd z.
\end{align}
Hence, \textbf{the relation between the conditional and marginal score is analogous to the relation between the conditional and marginal vector field}. Note that we can prove \cref{e:marginal_score} via
\begin{align}
\nabla\log p_t(x) = \frac{\nabla p_t(x)}{p_t(x)}
=\frac{\nabla \int p_t(x|z)\pdata(z)\dd z}{p_t(x)}
=\frac{\int \nabla p_t(x|z)\pdata(z)\dd z}{p_t(x)}
=\int \nabla \log p_t(x|z)\frac{ p_t(x|z)\pdata(z)}{p_t(x)}\dd z,
\end{align}
where we have used the rule $\partial_{y} \log y=1/y$ combined with the chain rule twice.

\begin{examplebox}[Score Function for Gaussian Probability Paths.] 
For the Gaussian path $p_t(x|z)=\mathcal{N}(x;\alpha_t z,\beta_t^2 I_d)$, we can use the form of the Gaussian probability density (see \cref{e:gaussian}) to get
\begin{align}
\label{eq:cond_score_gaussian}
    \nabla \log p_t(x|z) = \nabla\log \mathcal{N}(x;\alpha_t z,\beta_t^2 I_d) = -\frac{x-\alpha_t z}{\beta_t^2}.
\end{align}
\end{examplebox}

Note that the score function for a Gaussian probability path is a linear function of $x$ and z. The same is true for the conditional vector field $u_t(x|z)$ (see \cref{eq:conditional_gaussian_vf}). It is thus possible to convert between the two, as the next proposition illustrates.
\begin{proposition}[Conversion Formula for Gaussian Probability Paths]
\label{prop:conversion_formula_gaussian_prob_path}
For the Gaussian probability path $p_t(x|z)=\mathcal{N}(\alpha_t z,\beta_t^2 I_d)$, the conditional (resp. marginal) vector field and the conditional (resp. marginal) score are related by the following identities
\begin{align}
\uref_t(x|z)=&a_t\nabla\log p_t(x|z)+b_tx,\quad a_t=\left(\beta_t^2\frac{\dot{\alpha}_t}{\alpha_t}-\dot{\beta}_t\beta_t\right),\quad b_t=\frac{\dot{\alpha}_t}{\alpha_t} \label{eq:reparam_cond}\\
\uref_t(x)=&a_t\nabla\log p_t(x)+b_tx. \label{eq:reparam_uncond}
\end{align}
In particular, we note that the conditional (resp. marginal) vector field can be recovered from the conditional (resp. marginal) score, and vice versa.
\end{proposition}
\begin{proof}
For the conditional vector field and conditional score, we can derive:
\begin{align*}
\uref_t(x|z)=&\left(\dot{\alpha}_t-\frac{\dot{\beta}_t}{\beta_t}\alpha_t\right)z+\frac{\dot{\beta}_t}{\beta_t}x
\overset{(i)}{=}\left(\beta_t^2\frac{\dot{\alpha}_t}{\alpha_t}-\dot{\beta}_t\beta_t\right)\left(\frac{\alpha_tz-x}{\beta_t^2}\right)+\frac{\dot{\alpha}_t}{\alpha_t}x=\left(\beta_t^2\frac{\dot{\alpha}_t}{\alpha_t}-\dot{\beta}_t\beta_t\right)\nabla\log p_t(x|z)+\frac{\dot{\alpha}_t}{\alpha_t}x
\end{align*}
where in $(i)$ we just did some algebra. By taking integrals, the same identity holds for the marginal flow vector field and the marginal score function:
\begin{align*}
\uref(x)= \int \uref_t(x|z)\frac{ p_t(x|z)\pdata(z)}{p_t(x)}\dd z
=&\int\left[a_t\nabla\log p_t(x|z)+b_tx\right] \frac{ p_t(x|z)\pdata(z)}{p_t(x)}\dd z\\
\overset{(i)}{=}&a_t\nabla\log p_t(x)+b_tx
\end{align*}
where in $(i)$ we used \cref{e:marginal_score} and the fact that posterior density integrates to $1$.
\end{proof}
\Cref{prop:conversion_formula_gaussian_prob_path} is striking because it says that once we've learned $\uref_t$ we've also learned the  score function $\nabla\log p_t(x)$, and vice versa. Therefore, many diffusion models learn the score function $\nabla\log p_t(x)$ instead via a neural network. We will discuss this in \cref{subsec:score_matching}.
\begin{remarkbox}[Reparameterization of the Score]
The reparameterization formula for Gaussian probability paths in \cref{eq:reparam_cond} is possible because both sides (conditional vector field and conditional score) are \emph{linear} functions of $x$ and $z$. Once we marginalize (marginal vector field and marginal score), both sides are just a linear reparameterization of the posterior mean $\EE_{z|x}\left[z\right]$. It follows that any quantity that allows to recover $\EE_{z|x}\left[z\right]$ can in turn be used to recover the unconditional vector field and score. Further, doing so might even be preferable from a numerical/training stability standpoint. One common choice is the posterior mean itself, often referred to as the \textit{denoiser}. Formally, we define the \themebf{conditional and marginal denoiser} as
\begin{align}
\label{e:denoiser}
    D_t(x|z)=z, \quad 
D_t(x)
=\int z\,\frac{ p_t(x|z)\pdata(z)}{p_t(x)}\dd z \overset{(i)}{=} \frac{1}{\dot{\alpha}_t\beta_t-\alpha_t\dot{\beta}_t}(\beta_t \uref_t(x_t)-\dot{\beta}_tx_t).
\end{align}
Here, $(i)$ follows from an equivalent derivation as in \cref{prop:conversion_formula_gaussian_prob_path}. The denoiser has a very intuitive interpretation: it is the expected value of clean data $z$ given noisy data $x$.\footnote{Food for thought: will the denoiser always output a ``clean'' data point? Why or why not, and what might this depend on?} People often call such models \themebf{denoising diffusion models} as learning $D_t$ and learning $\uref_t$ are theoretically equivalent.
\end{remarkbox}

\subsection{Sampling with SDEs}

\begin{figure}[t!]
\centering
   \begin{subfigure}[b]{\textwidth}
   \centering
   \includegraphics[width=\textwidth]{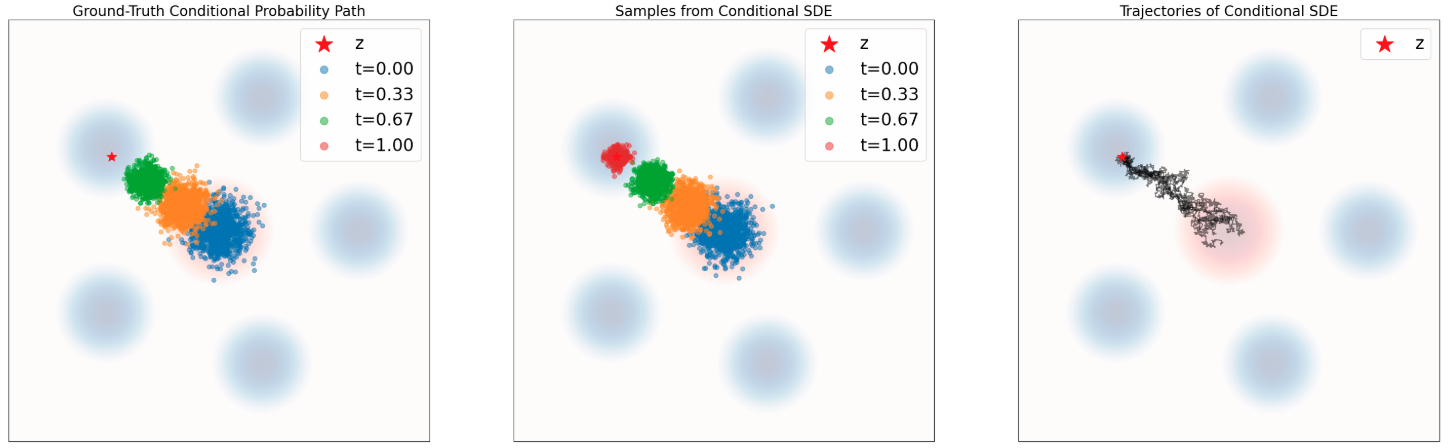}
   \label{fig:Ng1} 
\end{subfigure}
\begin{subfigure}[b]{\textwidth}
    \centering
   \includegraphics[width=\textwidth]{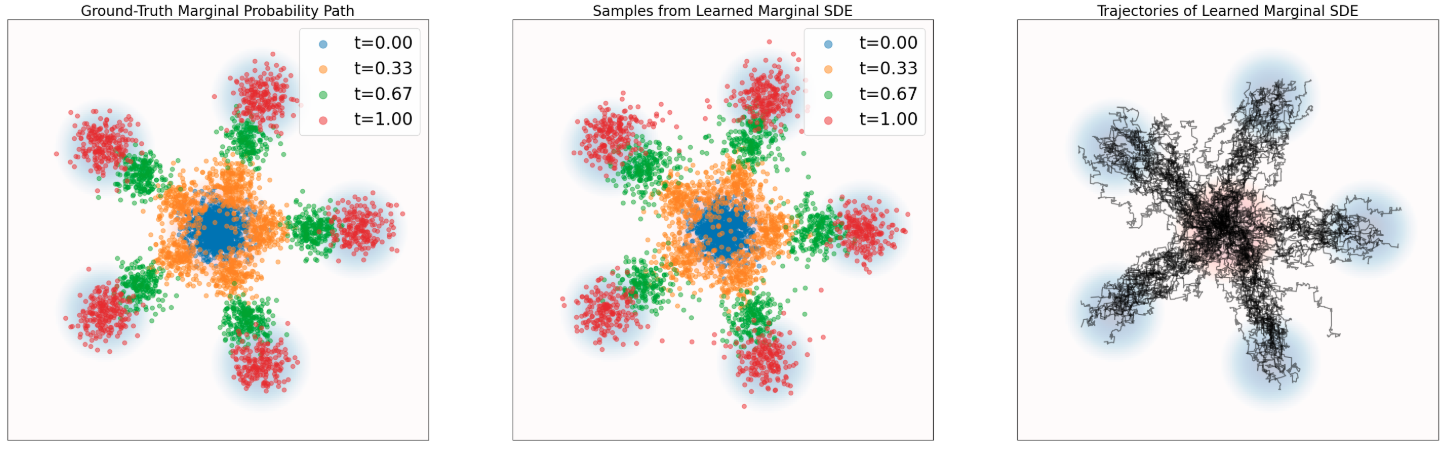}
\end{subfigure}
\caption{\label{fig:thm_sde_extension}Illustration of \cref{thm:langevin_trick}. Simulating a probability path with SDEs. This repeats the plots from \cref{fig:cond_marginal_path_simulation} with SDE sampling using \cref{eq:sde_extension}. Data distribution $\pdata$ in blue background. Gaussian $\pinit$ in red background. Top row: Conditional path. Bottom row: Marginal probability path. As one can see, the SDE transports samples from $\pinit$ into samples from $\delta_{z}$ (for the conditional path) and to $\pdata$ (for the marginal path).}
\end{figure}

So far, we have demonstrated how one can construct a trajectory $X_t$ of an ODE that follows a desired probability path $p_t$ via a marginal vector field $\uref_t$. But this approach is constrained to flow models. What about diffusion models? Using score functions, let us now extend this result to SDEs.
\begin{theorem}[SDE Extension Trick]
\label{thm:langevin_trick}
Define the conditional and marginal vector fields $\uref_t(x|z)$ and $\uref_t(x)$ as before. Then, for any diffusion coefficient $\sigma_t\geq 0$, we may construct an SDE by adding \textcolor{ForestGreen}{stochastic dynamics} to the dynamics of the \textcolor{RoyalBlue}{original ODE} as follows:
\begin{alignat}{2}
\label{eq:sde_extension}
X_0 &\sim \pinit,\qquad & \dd X_t
  &=\textcolor{RoyalBlue}{\uref_t(X_t)\dd t} + \textcolor{ForestGreen}{\frac{\sigma_t^2}{2}\nabla\log p_t(X_t)\dd t
    + \sigma_t\dd W_t} \\
&&
&=\Big[\textcolor{RoyalBlue}{\uref_t(X_t)}+\textcolor{ForestGreen}{\frac{\sigma_t^2}{2}\nabla\log p_t(X_t)}\Big]\dd t
    + \textcolor{ForestGreen}{\sigma_t}\dd W_t \notag \\
\label{eq:marginal_ode_follows_marginal_path_sde}
\Rightarrow\quad X_t &\sim p_t\quad (0\le t\le 1). &&
\end{alignat}

In particular, $X_1\sim \pdata$ for this SDE. We note that the \textcolor{ForestGreen}{stochastic dynamics} are closely related to the \themebf{Langevin dynamics}, and can be thought of as injecting noise while preserving the marginal distribution $p_t$. We discuss Langevin dynamics briefly in \cref{remark:langevin}.
\end{theorem}

We illustrate the dynamics described in \cref{thm:langevin_trick} in \cref{fig:thm_sde_extension}. As one can see, the trajectories are now zig-zagged, illustrating the stochastic nature of the SDE's evolution. As \cref{thm:langevin_trick} establishes however, the marginals $p_t$ stay the same. Note that the above result is striking in that we can choose \emph{any} diffusion coefficient $\sigma_{t}\geq 0$ even after having trained the networks. In theory, \cref{thm:langevin_trick} holds for any choice of $\sigma_t$. However, \emph{in practice}, we suffer from both \textit{training error} (the neural network does not perfectly approximate the marginal vector field and score) and simulation error (e.g. for $\sigma_{t}\gg0$, we would need to take prohibitively small step sizes in \cref{alg:sampling_diffusion_model}). In practice, for a fixed trained model, there is then an optimal $\sigma_{t}\geq 0$ which can be empirically determined \citep{karras2022elucidating, albergo2023stochastic, ma2024sit}.\footnote{Again, we stress that the existence of a ``best $\sigma_t$'' is an artifact of imperfectly trained models and finite compute budgets rather than a theoretical statement about the dynamics in their continuous limit.}

For Gaussian probability paths, we get the score function for free by having learned the marginal vector field. 
\begin{examplebox}[Gaussian SDE Extension Trick]
    By \cref{prop:conversion_formula_gaussian_prob_path}, for Gaussian probability paths, we can express the SDE from \cref{thm:langevin_trick} purely using score functions:
\begin{align}
\label{eq:sde_extension_gaussian}
    X_0\sim&\,\pinit,\quad \dd X_t =\left[\left(a_t+\frac{\sigma_t^2}{2}\right)\nabla\log p_t(X_t)+b_tX_t\right]\dd t +\sigma_t\dd W_t\\
\label{eq:marginal_ode_follows_marginal_path_sde_gaussian}
    \Rightarrow X_t\sim& \,p_t\quad (0\leq t\leq 1)
\end{align}
where $a_t,b_t$ are defined as in \cref{prop:conversion_formula_gaussian_prob_path}.
\end{examplebox}

In the remainder of this section, we will prove  \cref{thm:langevin_trick} via the \themebf{Fokker-Planck equation}, which extends the continuity equation from ODEs to SDEs. To do so, let us first define the \themebf{Laplacian} operator $\Delta$ via
\begin{align}
    \label{eq:laplacian_definition}
    \Delta w_t(x)=&\sum\limits_{i=1}^{d}\frac{\partial^2}{\partial x_i^2}w_t(x)=\divv(\nabla w_t)(x),
\end{align}
for scalar field $w_t:\R^d\to\R$.
\begin{theorem}[Fokker-Planck Equation]
\label{thm:fokker_planck}
Let $p_t$ be a probability path and let us consider the SDE
\begin{align*}
    X_0\sim \pinit, \quad \dd X_t = u_t(X_t)\dd t + \sigma_t\dd W_t.
\end{align*}
Then $X_t$ has distribution $p_t$ for all $0\leq t\leq 1$ if and only if the \themebf{Fokker-Planck equation} holds:
    \begin{align}
    \label{e:fokker_planck}
    \partial_t p_t(x) = -\divv (p_t u_t)(x)+\frac{\sigma_t^2}{2}\Delta p_t (x)\quad \text{ for all }x\in\R^d, 0\leq t\leq 1,
    \end{align}     
\end{theorem}
A self-contained proof of the Fokker-Planck equation can be found in \cref{subsec:proof_fokker_planck}. Note that \cref{thm:continuity_equation} is recovered from the Fokker-Planck equation when $\sigma_t=0$. The additional Laplacian term $\Delta p_t$ might be hard to rationalize at first. Those familiar with physics will note that the same term also appears in the heat equation (which is in fact a special case of the Fokker-Planck equation). Heat diffuses through a medium. We also add a diffusion process (not a physical but a mathematical one) and hence we add this additional Laplacian term. 
Let us now use the Fokker-Planck equation to help us prove \cref{thm:langevin_trick}.

\begin{proof}[Proof of \Cref{thm:langevin_trick}]
 By \cref{thm:fokker_planck}, we need to show that that the SDE defined in \cref{eq:sde_extension} satisfies the Fokker-Planck equation for $p_t$. We can do this by direction calculation:
\begin{align*}
    \partial_t p_t(x) \overset{(i)}{=}& - \divv(p_t\uref_t)(x)\\
    \overset{(ii)}{=}& - \divv(p_t\uref_t)(x) -\frac{\sigma_t^2}{2}\Delta p_t(x)+\frac{\sigma_t^2}{2}\Delta p_t(x)\\
    \overset{(iii)}{=}& - \divv(p_t\uref_t)(x) -\divv(\frac{\sigma_t^2}{2}\nabla p_t)(x)+\frac{\sigma_t^2}{2}\Delta p_t(x)\\
    \overset{(iv)}{=}& - \divv(p_t\uref_t)(x) -\divv(p_t\left[\frac{\sigma_t^2}{2}\nabla \log p_t\right])(x)+\frac{\sigma_t^2}{2}\Delta p_t(x)\\
    \overset{(v)}{=}&- \divv\left(p_t\left[\uref_t+\frac{\sigma_t^2}{2}\nabla \log p_t\right]\right)(x)+\frac{\sigma_t^2}{2}\Delta p_t(x),
\end{align*}
where in $(i)$ we used \cref{thm:continuity_equation}, in $(ii)$ we added and subtracted the same term, in $(iii)$ we used the definition of the Laplacian (\cref{eq:laplacian_definition}), in $(iv)$ we used that $\nabla\log p_t=\frac{\nabla p_t}{p_t}$, and in $(v)$ we used the linearity of the divergence operator. The above derivation shows that the SDE defined in \cref{eq:sde_extension} satisfies the Fokker-Planck equation for $p_t$. By \cref{thm:fokker_planck}, this implies $X_t\sim p_t$ for $0\leq t\leq 1$, as desired.
\end{proof}

\begin{figure}[!t]
    \centering
    \includegraphics[width=\textwidth]{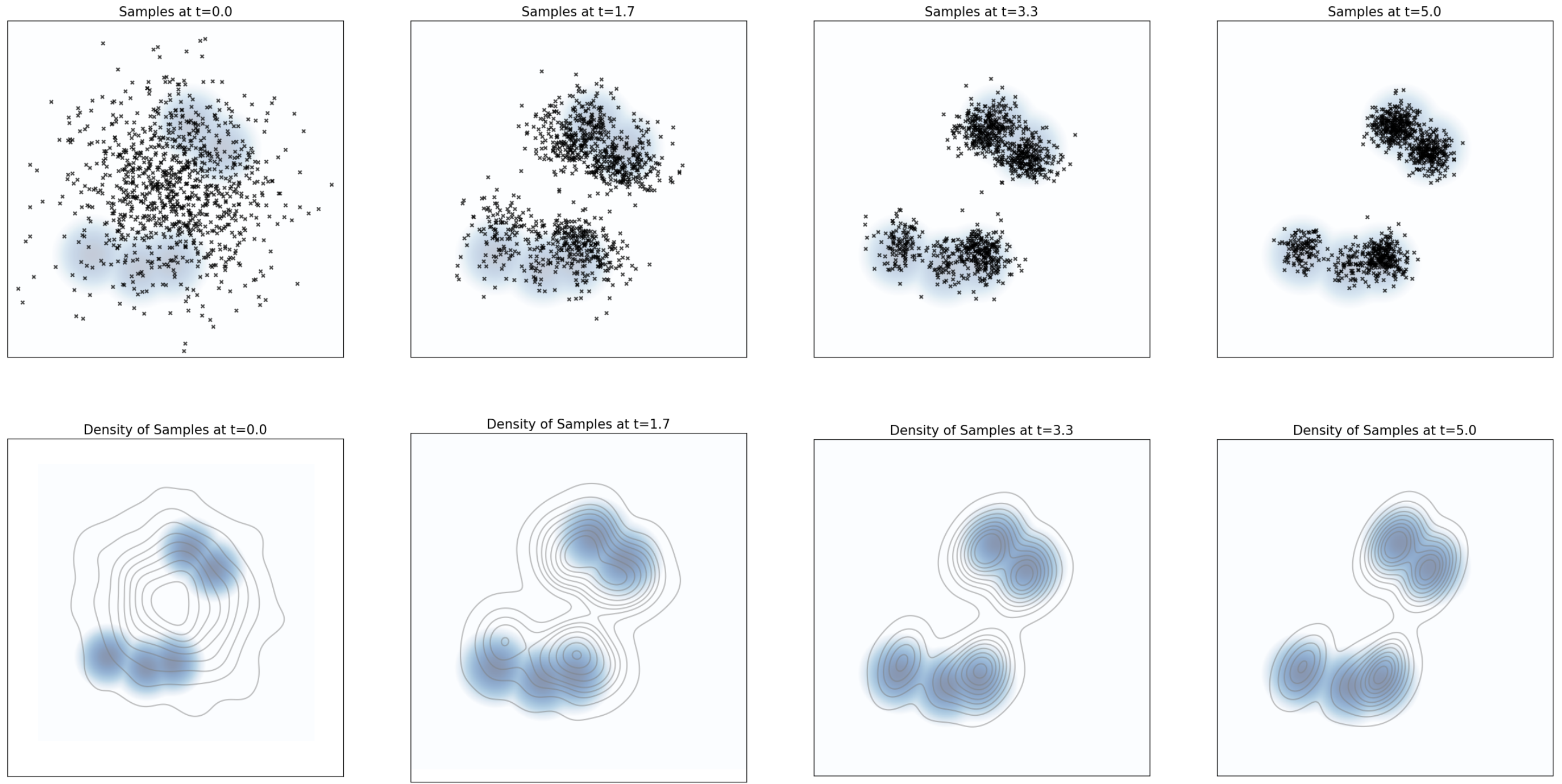}
    \label{fig:langevin}
    \caption{Top row: Particles evolving under the Langevin dynamics given by \cref{eq:langevin_dynamics}, with $p(x)$ taken to be a Gaussian mixture with 5 modes. Bottom row: A kernel density estimate of the same samples shown in the top row. As one can see, the distribution of samples converges to the equilibrium distribution $p$ (blue background colour).}
\end{figure}
\begin{remarkbox}[Optional: Langevin Dynamics]
\label{remark:langevin}
The above construction has a famous special case when the probability path is constant, i.e. $p_t=p$ for a fixed distribution $p$. In this case, we set $\uref_t=0$ and obtain the SDE
\begin{equation}
    \dd X_t = \frac{\sigma_t^2}{2}\nabla\log p(X_t)\dd t + \sigma_t dW_t \label{eq:langevin_dynamics},
\end{equation}
which is commonly known as \themebf{Langevin dynamics}. The fact that $p_t$ is constant implies that $\partial_tp_t(x)=0$. It follows immediately from \cref{thm:langevin_trick} that these dynamics satisfy the Fokker-Planck equation for the static path $p_t=p$ in \cref{thm:langevin_trick}. Therefore, we may conclude that $p$ is a stationary distribution of Langevin dynamics:
\begin{align*}
    X_0 \sim p\quad \Rightarrow \quad X_t \sim p\quad (t\geq 0).
\end{align*}
As with many Markov processes, these dynamics converge to the stationary distribution $p$ under rather general conditions. That is, if we instead we take $X_0 \sim p' \neq p$, so that $X_t \sim p_t'$, then under mild conditions $p_t \to p$. This fact makes Langevin dynamics extremely useful, and it accordingly serves as the basis for e.g., \textbf{molecular dynamics} simulations, and many other Markov chain Monte Carlo (MCMC) methods across Bayesian statistics and the natural sciences. In particular, the Ornstein-Uhlenbeck processes are recovered as the special case of the Langevin dynamics when $p$ is a Gaussian, and serve as the basis for initial formulations of diffusion models. 
% The Langevin dynamics also have elegant connections to the theories of gradients flows and optimal transport, both of which are beyond the scope of these notes.
\end{remarkbox}
\begin{remarkbox}[Optional: GLASS Flows, Stochastic evolution with ODEs]
The remarkable property of SDE sampling (compared to ODEs) is that the evolution becomes stochastic, i.e. the initial point $X_0$ does not fully determine $X_t$ for $t>0$. Perhaps surprisingly, it is also possible to get the same stochastic transitions purely via ODEs via a simple sampling trick called 
\emph{GLASS Flows} \citep{holderrieth2025glass}. This allows to exploit the stochastic nature of SDEs (e.g. via search algorithms) while keeping the efficiency of ODEs.
\end{remarkbox}
\subsection{Score Matching}
\label{subsec:score_matching}
It remains to show how we can learn the marginal score function $\nabla\log p_t(x)$. Of course, for Gaussian probability paths, we can simply transform $\uref_t(x)$ by \cref{prop:conversion_formula_gaussian_prob_path}. However, what about in general? It turns out that we can also learn marginal score functions directly. To approximate the marginal score $\nabla\log p_t$, we use a neural network that we call \themebf{score network} $s_t^\theta:\mathbb{R}^d\times[0,1]\to\mathbb{R}^d$. In the same way as before, we can design a \themebf{score matching} loss and a \themebf{denoising score matching} loss:
\begin{align*}
    \mathcal{L}_{\text{SM}}(\theta) &=\mathbb{E}_{t\sim\text{Unif},\,z\sim \pdata,\, x\sim p_t(\cdot|\dap)}\left[\left\|s_t^\theta(x) - \nabla\log p_t(x)\right\|^2\right] &&\blacktriangleright\,\,\text{score matching loss}\\
    \mathcal{L}_{\text{CSM}}(\theta) &=\mathbb{E}_{t\sim\text{Unif},\,z\sim \pdata,\, x\sim p_t(\cdot|\dap)}\left[\left\|s_t^\theta(x) - \nabla\log p_t(x|\dap)\right\|^2\right]  &&\blacktriangleright\,\,\text{conditional score matching loss}
\end{align*}
where again the difference is using the marginal score $\nabla\log p_t(x)$ vs. using the conditional score $\nabla\log p_t(x|z)$. %\footnote{It would be more consistent to call it \emph{conditional} score matching loss but \emph{denoising} score matching is a term used in the literature. We will realize later why it is called \emph{denoising} score matching.} 
As before, we ideally would want to minimize the score matching loss but can't because we don't know $\nabla\log p_t(x)$. But similarly as before, the denoising score matching loss is a tractable alternative:
\begin{theorem}
\label{thm:dsm_loss}
The score matching loss equals the denoising score matching loss up to a constant:
\begin{align*}
\mathcal{L}_{\text{SM}}(\theta) = \mathcal{L}_{\text{CSM}}(\theta) + C,
\end{align*}
where $C$ is independent of parameters $\theta$. Therefore, their gradients coincide:
\begin{align*}
\nabla_\theta \mathcal{L}_{\text{SM}}(\theta) = \nabla_\theta \mathcal{L}_{\text{CSM}}(\theta).
\end{align*}
In particular, for the minimizer $\theta^*$, it will hold that $s_t^{\theta^*}=\nabla\log p_t$. 
\end{theorem}
\begin{proof}
Note that the formula for $\nabla\log p_t$ (\cref{e:marginal_score}) looks the same as the formula for $\uref_t$ (\cref{eq:marginal_vector_field}). Therefore, the proof is identical to the proof of \cref{thm:fm_loss} replacing $\uref_t$ with $\nabla\log p_t$.
\end{proof}

\begin{examplebox}[Denoising Diffusion Models: Score Matching for Gaussian Probability Paths]
Let us instantiate the denoising score matching loss for the case of $p_t(x|z)=\mathcal{N}(\alpha_tz,\beta_t^2I_d)$. As we derived in \cref{eq:cond_score_gaussian}, the conditional score $\nabla\log p_t(x|z)$ has the formula
\begin{align}
\label{e:marginal_vf_cond_score_restated}
    \nabla\log p_t(x|z) = -\frac{x-\alpha_t z}{\beta_t^2}.
\end{align}
Plugging in this formula, the conditional score matching loss becomes:
\begin{align*}
        \mathcal{L}_{\text{CSM}}(\theta) &=\mathbb{E}_{t\sim\Unif,\,z\sim \pdata,\,x\sim p_t(\cdot|\dap)}\left[\left\|s_t^\theta(x)+\frac{x-\alpha_t z}{\beta_t^2}\right\|^2\right]\\&\overset{(i)}{=}\mathbb{E}_{t\sim\Unif,\,z\sim \pdata,\,\epsilon\sim \mathcal{N}(0,I_d)}\left[\left\|s_t^\theta(\alpha_tz+\beta_t\epsilon)+\frac{\epsilon}{\beta_t}\right\|^2\right]\\
        &=\mathbb{E}_{t\sim\Unif,\,z\sim \pdata,\,\epsilon\sim \mathcal{N}(0,I_d)}\left[\frac{1}{\beta_t^2}\left\|\beta_ts_t^\theta(\alpha_tz+\beta_t\epsilon)+\epsilon\right\|^2\right]
\end{align*}
where in $(i)$ we plugged in \cref{eq:sampling_procedure_gaussian_path_restated_for_fm} and replaced $x$ by $\alpha_tz+\beta_t\epsilon$. Note that the network $s_t^\theta$ essentially learns to predict the noise that was used to corrupt a data sample $z$. This explains why the above training loss is  called \themebf{denoising score matching}. It was soon realized that the above loss is numerically unstable for $\beta_t\approx 0$ close to zero (i.e. denoising score matching only works if you add a sufficient amount of noise). In some of the first works on denoising diffusion models (see \textbf{Denoising Diffusion Probabilitic Models}, \citep{ho2020denoising}) it was therefore proprosed to drop the constant $\frac{1}{\beta_t^2}$ in the loss and reparameterize $s_t^\theta$ into a \themebf{noise predictor} network $\epsilon_t^\theta:\mathbb{R}^d\times[0,1]\to\mathbb{R}^d$ via:
\begin{align*}
    -\beta_t s_t^\theta(x) = \epsilon_t^\theta(x)\quad \Rightarrow \quad \mathcal{L}_{\text{DDPM}}(\theta) =&\mathbb{E}_{t\sim\Unif,z\sim \pdata, \epsilon\sim \mathcal{N}(0,I_d)}\left[\|\epsilon_t^\theta(\alpha_tz+\beta_t\epsilon)-\epsilon\|^2\right]
\end{align*}
As before, the network $\epsilon_t^\theta$ essentially learns to predict the noise that was used to corrupt a data sample $z$. In \cref{alg:training_score_matching_gaussian_paths}, we summarize the training procedure.
\end{examplebox}

\begin{algorithm}[h]
\caption{Score Matching Training Procedure for Gaussian probability path}
\label{alg:training_score_matching_gaussian_paths}
\begin{algorithmic}[1]
\REQUIRE A dataset of samples $z\sim \pdata$, score network $s_t^\theta$ or noise predictor $\epsilon_t^\theta$
\FOR{each mini-batch of data}
    \STATE Sample a data example $\dap$ from the dataset.
    \STATE Sample a random time $t \sim \text{Unif}_{[0,1]}$.
    \STATE Sample noise $\epsilon\sim\mathcal{N}(0,I_d)$
    \STATE Set $x_t=\alpha_t z + \beta_t\epsilon$\hfill (\text{General case: }$x_t\sim p_t(\cdot|z)$)
    %\IF{Flow matching}
    \STATE Compute loss
    \begin{align*}
        \mathcal{L}(\theta) =& \|s_t^\theta(x_t)+\frac{\epsilon}{\beta_t}\|^2 \quad &(\text{General case: }=\|s_t^\theta(x_t)-\nabla\log p_t(x_t|z)\|^2)\\
    \text{Alternatively: }\mathcal{L}(\theta) =& \|\epsilon_t^\theta(x_t)-\epsilon\|^2
    \end{align*}
    %\ENDIF
    % \IF{Score matching}
    % \STATE
    % \begin{align*}
    %     \mathcal{L}(\theta) =& \|s_t^\theta(x_t)+\frac{\epsilon}{\beta_t}\|^2 \quad &(\text{General case: }=\|s_t^\theta(x_t)+\nabla\log p_t(x_t|z)\|^2)
    % \end{align*}
    % \ENDIF
    \STATE Update the model parameters $\theta$ via gradient descent on $\mathcal{L}(\theta)$.
\ENDFOR
\end{algorithmic}
\end{algorithm}

Let us summarize the results of this section:
\begin{summarybox}[Score Functions, Score Matching, and Stochastic Sampling]
Let $p_t(x|z),p_t(x)$ be the conditional and marginal probability path. The \themebf{conditional score function} is given by $\nabla\log p_t(x|z)$ and the \themebf{marginal score function} is given by $\nabla\log p_t(x)$. For every diffusion coefficient $\sigma_{t}\geq 0$, the trajectories of the following SDE follow the probability path:
\begin{align}
\label{eq:sde_extension_restated}
X_0\sim&\pinit,\quad \dd X_t =\left[\uref_t(X_t)+\frac{\sigma_t^2}{2}\nabla\log p_t(X_t)\right]\dd t +\sigma_t\dd W_t\\
    \Rightarrow X_t\sim& p_t\quad (0\leq t\leq 1),
\end{align}
where is $\uref_t(x)$ be the marginal vector field as before (see \cref{eq:marginal_vector_field}).

\paragraph{Score Matching.} To learn the marginal score function $\nabla\log p_t(x)$, we can use a \themebf{score network} $s_t^\theta$ and train it via \themebf{denoising score matching}
\begin{align}
    \label{eq:dsm}
\mathcal{L}_{\text{CSM}}(\theta) &= \mathbb{E}_{z\sim \pdata, \,t\sim \text{Unif},\,x\sim p_t(\cdot|\dap)}[\|s_t^\theta(x) - \nabla\log p_t(x|\dap)\|^2] \quad &(\text{denoising score matching loss})
\end{align}
% For every diffusion coefficient $\sigma_t\geq 0$, simulating the SDE (e.g. via \cref{alg:sampling_diffusion_model})
% \begin{align}
% \label{eq:sde_sampling_restated_again}
%     X_0\sim&\pinit,\quad \dd X_t =\left[u_t^\theta(X_t)+\frac{\sigma_t^2}{2}s_t^\theta(X_t)\right]\dd t +\sigma_t\dd W_t
% \end{align}
% will result in generating approximate samples from $\pdata$.

\paragraph{Gaussian Probability Paths.} For the - most important -  case of a Gaussian probability path $p_t(x|z)=\mathcal{N}(x;\alpha_tz,\beta_t^2I_d)$, there is no need to train $s_t^\theta$ and $u_t^\theta$ separately as we can convert them via the formula:
\begin{align*}
u_t^\theta(x)=&a_t s_t^\theta(x)+b_tx,\quad a_t=\left(\beta_t^2\frac{\dot{\alpha}_t}{\alpha_t}-\dot{\beta}_t\beta_t\right), b_t=\frac{\dot{\alpha}_t}{\alpha_t}
\end{align*}
After training, we can simulate the following SDE
\begin{align}
X_0\sim\pinit,\quad \dd X_t =&\left[\left(1+\frac{\sigma_{t}^2}{2a_t}\right)u_t^\theta(X_t)-\frac{\sigma_{t}^2b_t}{2a_t}X_t\right]\dd t +\sigma_t\dd W_t\\
=&\left[\left(a_t+\frac{\sigma_{t}^2}{2}\right)s_t^\theta(X_t)+b_tX_t\right]\dd t +\sigma_t\dd W_t
\end{align}
for any diffusion coefficient $\sigma_{t}\geq 0$ 
to obtain approximate samples $X_1\sim \pdata$. One can empirically find the optimal $\sigma_t\geq 0$.
\end{summarybox}
% \paragraph{Adding noise to data.}

% \paragraph{Frame as probability path.}

% \paragraph{How do I go the other way?} Transport mass - using Fokker-Planck equation.

% \paragraph{Weak time-reversal.} Probability flow ODE.

% \paragraph{Adding Langevin dynamics.}

% \paragraph{Special case: Proper time-reversal.}

% \paragraph{Specify marginals via an SDE.} Example of Ohrnstein-Uhlenbeck process.

% Instantaneous Change of Variables (evaluating log-likelihood).
% Deterministic sampling. Stochastic sampling:
% Linear ODEs, SDEs with affine drift coefficients, Time-reversal idea.
\newpage
\section{Guidance: How To Condition on a Prompt}
\label{sec:guidance}

So far, the generative models we considered were \themebf{unguided}, e.g. an image model would simply generate \themeit{some} image. Mathematically speaking, this meant that our model returned samples from an \emph{unconditional} data distribution $\pdata(z)$.  However, in most cases, our goal is not to merely generate an arbitrary object, but to generate an object \textbf{\sffamily conditioned on some additional information}. In other words, we want to \textbf{guide} the model to generate objects of a certain kind.  For example, one might imagine a generative model for images which takes in a text prompt $y$, and then generates an image $x$ that fits to the text prompt $y$. As discussed in \cref{sec:introduction}, this means that we want to sample from $\pdata(z|y)$, that is, the \textbf{guided data distribution} \textbf{\sffamily conditioned on $y$}. We are going to discuss this in this section.

\begin{remarkbox}[Terminology]
To avoid a notation and terminology clash with the use of the word ``\text{conditional}'' to refer to conditioning on $z \sim \pdata$ (conditional probability path/vector field), we will make use of the term \themebf{guided} to refer specifically to conditioning on $y$ such as a text prompt.
\end{remarkbox}

\subsection{Vanilla Guidance}

First, we discuss the ``standard'' way of how one would go about building a guided generative model. The short answer is as follows: We simply provide the input prompt $y$ to the network during training and inference and do everything in the same way as before. We formalize this in the following. We think of a conditioning variable or prompt $y$ to live in a space $\mathcal{Y}$. When $y$ corresponds to a text-prompt, for example, $\mathcal{Y}$ is the space of all texts. When $y$ corresponds to some discrete class label, $\mathcal{Y}$ would be discrete. We pose no constraints on $\mathcal{Y}$.

% \begin{remarkbox}[Guided vs. Conditional Terminology]
%     In these notes, we opt to use the term \themebf{guided} in place of \themebf{conditional} to refer to the act of conditioning on $y$. Here, we will refer to e.g., a \themebf{guided} vector field $\uref_t(x|y)$ and a \themebf{conditional} vector field $\uref_t(x|z)$. This terminology is consistent with other works such as \cite{lipman2024flow}. 
% \end{remarkbox}

% The goal of \themebf{guided generative modeling} is thus to be able to sample from $\pdata(x|y)$ \themeit{for any such $y$}. In the language of flow and score matching, and in which our generative models correspond to the simulation of ordinary and stochastic differential equations, this can be phrased as follows.

% \begin{ideabox}[Guided Generative Model]
We define a \themebf{guided diffusion model} to consist of a \themebf{guided vector field} $u_t^{\theta}(\cdot | y)$, parameterized by some neural network, and a time-dependent diffusion coefficient $\sigma_t$, together given by
\begin{align*}
    \textbf{\sffamily Neural network:}&\, u^\theta: \mathbb{R}^d \times \mathcal{Y} \times [0,1] \to \mathbb{R}^d,\,\, (x,y,t) \mapsto u_t^{\theta}(x|y)\\
    \textbf{\sffamily Fixed:}&\, \sigma_t: [0,1] \to [0,\infty),\,\, t \mapsto \sigma_t
\end{align*}
Notice the difference from summary \ref{summary:diffusion_model}: we are additionally guiding $u_t^\theta$ with the input $y\in \mathcal{Y}$. For any such $y \in  \mathcal{Y}$, samples may then be generated from such a model as follows:
\begin{align*}
    \textbf{\sffamily Initialization:}\quad X_0&\sim\pinit \quad  &&\blacktriangleright\,\,\text{Initialize with simple distribution (such as a Gaussian)}\\
    \textbf{\sffamily Simulation:}\quad \dd X_t &= u_t^\theta(X_t|y)\dd t + \sigma_t\dd W_t\quad &&\blacktriangleright\,\,\text{Simulate SDE from $t=0$ to $t=1$.}\\
    \textbf{\sffamily Goal:}\quad X_1 &\sim  \pdata(\cdot | y) \quad &&\blacktriangleright\,\,\text{Goal is for $X_1$ to be distributed like $\pdata(\cdot|y)$.}
\end{align*}
When $\sigma_t = 0$, we say that such a model is a \themebf{guided flow model}. In the following, we restrict ourselves to flow matching and flow models to make things more concise but everything applies similarly to the general case.

Next, we discuss: How would we train a guided flow model $u_t^\theta(x|y)$? A simple trick might to fix our choice of $y$, and to take our data distribution as $p_{\text{data}}(x|y)$. Then we have recovered the unguided generative problem as before, and we can accordingly construct a generative model using the conditional flow matching objective, viz.,
\begin{equation}
    \mathbb{E}_{z \sim p_{\text{data}}(\cdot|y), x \sim p_t(\cdot|z)} \lVert u_t^{\theta}(x|y) - \uref_t(x|z)\rVert^2.
\end{equation}
Note that the label $y$ does not affect the conditional probability path $p_t(\cdot|z)$ or the conditional vector field $\uref_t(x|z)$ (although in principle, we could make it dependent). % Since the label doesn't actually affect the guided conditional probability path, the above can be further simplified with $\uref_t(x|z,y) = \uref_t(x|z)$ (we might say: ``the guided conditional probability path is the same as the conditional probability path''), which we have worked with already. 
Expanding the expectation over all such choices of $y$,  we thus obtain a \themebf{guided conditional flow matching objective}
\begin{equation}
    \label{eq:guided_cfm}
    \mathcal{L}_{\text{CFM}}^{\text{guided}}(\theta) = \mathbb{E}_{(z,y) \sim p_{\text{data}}(z,y),\,t\sim \text{Unif}[0,1],\,x\sim p_t(\cdot|z)} \lVert u_t^{\theta}(x|y) - \uref_t(x|z)\rVert^2.
\end{equation}
One of the main differences between the guided objective in \cref{eq:guided_cfm} and the unguided objective from \cref{eq:cfm} is that here we are sampling $(z,y) \sim \pdata$ rather than just $z \sim \pdata$. The reason is that our data distribution is now, in principle, a joint distribution over e.g., both images $z$ and text prompts $y$. In practice, this means that a PyTorch implementation of \cref{eq:guided_cfm} would involve a dataloader which returned batches of \textbf{both $z$ and $y$}.

\begin{figure}[!t]
    \centering
    \includegraphics[width=\linewidth]{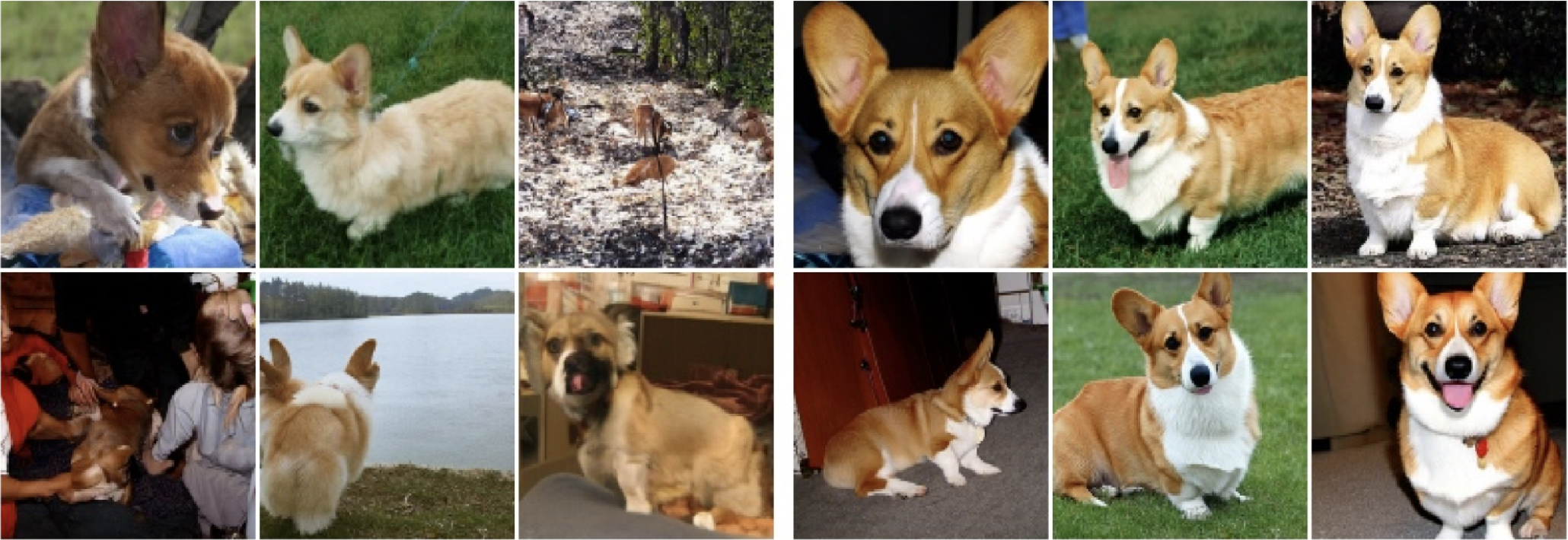}
    \caption{Image generation with prompt/class $y=$``corgi dog''. Left: samples generated with vanilla guidance - the images do not fit well to the prompt. Right: samples generated with classifier guidance and $w = 4$. As shown, classifier-free guidance improves the adherence to the prompt. Figure taken from \cite{cfg}.}
    \label{fig:guidance}
\end{figure}

\begin{figure}[!t]
    \centering
    \includegraphics[width=0.9\linewidth]{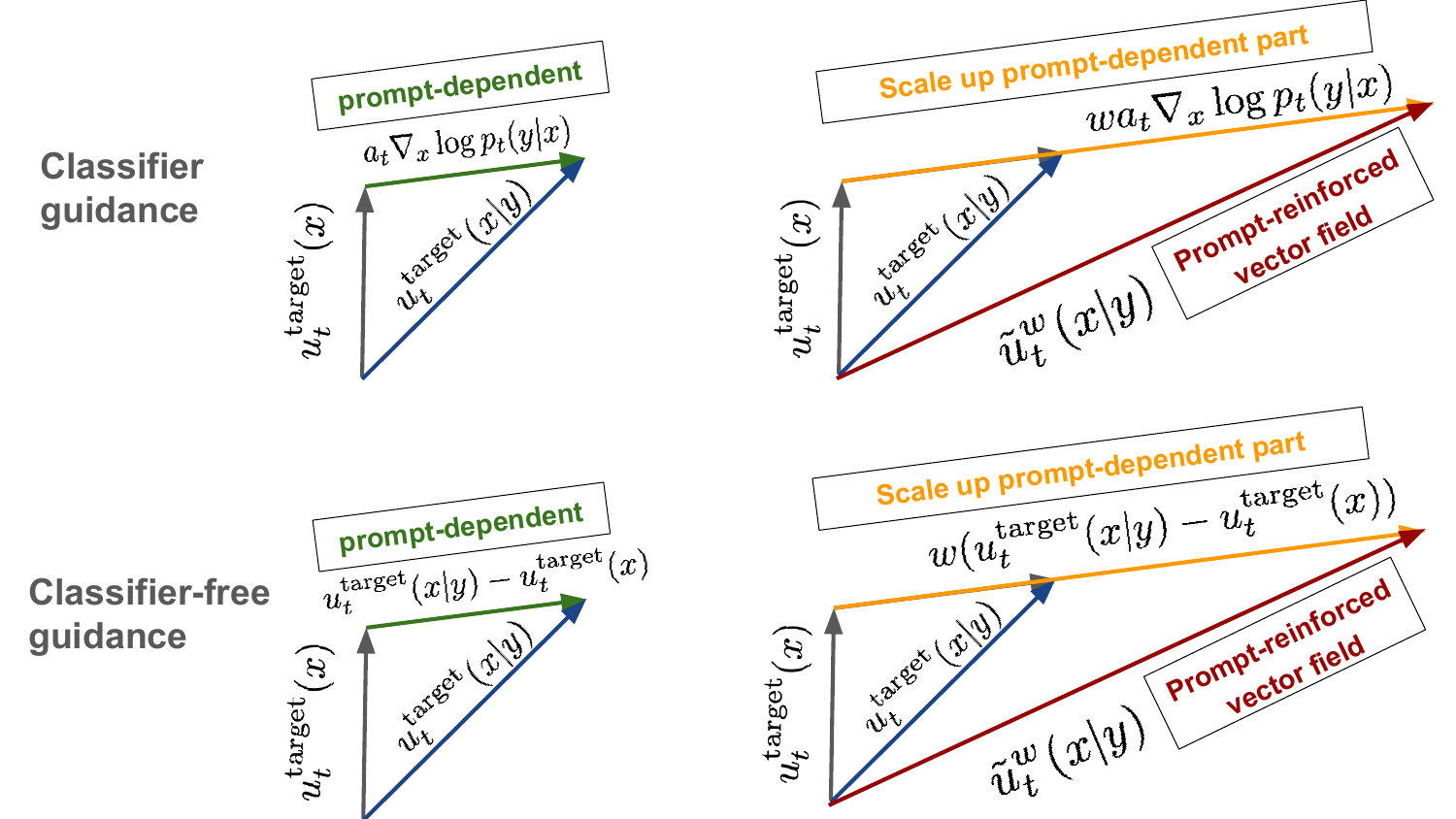}
    \caption{Illustration of classifier and classifier-free guidance. Classifier guidance decomposes the guided vector field $\uref_t(x|y)$ and the gradient of a classifier $\log p_t(y|x)$ and scales up the classifier with guidance scale $w>1$. Classifier-free guidance scales up the difference between both vector fields, thereby achieving the same effect but without having to train a separate classifier model.}
\label{fig:classifier_and_cfg_illustrative_figure}
\end{figure}

\subsection{Classifer-Free Guidance}

In theory, vanilla guidance should lead to a faithful generation procedure of $\pdata(\cdot|y)$. However, it was soon empirically realized that images samples with this procedure did not fit well enough to the desired label $y$ (see  \cref{fig:guidance}). This can have a diversity of reasons: the model might underfit (i.e. we do not actually learn the true marginal vector field) or our data might be imperfect (e.g. text-image pairs from the world wide web have a lot of errors). Therefore, to truly generate samples that fit better to a prompt, we have to find a way to  artificially reinforce the prompt variable $y$. The main technique for doing so is called  \themebf{classifier-free guidance} that is widely used in the context of state-of-the-art diffusion models, and which we discuss next. 

\paragraph{Classifier Guidance.} For simplicity, we will focus here on the case of Gaussian probability paths. Recall from \cref{eq:gaussian_conditional_probability_paths}
 that a Gaussian conditional probability path is given by $p_t(\cdot|\dap) = \mathcal{N}(\alpha_t \dap,\beta_t^2 I_d)$ 
where the \text{noise schedulers} $\alpha_t$ and $\beta_t$ are continuously differentiable, monotonic, and satisfy $\alpha_0 = \beta_1 = 0$ and $\alpha_1 = \beta_0 = 1$. Further, recall that we can use \cref{prop:conversion_formula_gaussian_prob_path} to rewrite the guided vector field $\uref_t(x|y)$ in the following form using the guided score function $\nabla\log p_t(x|y)$
\begin{equation}
    \uref_t(x|y) = a_t\nabla \log p_t(x|y)+b_tx,
\end{equation}
Next, realize that $p_t(x|y)$ is a conditional density. Hence, we can use  Bayes' rule to rewrite the guided score as
\begin{align}
p_t(x|y)=&\frac{p_t(x)p_t(y|x)}{p_t(y)}\\
\nabla \log p_t(x|y) =& \nabla \log \left(\frac{p_t(x)p_t(y|x)}{p_t(y)}\right) = \nabla \log p_t(x) + \nabla \log p_t(y|x), 
\label{eq:bayes_rule}
\end{align}
where we used that the gradient $\nabla$ is taken with respect to the variable $x$, so that $\nabla \log p_t(y) = 0$. We may thus rewrite 
\begin{align*}
    \uref_t(x|y) = b_tx + a_t(\nabla \log p_t(x) + \nabla \log p_t(y|x)) = \uref_t(x) + a_t \nabla \log p_t(y|x).
\end{align*}
Notice the shape of the above equation: The guided vector field $\uref_t(x|y)$ is a sum of the unguided vector field $\uref_t(x)$ \emph{plus} a gradient of the likelihood $p_t(y|x)$ of the guidance variable $y$. As people observed that their image $x$ did not fit their prompt $y$ well enough, it was a natural idea to scale up the contribution of the $\nabla \log p_t(y|x)$ term, yielding
\begin{align}
    \tilde{u}_t(x|y) = \uref_t(x) + w a_t \nabla \log p_t(y|x),\quad &(\text{classifier guidance})
\end{align}
where $w > 1$ is known as the \themebf{guidance scale}. How can we learn the term $\log p_t(y|x)$? Note that this can be considered as a sort of classifier of noised data (i.e. it gives the log-likelihoods of $y$ given $x$). So we can simply learn it via supervised learning. This leads to \themebf{classifier guidance} \cite{classifier_guidance, yangsong_sde} (see \cref{fig:classifier_and_cfg_illustrative_figure} for an illustration). Classifier guidance was largely superseded by classifier-free guidance, which is why we will not discuss it further here. However, it forms the basis for the classifier-free guidance, as we will see next. Finally, note that this is a heuristic: for $w \neq 1$, it holds that $\tilde{u}_t(x|y) \neq \uref_t(x|y)$, i.e. therefore not the ``true'' guided vector field.

\paragraph{Classifier-Free Guidance.} While classifier guidance is possible in principle, it comes with difficulties: The first thing is that we need to train a classifier alongside a flow/diffusion model - so we have 2 networks instead of 1. Further, if the $y$ is high-dimensional, e.g. a text prompt and not just a class, then $p_t(y|x)$ might be very hard to learn and the gradient $\nabla\log p_t(y|x)$ hard to obtain. For this reason, \themebf{classifier-free guidance} \citep{cfg} was introduced. Classifier-free guidance results in the theoretically equivalent effect as classifier guidance but without having to train a separate classifier.

To do so, we may again apply the equality $$\nabla \log p_t(x|y) = \nabla \log p_t(x) + \nabla \log p_t(y|x)$$ to obtain 
\begin{align*}\tilde{u}_t(x|y) &= \uref_t(x) + w a_t \nabla \log p_t(y|x)\\
&= \uref_t(x) + w a_t (\nabla \log p_t(x|y) - \nabla \log p_t(x))\\
&= \uref_t(x) - (w b_tx + w a_t \nabla \log p_t(x)) + (w b_t x + w a_t \nabla \log p_t(x|y))\\
&= (1-w) \uref_t(x) + w \uref_t(x|y).\end{align*}
We may therefore express the scaled guided vector field $\tilde{u}_t(x|y)$ as the linear combination of the unguided vector field $\uref_t(x)$ with the guided vector field $\uref_t(x|y)$. The idea might then to to train both an unguided $\uref_t(x)$ (using e.g., \cref{eq:cfm}) as well as a guided $\uref_t(x|y)$ (using e.g., \cref{eq:guided_cfm}), and then combine them at inference time to obtain $\tilde{u}_t(x|y)$. "But wait!", you might ask, "wouldn't we need to train two models then !?". It turns out that we can train both in one model: we may augment our label set with a new, additional $\varnothing$ label that denotes \textbf{the absence of conditioning}. We can then treat $\uref_t(x)=\uref_t(x|\varnothing)$. With that, we do not need to train a separate model to reinforce the effect of a hypothetical classifier. This approach of training a conditional and unconditional model in one (and subsequently reinforcing the conditioning) is known as \themebf{classifier-free guidance} (CFG) \cite{cfg} (see \cref{fig:classifier_and_cfg_illustrative_figure} for an illustration). 

\begin{remarkbox}[Derivation for general probability paths]
Note that the construction
\begin{equation*}
    \tilde{u}_t(x|y) = (1-w) \uref_t(x) + w \uref_t(x|y),
\end{equation*}
is equally valid for any choice probability path, not just a Gaussian one. When $w=1$, it is straightforward to verify that $\tilde{u}_t(x|y)=\uref_t(x|y)$. Our derivation using Gaussian paths was simply to illustrate the intuition behind the construction, and in particular of amplifying the contribution of a hypothetical ``classifier'' $\nabla \log p_t(y|x)$.
\end{remarkbox}

\paragraph{Training and Classifier-Free Guidance.} We must now amend the guided conditional flow matching objective from \cref{eq:guided_cfm} to account for the possibility of $y = \varnothing$. The challenge is that when sampling $(z,y) \sim \pdata$, we will never obtain $y = \varnothing$. It follows that we must introduce the possibility of $y = \varnothing$ artificially. To do so, we will define some hyperparameter $\eta$ to be the probability that we discard the original label $y$, and replace it with $\varnothing$. We thus arrive at our \themebf{CFG conditional flow matching training objective}
\begin{align}
    \mathcal{L}_{\text{CFM}}^{\text{CFG}}(\theta) &= \,\,\mathbb{E}_{\square} \lVert u_t^{\theta}(x|y) - \uref_t(x|z)\rVert^2\\
    \square &= (z,y) \sim p_{\text{data}}(z,y),\, t \sim \text{Unif}[0,1],\, x \sim p_t(\cdot|z),\text{replace }y=\varnothing\text{ with prob. }\eta
\end{align}

\begin{algorithm}[h]
\caption{Classifier-free guidance training for Gaussian probability path $p_t(x|z)=\mathcal{N}(x;\alpha_tz,\beta_t^2I_d)$}
\label{alg:cfg_training}
\begin{algorithmic}[1]
\REQUIRE Paired dataset $(z,y)\sim \pdata$, neural network $u_t^\theta$
\FOR{each mini-batch of data}
    \STATE Sample a data example $(\dap,y)$ from the dataset.
    \STATE Sample a random time $t \sim \text{Unif}_{[0,1]}$.
    \STATE Sample noise $\epsilon\sim\mathcal{N}(0,I_d)$
    \STATE Set $x=\alpha_t z + \beta_t \epsilon$
    % \hfill (\text{General case: }$x\sim p_t(\cdot|z)$)
    % %\IF{Flow matching}
    \STATE With probability $p$ drop label: $y\leftarrow \varnothing$
    \STATE Compute loss
    \begin{align*}
        \mathcal{L}(\theta) =& \|u_t^\theta(x|y)-(\dot{\alpha}_tz+\dot{\beta}_t\epsilon)\|^2 
    \end{align*}
    \STATE Update the model parameters $\theta$ via gradient descent on $\mathcal{L}(\theta)$.
\ENDFOR
\end{algorithmic}
\end{algorithm}

We summarize our findings below.

\begin{figure}[!t]
    \centering
    \includegraphics[width=\linewidth]{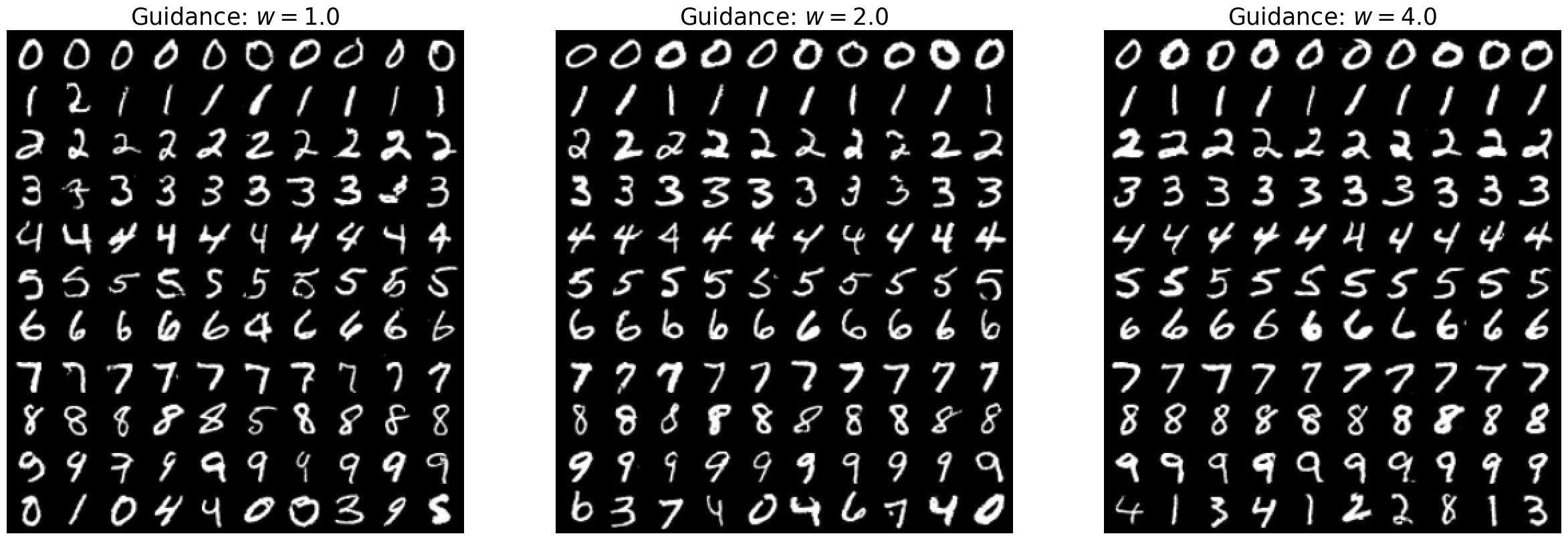}
    \caption{The effect of classifier-free guidance applied at various guidance scales for the MNIST dataset of hand-written digits. Left: Guidance scale set to $w = 1.0$. Middle: Guidance scale set to $w = 2.0$. Right: Guidance scale set to $w = 4.0$. You will generate a similar image yourself in the lab three!}
    \label{fig:mnist_guidance}
\end{figure}

\begin{summarybox}[Classifier-Free Guidance for Flow Models]
Given the unguided marginal vector field $\uref_t(x|\varnothing)$, the guided marginal vector field $\uref_t(x|y)$, and a \themebf{guidance scale} $w > 1$, we define the \themebf{classifier-free guided vector field} $\tilde{u}_t(x|y)$ by 
\begin{equation}
    \tilde{u}_t(x|y) = (1-w) \uref_t(x|\varnothing) + w \uref_t(x|y).
    \label{eq:flow_cfg}
\end{equation}
By approximating $\uref_t(x|\varnothing)$ and $\uref_t(x|y)$ using the same neural network, we may leverage the following \themebf{classifier-free guidance CFM} (CFG-CFM) objective, given by
\begin{align}
    \label{eq:cfg_guided_cfm}
    \mathcal{L}_{\text{CFM}}^{\text{CFG}}(\theta) &= \,\,\mathbb{E}_{\square} \lVert u_t^{\theta}(x|y) - \uref_t(x|z)\rVert^2\\
    \square &= (z,y) \sim p_{\text{data}}(z,y),\, t \sim \text{Unif}[0,1],\, x \sim p_t(\cdot|z),\text{replace }y=\varnothing\text{ with prob. }\eta
\end{align}
In plain English, $\mathcal{L}_{\text{CFM}}^{\text{CFG}}$ might be approximated by
\begin{alignat*}{3}
    (z,y) &\sim \pdata(z,y) \quad\quad\quad\quad && \blacktriangleright \quad \text{Sample $(z,y)$ from data distribution.}\\
    t &\sim \text{Unif}[0,1) \quad\quad\quad\quad && \blacktriangleright \quad \text{Sample $t$ uniformly on $[0,1)$.}\\
    x &\sim p_t(x|z) \quad\quad\quad\quad && \blacktriangleright \quad \text{Sample $x$ from the conditional probability path $p_t(x|z)$.}\\
    \text{with prob.}&\,\eta,\, y \gets \varnothing \quad\quad\quad\quad && \blacktriangleright \quad \text{Replace $y$ with $\varnothing$ with probability $\eta$.}\\
    \widehat{\mathcal{L}_{\text{CFM}}^{\text{CFG}}(\theta)} &=  \lVert u_t^{\theta}(x|y) - \uref_t(x|z)\rVert^2 \quad\quad\quad\quad && \blacktriangleright \quad \text{Regress model against conditional vector field.}
\end{alignat*}
At inference time, for a fixed choice of $y$, we may sample via
\begin{alignat*}{3}
    \textbf{\sffamily Initialization:}\quad X_0&\sim\pinit(x) \quad  && \blacktriangleright\,\,\text{Initialize with simple distribution (such as a Gaussian)}\\
    \textbf{\sffamily Simulation:}\quad \dd X_t &= \tilde{u}_t^\theta(X_t|y)\dd t \quad && \blacktriangleright\,\,\text{Simulate ODE from $t=0$ to $t=1$.}\\
    \textbf{\sffamily Samples:}\quad X_1& \quad && \blacktriangleright\,\,\text{Goal is for $X_1$ to adhere to the guiding variable $y$.}
\end{alignat*}
\end{summarybox}
Note that the distribution of $X_1$ is not necessarily aligned with $X_1 \sim  \pdata(\cdot | y)$ anymore if we use a weight $w>1$. However, empirically, this shows better alignment with conditioning. Classifier-free guidance is therefore a  \textbf{heuristic} that is predominantly justified by its excellent empirical results. In fact, almost any image or video that you see that is AI-generated relied heavily on classifier-free guidance $w\geq 4$. In \cref{fig:guidance}, we illustrate class-based classifier-free guidance on 128x128 ImageNet, as in \cite{cfg}. Similarly, in \cref{fig:mnist_guidance}, we visualize the affect of various guidance scales $w$ when applying classifier-free guidance to sampling from the MNIST dataset of handwritten digits.

\begin{remarkbox}[Guidance for Diffusion Models]
It is straight-forward to extend the discussion from flow models to diffusion models. One simply replaces $u_t^\theta(x|y)$ by $\tilde{u}_t^\theta(x|y)$ and samples using SDEs as discussed in \cref{sec:training_generative_models}.
\end{remarkbox}

\newpage
\section{Building Large-Scale Image or Video Generators}
\label{sec:image_generation}
In the previous sections, we learned how to train a flow matching or diffusion model to sample from a distribution $\pdata(x|y)$. This recipe is general and can be applied to a variety of different data types and applications. In this section, we examine in depth the particular cases of large-scale image and video generation, and including well-known models such as \themeit{FLUX 2.0, Stable Diffusion 3, Nano Banana} and \themeit{VEO-3 or Meta Movie Gen Video}. Finally, we'll apply what we've learned so far in the lab to build our own version of  such models from scratch! This section is broadly arranged as follows:
\begin{itemize} 
    \item[1.] \themebf{Neural network architectures:} We first discuss how raw conditioning input, including the time $t$, and guidance variable $y_{\text{raw}}$ (i.e., a discrete class label or raw text), is converted, or \themebf{embedded} into a vector-valued form digestible by the model $u_t^\theta(x|y)$ itself. Then we discuss popular architectural choices for $u_t^\theta(x|y)$, including the \themebf{U-Net} and \themebf{diffusion transformer}.
    \item[2.] \themebf{Latent Space:} We discuss \themebf{variational autoencoders}, which allow for generative modeling in a lower dimensional \themebf{latent space}, thereby enabling ultra high-resolution image generation.
    \item[3.]\themebf{Case Studies:} Finally, we will examine in depth the two state-of-the-art image and video models mentioned above - \themeit{Stable Diffusion} and \themeit{Meta MovieGen} - to give you a taste of how things are done at scale.
\end{itemize}

\subsection{Neural Network Architectures}
\label{sec:image_architecture}

Let us first turn our attention toward the design of scalable neural network architectures for flow and diffusion models targeting image-like modalities (e.g., images and videos). Specifically, we'll explore how the task of the (guided) vector field $u_t^\theta(x|y)$ with parameters $\theta$ is implemented in practice. Note that the neural network must have 3 inputs: a vector $x\in\R^d$, a conditioning variable $y\in\mathcal{Y}$, and a time value $t\in [0,1]$, as well as one output, a vector $u_t^\theta(x|y)\in\mathbb{R}^d$. For low-dimensional distributions (e.g. the toy distributions we have seen in previous sections), it is sufficient to parameterize $u_t^\theta(x|y)$ as a multi-layer perceptron (MLP), otherwise known as a fully connected neural network. That is, in this simple setting, a forward pass through $u_t^\theta(x|y)$ would involve concatenating our input $x$, $y$, and $t$, and passing them through an MLP. However, for complex, high-dimensional distributions, such as those over images, videos, and proteins, an MLP will likely not suffice, and it is common to use special, application-specific architectures. For the remainder of this subsection, we will consider the case of \textbf{images} (and by extension, videos). First, we'll consider how the raw conditioning information - the time $t$ and the conditioning variable $y$ - are \themebf{embedded} into a vector-valued form digestible by the actual model. Second, we'll consider two common architectural architectural choices for such a model: the \themebf{U-Net} \citep{ronneberger2015u, ho2020denoising, unet_cite_1, classifier_guidance}, and the \themebf{diffusion transformer} (DiT) \citep{dosovitskiy2020image, dit, ma2024sit}.

\subsubsection{Embedding the Conditioning Variables}
\label{subsubsec:encoding_the_conditioning_variables}

\paragraph{Embedding Time.} For simple toy models, concatenating the raw value of $t$ to the input is sufficient to train a reasonably performant network. In practice, the scalar time is often embedded in a higher dimensional space using \themebf{Fourier features}, allowing the model to more faithfully capture high-frequency time dependence \citep{tancik2020fourierfeaturesletnetworks}. Explicitly, the featurization is given by
\begin{equation}
    \text{TimeEmb}(t) = \sqrt{\frac{2}{d}}\begin{bmatrix}
    \cos(2\pi w_1 t) & \cdots & \cos(2\pi w_{d/2} t) & \sin(2\pi w_1 t) & \cdots & \sin(2\pi w_{d/2} t)
    \end{bmatrix}^T,
\end{equation}
where the frequencies $w_i$ are set in the following way
\begin{align}
w_i \;=\; w_{\min}\left(\frac{w_{\max}}{w_{\min}}\right)^{\frac{i-1}{d/2-1}},
\qquad i=1,\ldots,d/2.
\end{align}
This choice of $\text{TimeEmb}$ is a standard choice but this exact form is not strictly necessary. Rather, the above is simply a convenient way of obtaining a normed embedding of dimension $d$, i.e. $\|\text{TimeEmb}(t)\|=1$ (because $\sin^2+\cos^2=1$).

\paragraph{Embedding Class Labels.}  When $y_{\text{raw}} \in \mathcal{Y} \triangleq \{0,\dots, N\}$ is just a class label, then it is often easiest to simply learn a separate embedding vector for each of the $N+1$ possible values of $y_{\text{raw}}$, and set $y$ to this embedding vector. One would consider the parameters of these embeddings to be included in the parameters of $u_t^\theta(x|y)$, and would therefore learn these during training.

\paragraph{Embedding Textual Input} When $y_{\text{raw}}$ is a text-prompt, the situation is more complex, and approaches largely rely on frozen, pre-trained models. Such models are trained to embed a discrete text input into a continuous vector that captures the relevant information. One such model is known as \themebf{CLIP} (Contrastive Language-Image Pre-training). CLIP is trained to learn a shared embedding space for both images and text-prompts, using a training loss designed to encourage image embeddings to be close to their corresponding prompts, while being farther from the embeddings of other images and prompts \cite{clip}. We might therefore take $y = \text{CLIP}(y_{\text{raw}}) \in \mathbb{R}^{d_{\text{CLIP}}}$ to be the embedding produced by a frozen, pre-trained CLIP model. In certain cases, it may be undesirable to compress the entire sequence into a single representation. In this case, one might additionally consider embedding the prompt using a pre-trained transformer so as to obtain a sequence of embeddings. It is also common to combine multiple such pretrained embeddings when conditioning so as to simultaneously reap the benefits of each model \cite{sd3, moviegen}. For our purposes, one can simply assume that after applying such a model the prompt embedding has shape
\begin{align*}
\text{PromptEmbed}(y_{\text{raw}})\in \mathbb{R}^{S\times k}
\end{align*}

\begin{figure}[!t]
\centering
\begin{subfigure}{.5\textwidth}
  \centering
  \includegraphics[width=0.95\linewidth]{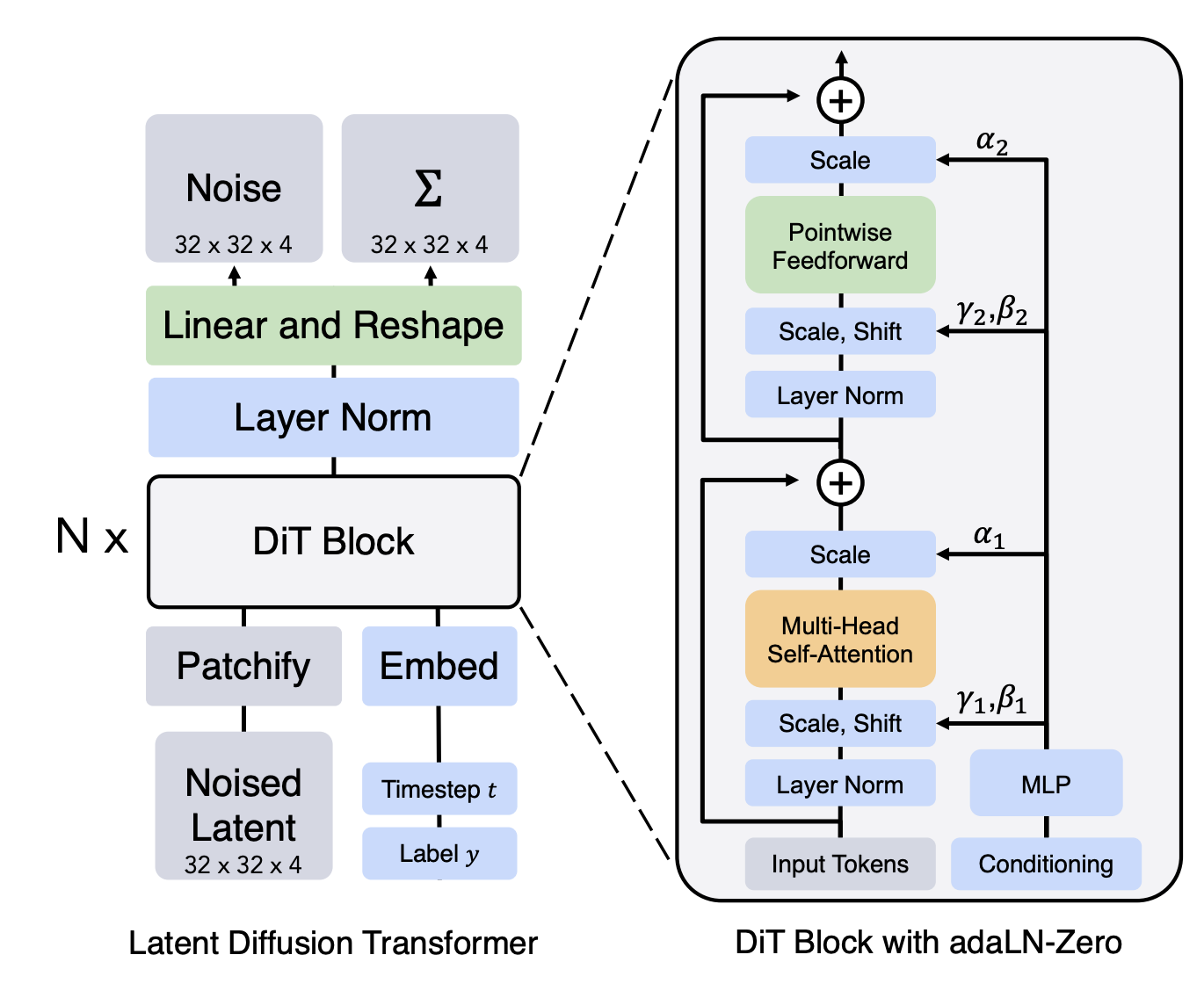}
  \label{fig:sub1}
\end{subfigure}%
\begin{subfigure}{.5\textwidth}
  \centering
  \includegraphics[width=0.95\linewidth]{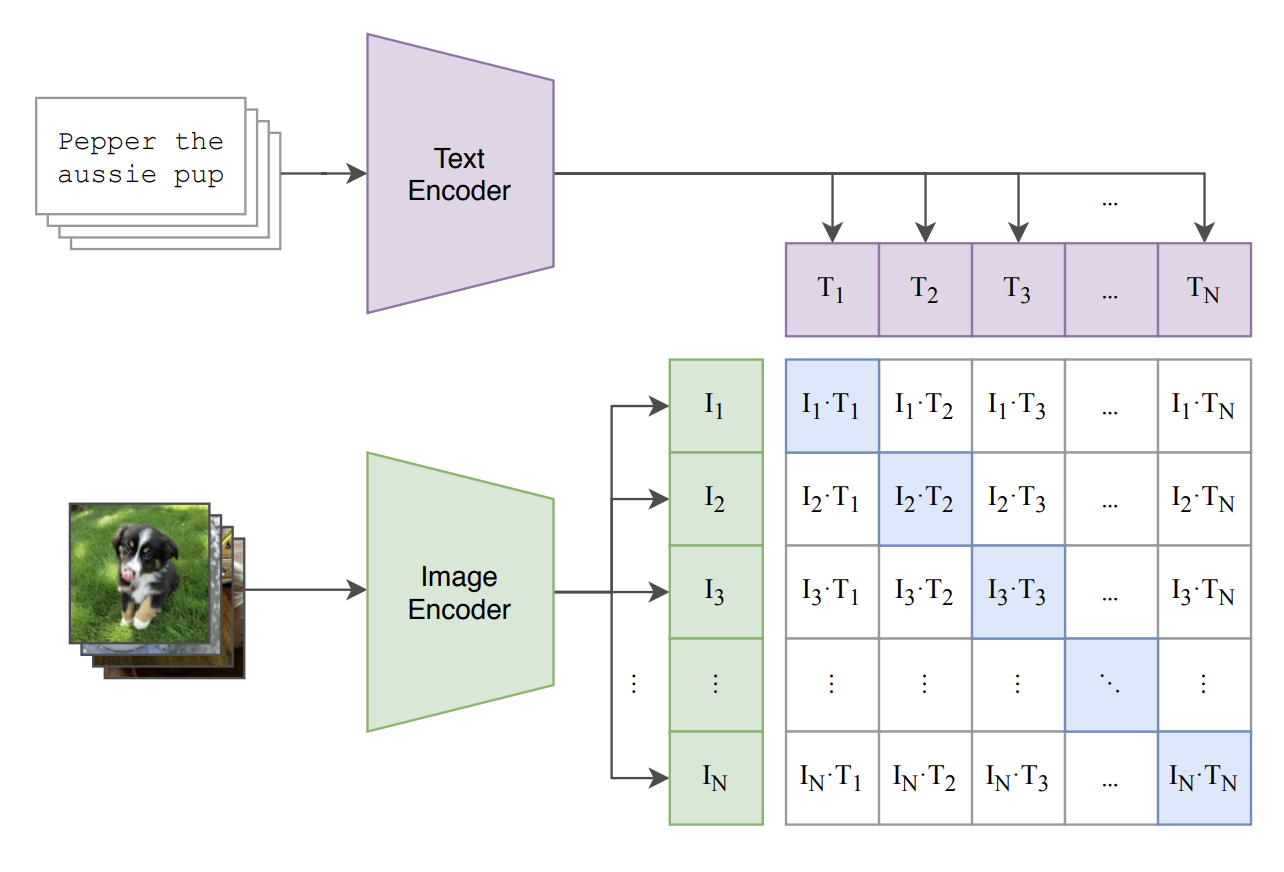}
  \label{fig:sub2}
\end{subfigure}
\caption{Left: An overview of the diffusion transformer architecture, taken from \cite{dit}. Right: A schematic of the contrastive CLIP loss, in which a shared image-text embedding space is learned, taken from \cite{clip}.}
\label{fig:test}
\end{figure}

% \paragraph{Feeding in the Embedding.} Suppose now that we have obtained our embedding vector $y \in \mathbb{R}^{d_y}$. Now what? The answer varies, but usually it is some variant of the following: feed it individually into every sub-component of the architecture for images. Let us briefly describe how this is accomplished in the U-Net implementation used in lab three, as depicted in \cref{fig:unet}. At some intermediate point within the network, we would like to inject information from $y \in \mathbb{R}^{d_y}$ into the current activation $x^{\text{intermediate}}_t \in \mathbb{R}^{C \times H \times W}$. We might do so using the procedure below, given in PyTorch-esque pseudocode.

% \begin{align*}
%     y &= \text{MLP}(y) \in \mathbb{R}^C\quad 
%     && \blacktriangleright\,\,\text{Map $y$ from $\mathbb{R}^{d_y}$ to $\mathbb{R}^C$.}\\
%     y &= \text{reshape}(y) \in \mathbb{R}^{C \times 1 \times 1}\quad 
%     && \blacktriangleright\,\,\text{Reshape $y$ to ``look'' like an image.}\\
%     x^{\text{intermediate}}_t &= \text{broadcast\_add}(x^{\text{intermediate}}_t,y) \in \mathbb{R}^{C \times H \times W}\quad && \blacktriangleright\,\,\text{Add $y$ to $x^{\text{intermediate}}_t$ pointwise.}
% \end{align*}

% One exception to this simple-pointwise conditioning scheme is when we have a sequence of embeddings as produced by some pretrained language model. In this case, we might consider using some sort of cross-attention scheme between our image (suitably patchified) and the tokens of the embedded sequence. We will see multiple examples of this in \cref{sec:large_scale_models}.

\subsubsection{Diffusion Transformers}
\label{subsec:transformers}
Before we dive into the specifics of these architectures, let us recall from the introduction that an image is simply a vector $x \in \mathbb{R}^{C_{\text{image}} \times H \times W}$. Here $C_{\text{image}}$ denotes the number of \themebf{channels} (an RGB image typically would have $C_{\text{input}} = 3$ color channels), and $H$ and $W$ respectively denote the \themebf{height} and \themebf{width} of the image in pixels. One particularly prominent architectural class are so-called \themebf{diffusion transformers} (DiTs), and their variants, which use the \themebf{attention} mechanism to construct the network \cite{attention, dit, ma2024sit}. There are different flavors of diffusion transformers. We explain here a generic design, and note though that specific instantiations of DiTs might differ depending on model and application. For the remainder of this section, we will use $d$ to denote the hidden dimension, $L$ to denote the number of transformer layers, and $h$ to denote the number of heads per layer. Diffusion transformers are based on \themebf{vision transformers} (ViTs), whose main idea is essentially to divide up an image into patches, embed the patches to obtain a sequence of tokens, and process the resulting tokens via standard attention \cite{vit}. A final depatchification operation is applied at the end to recover an image of the correct shape. The initial patchification operation is simply a restructuring of the image tensor $x\in \mathbb{R}^{C\times H\times W}$:
\begin{align*}
    \text{Patchify}(x)\in \mathbb{R}^{N\times C'}
\end{align*}
where $C'=CP^2, N=(H/P)\cdot(W/P)$ for $P$ the patch size. Next, we apply a linear transformation to the output giving us the final patch embedding
\begin{align*}
    \text{PatchEmb}(x)=\text{Patchify}(x)W \in \mathbb{R}^{N\times d}
\end{align*}
where $W\in \mathbb{R}^{C'\times d}$ is a learnable weight matrix. The inputs to the diffusion transformer are then the time embedding, the prompt embedding, and the patchified image tensor given by (see \cref{subsubsec:encoding_the_conditioning_variables}):
\begin{align*}
\tilde{t} &=\text{TimeEmb}(t)\in \mathbb{R}^{d}\\
\tilde{y}&=\text{PromptEmb}(y)\in\mathbb{R}^{S \times d}\\
\tilde{x}_{0}&=\text{PatchEmb}(x)\in\mathbb{R}^{N\times d}
\end{align*}
Note that all elements have now the desired hidden dimension of the transformer. The diffusion transformer then iteratively updates $\tilde{z}_i$ via for $i=0,\cdots,L-1$ via transformer layers in a \themebf{DitBlock} (see  \cref{remark:dit_layer} for details):
\begin{align}
\tilde{x}_{i+1}=\text{DiTBlock}(\tilde{x}_i,\tilde{t},\tilde{y})\in\mathbb{R}^{N\times d}\quad (i=0,\dots,L-1).
\end{align}
where $N$ is the number of layers.
Finally, a final operation applies a depatchification operation which maps the DiT output back to the desired output shape:
\begin{align*}
u=\text{Depatchify}(\tilde{x}_N\tilde{W})\in \mathbb{R}^{C\times H\times W},
\end{align*}
where $\tilde{W}\in \mathbb{R}^{d\times C'}$. The final tensor $u$ then serves as the output of the model and the predicted velocity $u_t^\theta(x|y)$.
\begin{remarkbox}[DiT Block]
\label{remark:dit_layer}
For completeness, we present a brief mathematical description of a single DiT layer. While we attempt to include enough detail to allow for a general understanding of the DiT model family, we remind the reader that these choose to emphasize key algorithmic choices rather than architectural details. Now, let $x\in\mathbb{R}^{N\times d}$ denote the current sequence of patch tokens (here $x=\tilde x_i$),
and let $y\in\mathbb{R}^{S \times d}$ denote the embedded guiding variable (here $y=\tilde y$).
Then, a typical DiT block updates $x$ using (i) self-attention on patches, (ii) cross-attention to the prompt,
and (iii) time conditioning via adaptive normalization (AdaLN).

\paragraph{Scaled Dot Product Attention.}
Given queries $Q\in\mathbb{R}^{N\times d_h}$, keys $K\in\mathbb{R}^{M\times d_h}$, and values $V\in\mathbb{R}^{M\times d_h}$,
\[
\mathrm{Attn}(Q,K,V)
\;=\;
\mathrm{softmax}\!\left(\frac{QK^\top}{\sqrt{d_h}}\right)V
\;\in\;\mathbb{R}^{N\times d_h},
\]
where the softmax is applied row-wise.

\paragraph{Multi-Head Attention.}
Let $h$ denote the number of heads and $d_h=\frac{d}{h}$ the per-head dimension.
For each head $h\in\{1,\dots,n_{\text{heads}}\}$, learn projection matrices
$W_Q^{(h)},W_K^{(h)},W_V^{(h)}\in\mathbb{R}^{k\times d_h}$.
Define
\[
\text{head}_h(x,z)
\;=\;
\mathrm{Attn}\!\big(xW_Q^{(h)},\,zW_K^{(h)},\,zW_V^{(h)}\big),
\]
where the \emph{source sequence} $z$ is either
\[
z=x \quad \text{(self-attention on patches)},\qquad
z=y \quad \text{(cross-attention to the prompt)}.
\]
Concatenate heads and apply an output projection $W_O\in\mathbb{R}^{d\times d}$:
\[
\mathrm{MultiHeadattention}(x,z)
\;=\;
\mathrm{Concat}\big(\text{head}_1(x,z),\dots,\text{head}_{h}(x,z)\big)\,W_O
\;\in\;\mathbb{R}^{N\times d}.
\]

\paragraph{Time Conditioning via Adaptive Normalization.}
Let $\tilde t\in\mathbb{R}^d$ be the timestep embedding. A standard choice in DiTs is to use $\tilde t$
to produce per-channel scale/shift parameters that modulate normalized activations \citep{perez2018film}.
Concretely, let $g:\mathbb{R}^d\to\mathbb{R}^{2d}$ be an MLP and set
\[
(\gamma,\beta) = g(\tilde t),
\]
where $\gamma,\beta\in\mathbb{R}^{d}$ (or, depending on the implementation, separate $(\gamma,\beta)$ pairs for different
sub-layers such as attention and MLP). Given a token matrix $x\in\mathbb{R}^{N\times d}$ and a normalization operator
$\mathrm{Norm}(\cdot)$ (e.g.\ LayerNorm), define the modulated normalization
\[
\mathrm{AdaNorm}_{\tilde t}(x)
\;=\;
\bigl(1+\gamma\bigr)\odot \mathrm{Norm}(H) \;+\; \beta,
\]
where $\odot$ denotes elementwise multiplication with broadcasting over the token dimension.

\paragraph{Putting It Together.}
The combined operation, and thus the DitBlock, is given by.
\begin{align*}
x &\gets x + g_\text{self}(\tilde{t})\odot \mathrm{MultiHeadattention}\!\big(\mathrm{AdaNorm}_{\tilde t}(x),\,\mathrm{AdaNorm}_{\tilde t}(x)\big)\\
x &\gets x + g_\text{cross}(\tilde{t})\mathrm{MultiHeadattention}\!\big(\mathrm{AdaNorm}_{\tilde t}(x),\,y\big)\\
x &\gets x + g_\text{MLP}(\tilde{t})\mathrm{MLP}\!\big(\mathrm{AdaNorm}_{\tilde t}(x)\big),
\end{align*}
where the MLP is a position-wise feed-forward network, and the $g_{\cdots}$ are learnable gating parameters. The output $x\in \RR^{N\times d}$ becomes the next-layer
patch-token sequence (in our notation, $\tilde x_{i+1}$). Finally, we note that class-conditioned DiT's, such as the one implemented in the lab, are typically simpler and eschew the cross attention layer in favor of a time and class-based AdaNorm conditioning.
\end{remarkbox}

\subsubsection{U-Net}
\begin{figure}
    \centering
    \includegraphics[width=\textwidth]{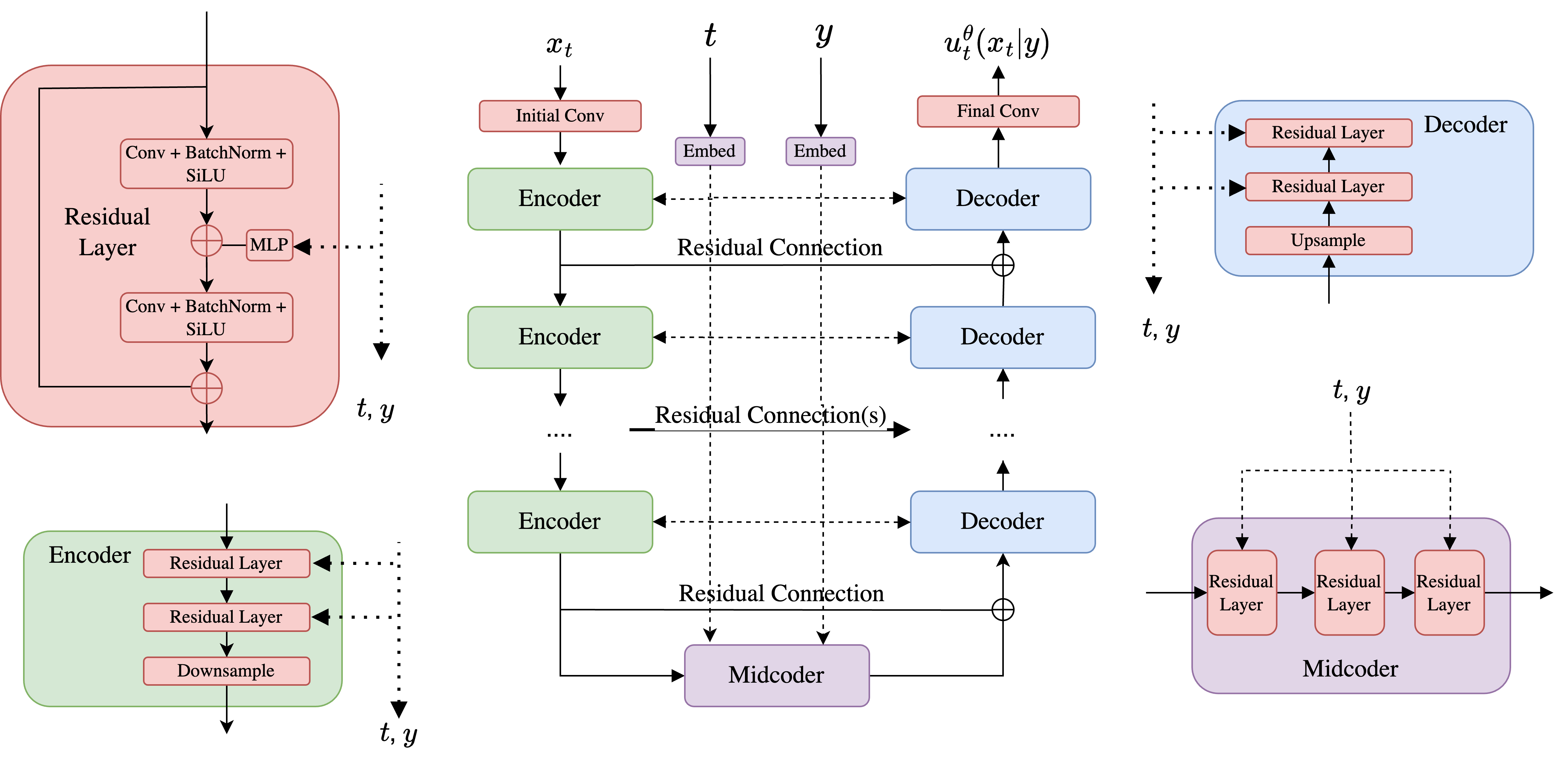}
    \caption{A simplified U-Net architecture (an architecture like this was used in lab 03 of the 2025 version of this course).}
    \label{fig:unet}
\end{figure}

The \themebf{U-Net} architecture \citep{ronneberger2015u} is an alternative architecture to the DiT architecture and is a specific type of convolutional neural network. Originally designed for image segmentation, its crucial feature is that both its input and its output have the shape of images (possibly with a different number of channels). This makes it ideal for parameterizing a vector field $x\mapsto u_t^\theta(x|y)$, as for fixed $y,t$ its input has the shape of an image and its output does, too. Accordingly, U-Nets have seen widespread use across much of the early literature on diffusion models \citep{ho2020denoising, unet_cite_1, classifier_guidance}. A U-Net consists of a series of \themebf{encoders} $\mathcal{E}_i$, and a corresponding sequence of \themebf{decoders} $\mathcal{D}_i$, along with a latent processing block in between, which we shall refer to as a \themebf{midcoder}.\footnote{Midcoder is a completely non-standard term used here to refer to the bottom-most part of the U-Net stack, and in analogy with the encoder and decoder.} For sake of example, let us walk through the path taken by an image $x_t \in \mathbb{R}^{3 \times 256 \times 256}$ (we have taken $(C_{\text{input}}, H, W) = (3, 256, 256)$) as it is processed by the U-Net:
\begin{alignat*}{3}
    x^{\text{input}}_t &\in \mathbb{R}^{3 \times 256 \times 256} \quad  
    && \blacktriangleright\,\,\text{Input to the U-Net.}\\
    x^{\text{latent}}_t = \mathcal{E}(x^{\text{input}}_t) &\in \mathbb{R}^{512 \times 32 \times 32} \quad && \blacktriangleright\,\,\text{Pass through encoders to obtain latent.}\\
    x^{\text{latent}}_t = \mathcal{M}(x^{\text{latent}}_t) &\in \mathbb{R}^{512 \times 32 \times 32} \quad && \blacktriangleright\,\,\text{Pass latent through midcoder.}\\
    x^{\text{output}}_t = \mathcal{D}(x^{\text{latent}}_t) &\in \mathbb{R}^{3 \times 256 \times 256} \quad && \blacktriangleright\,\,\text{Pass through decoders to obtain output.}
\end{alignat*}
Notice that as the input passes through the encoders, the number of channels in its representation increases, while the height and width of the images are decreased. Both the encoder and the decoder usually consist of a series of convolutional layers (with activation functions, pooling operations, etc. in between). Not shown above are two points: First, the input $x^{\text{input}}_t\in \mathbb{R}^{3 \times 256 \times 256}$ is often fed into an initial pre-encoding block to increase the number of channels before being fed into the first encoder block. Second, the encoders and decoders are often connected by \themebf{residual connections}. The complete picture is shown in \cref{fig:unet}.
At a high level, most U-Nets involve some variant of what is described above. However, certain of the design choices described above may well differ from various implementations in practice. In particular, we opt above for a purely-convolutional architecture whereas it is common to include attention layers as well throughout the encoders and decoders. The U-Net derives its name from the ``U''-like shape formed by its encoders and decoders (see \cref{fig:unet}).

% In lab three, you'll have a chance to implement your own diffusion transformer from scratch. For the case of MNIST, in which case our guiding variable lies in the discrete set $\{0, \dots 0, \varnothing\}$, the diffusion transformer implementation from the lab proceeds as follows.

\subsection{Working in Latent Space: (Variational) Autoencoders}
Thus far, we have operated in the data space $\RR^d$. However, the cost of modeling directly within such a space quickly becomes prohibitively expensive as one scales to increasingly higher resolution images. For example, a $1024 \times 1024$ image with three RGB color channels corresponds to a total dimension of $d=H\cdot W\cdot 3\approx3*10^6$! Note that the dimension increases further for videos as everything scales with the number of frames $T$. As you can imagine, training over such a space quickly becomes infeasible. Unlike image classification, whose low-dimensional outputs allow for narrowing convolutional stacks, our flow-based modeling approach requires that our output $u_t^\theta(x)\in\mathbb{R}^d$ be just as large as our input. The important question thus becomes: \textbf{How can we model high-dimensional images within a reasonable memory and computation budget?}

\subsubsection{Standard Autoencoders}
A natural answer to this question lies in \themebf{compression}: perhaps the actual space of images, for example, lies near a much lower-dimensional manifold of the high dimensional image space. More concretely, we might consider an \themebf{encoder} $\mu_{\phi}: \mathbb{R}^d \to \mathbb{R}^k$, together with some \themebf{decoder} $\mu_{\theta}: \mathbb{R}^k \to \mathbb{R}^d$, which together map raw images $x \in \RR^d$ to and from latents $z\in\RR^k$, respectively. The dimension $k$ is typically chosen to be much smaller than $d$. For images, in which, for example, $d = 3 \times 1024 \times 1024$, it is not uncommon to downsample to obtain e.g., $k = 3 \times \tfrac{1024}{16} \times \tfrac{1024}{16}$. Together, $\mu_{\phi}$ and $\mu_{\theta}$ are referred to as an \themebf{autoencoder}. Ideally, $\mu_{\phi}$ and $\mu_{\theta}$ are chosen so as to achieve high reconstruction quality, or in other words, so that $\mu_{\theta}(\mu_{\phi}(x))$ resembles $x$ on average. Accordingly, autoencoders are usually trained with the \themebf{reconstruction loss}
\begin{align*}
\mathcal{L}_{\text{Recon}}(\phi,\theta)=&\mathbb{E}_{x\sim \pdata}\left[\|\mu_\theta(\mu_\phi(x))-x\|^2\right].
\end{align*}
which measures the squared error between the original data point $x$ and the reconstructed one $\mu_\theta(\mu_\phi(x))$.

\paragraph{Amenability to Generative Modeling.} Unfortunately, the reconstruction loss above is not enough to train a ``good'' autoencoder. Recall that our eventual goal is to train a generative model in the latent space, and targeting the latent distribution $p_{\text{latent}}(z)$ given by $z = \mu_{\phi}(x), x \sim \pdata$. A generative model for $\pdata(x)$ is then realized by passing the output of our latent generative model through the decoder $\mu_{\theta}$. A subtle issue arises with autoencoders as we have currently formulated them in that we have little to no control over $p_{\text{latent}}(z)$, and thus essentially no guarantee that $p_{\text{latent}}(z)$ is even well-behaved enough to be amenable to training such a generative model (i.e., nice, simple, Gaussian-like). While transforming our data in latent space might have compressed it, we might have transformed the data distribution $\pdata$ into a very hard-to-learn distribution $\platent$. Therefore, the question is: how can we make sure that the latent distribution $\platent$ is still well-behaved and easy-to-learn? To allow for more explicit regularization of the latent distribution, we will now recast the concept of autoencoder in a more general probabilistic framework leading to the concept of a variational autoencoder.

\subsubsection{Variational Autoencoders} A \themebf{variational autoencoder} (VAE) is obtained from our (deterministic) standard autoencoder formulation by relaxing the constraint that the encoder and decoder are deterministic functions. In particular, let us consider an encoder $q_\phi(z|x)$ with parameters $\phi$, and a decoder $p_\theta(x|z)$ with parameters $\theta$. The most common choice is to take 
\begin{align}
\label{eq:normal_vae}
q_\phi(z|x)=\mathcal{N}(z;\mu_\phi(x),\diag(\sigma_\phi^2(x))),\quad p_\theta(x|z)=\mathcal{N}(x;\mu_\theta(z),\sigma_\theta^2(z)I_d)
\end{align}
where $\mu_\phi(x)\in \mathbb{R}^k$, $\sigma_\phi^2(x)\in\mathbb{R}_{\geq 0}^k$, $\mu_\theta(z)\in \mathbb{R}^d$, and $\sigma_\theta^2(z)\in\mathbb{R}_{\geq 0}$ are parameterized as neural networks and $\diag$ denotes the diagonal matrix. To encode or decode a variable, we sample 
\begin{align*}
z &\sim q_\phi(\cdot|x) &\quad& (\text{encode}) \\
x &\sim p_\theta(\cdot|z) &\quad& (\text{decode})
\end{align*}
Finally, we note that when $\sigma_\phi(x) = 0$ and $\sigma_\theta(x) = 0$ always, we recover a standard autoencoder. Let us examine what a reconstruction loss looks like. A natural objective is the following:
\begin{align}
\mathcal{L}_{\text{VAE-Recon}}(\phi,\theta)=&-\EE_{x \sim \pdata(x),z\sim q_\phi(\cdot|x)} \left[\log p_\theta(x|z)\right]
\end{align}
Note the two changes: Instead of a deterministic encoding, we now sample $z\sim q_\phi(z|x)$. Further, we now take the negative log-likelihood of $x$ under decoding, i.e. the loss effectively asks: how likely would our original data point $x$ be if we  encoded and decoded it - and we take all possible decodings/encodings into account as things have become random now. For the Gaussian case, this reconstruction loss becomes:
\begin{align}
\mathcal{L}_{\text{VAE-Recon}}(\phi,\theta)=&\EE_{x \sim \pdata(x),z\sim q_\phi(z|x)} \left[\frac{1}{2\sigma^2_\theta(z)}\|x-\mu_\theta(z)\|^2+\frac{d}{2}\log \sigma_{\theta}^2(z)\right]+\text{const}
\end{align}
where we used the density of the normal distribution (see \cref{e:gaussian}) Hence, the VAE reconstruction loss is not that different from the standard AE reconstruction loss, we simply have to take into account all possible encodings $z\sim q_\phi(\cdot|x)$. The second term depending on the decoder variance controls the tradeoff between reconstruction accuracy and predictive uncertainty. Many implementations, including that in the lab, fix $\sigma_\phi(x)$ and $\sigma_\theta(z)$ to learned scalar constants (that is, independent of $x$ and $z$, respectively), thereby avoiding pathological behavior and numerical stability when learning variances. Therefore, the VAE reconstruction loss in this case then becomes basically the standard autoencoder reconstruction loss up to stochasticity in the encoding and constants:
\begin{align}
\mathcal{L}_{\text{VAE-Recon}}(\phi,\theta)=&\EE_{x \sim \pdata(x),z\sim q_\phi(z|x)} \left[\frac{1}{2\sigma_\theta^2}\|x-\mu_\theta(z)\|^2\right]+\text{const}
\end{align}

Let us now revisit our goal: We want to create an encoding of our data distribution $\pdata(x)$ such that after mapping it into a latent space, the distribution becomes ``nice'' or easy-to-learn. Toward this end, let us now introduce a \themebf{prior distribution} $\prior(z)$ over latents $z$. For our purposes, we will take $\prior=\mathcal{N}(0,I_k)$ to be an isotropic Gaussian. This choice of prior distribution $\prior$ effectively represents the ``ideal'' case for what the latent distribution should look like. A normal distribution would be very easy to learn, and would therefore satisfy our goal of obtaining a ``trainable'' latent distribution. The big idea is thus to regularize our encoder so as to ensure that the encoded data distribution is as close as possible to the $\prior$, which we accomplish via the auxiliary loss
\begin{align}
\mathcal{L}_{\text{VAE-Prior}}(\phi)=&\EE_{x \sim \pdata(x)} \left[\dkl{q_\phi(\cdot|x)}{\prior}\right],
\end{align}
and where $D_{KL}$ is the  \themebf{Kullback-Leibler (KL) divergence}. The KL-divergence is a fundamental way of measuring how different two probability distributions are. Explaining it in detail would go beyond the scope of this work but we give a brief background in \cref{remark:kl_divergence} as a reminder for the reader.  The loss $\mathcal{L}_{\text{VAE-Prior}}$ defined here now is very intuitive: We want that the encoding distributions looks like a Gaussian distribution for any data point $x$. If we do this for all $x$, it is natural to expect that then our latent distribution will look a Gaussian as well.
\begin{remarkbox}[Background on KL-divergence]
\label{remark:kl_divergence}
For two probability densities $q,p$, the \themebf{Kullback-Leibler divergence} (KL-divergence) is defined as
\begin{align*}
    \dkl{q(x)}{p(x)}=\int q(x) \log \frac{q(x)}{p(x)}=\mathbb{E}_{X\sim q}\left[\log \frac{q(X)}{p(X)}\right].
\end{align*}
The KL divergence is a standard measure of dissimilarity between distributions. In particular, the KL divergence satisfies the following useful properties:
\begin{align}
\label{eq:kl_non_negative}
\dkl{q(x)}{p(x)}&\geq 0,\\
\dkl{q(x)}{p(x)}&=0\quad \Leftrightarrow \quad q=p.
\end{align}
i.e. it is always non-negative and it is zero if and only the two probability distributions coincide.
\end{remarkbox}

To define the loss function for a variational autoencoder, we can now combine both the reconstruction and the prior loss with a parameter weight $\beta\geq 0$ to \themebf{VAE training objective} given by
\begin{align}
\mathcal{L}_{\text{VAE}}(\phi,\theta)&=\mathcal{L}_{\text{VAE-Recon}}(\phi,\theta)+\beta\mathcal{L}_{\text{VAE-Prior}}(\phi)\\
&=-\EE_{x \sim \pdata(x),z\sim q_\phi(z|x)} \left[\log p_\theta(x\mid z)\right]+\beta\EE_{x \sim \pdata(x)} \left[D_{KL}(q_\phi(\cdot|x)||\prior)\right]
\end{align}
where the first summand enforces that latent variables can be efficiently decoded back to data and the second summand enforces that our latent distribution is close to being a Gaussian. The parameter $\beta$ controls the strength of each. To make this loss more specific, let us derive the KL divergence for the Gaussian case:
\begin{examplebox}[KL Divergence Between Isotropic Gaussians]
\label{example:kl_divergence_between_isotropic_normals}
Let $q(x)=\mathcal N(x;\mu_q,\diag(\sigma_q^2))$ and $p(x)=\mathcal N(x;\mu_p,\diag(\sigma_p^2))$ be Gaussians with diagonal covariance matrices, with $\sigma_q,\sigma_{p}\in\mathbb{R}_{\geq 0}^d$, and where $x \in \RR^d$. Then
\begin{align}
\dkl{q}{p}
=\frac12\left(
\mathcal{K}\left(\frac{\sigma_{q}^2}{\sigma_{p}^2}\right)+ \frac{\|\mu_q-\mu_p\|^2}{\sigma_p^2}
\right),\quad \text{where }\mathcal{K}(\alpha)=\sum\limits_{i=1}^{d}\alpha_i-\log \alpha_i-1.
\label{eq:kl_isotropic_gaussians}
\end{align}
The expression above is intuitive: If the mean and variances coincide, that then $\dkl{q}{p}=0$. Further, it increases with the squared error $\|\mu_{q}-\mu_{p}\|^2$ between the mean vectors. Finally, the function $\mathcal{K}(\alpha)$ has a unique minimum at $\alpha=1$ so that $\dkl{q}{p}$ is minimized when $\sigma_{q}=\sigma_{p}$.
\begin{proof}
We do the proof for $d=1$ (proof is analogous for $d>1$ by summing up each dimension). Given the density of the normal distribution, we know that (see \cref{e:gaussian}):
\[
\log q(x)= -\frac {1}{2}\log(2\pi\sigma_q^2)-\frac{1}{2\sigma_q^2}\|x-\mu_q\|^2,
\qquad
\log p(x)= -\frac 12\log(2\pi\sigma_p^2)-\frac{1}{2\sigma_p^2}\|x-\mu_p\|^2
\]
Then
\begin{align}
\label{eq:kl_normal_distributions}
D_{\mathrm{KL}}(q\|p)
&=\E_{x\sim q}\big[\log q(x)-\log p(x)\big]=\frac 12\log\frac{\sigma_p^2}{\sigma_q^2}
+\frac{1}{2\sigma_p^2}\E_q\!\left[\|x-\mu_p\|^2\right]
-\frac{1}{2\sigma_q^2}\E_q\!\left[\|x-\mu_q\|^2\right].
\end{align}
For $x\sim \mathcal N(\mu_q,\sigma_q^2 I)$ we have
\[
\E_q\!\left[\|x-\mu_q\|^2\right]=\mathrm{tr}(\sigma_q^2 I)=\sigma_q^2.
\]
Combining this with the fact that $x-\mu_p=(x-\mu_q)+(\mu_q-\mu_p)$, and $\E_q[x-\mu_q]=0$, we obtain
\[
\E_q\!\left[\|x-\mu_p\|^2\right]
=\E_q\!\left[\|x-\mu_q\|^2\right]+\|\mu_q-\mu_p\|^2
=\sigma_q^2+\|\mu_q-\mu_p\|^2.
\]
Plugging these into \cref{eq:kl_normal_distributions} yields \eqref{eq:kl_isotropic_gaussians}.
\end{proof}
\end{examplebox}
Let us now assume a Gaussian shape of the encoder. Then we obtain: 
\begin{align}
\mathcal{L}_{\text{VAE-Prior}}(\phi)=&\EE_{x \sim \pdata(x)} \left[\dkl{q_\phi(\cdot|x)}{\mathcal{N}(0,I_k)}\right]=\mathbb{E}\left[\frac{1}{2}
\mathcal{K}\left(\sigma_{\phi}^2(x)\right)+ \frac{1}{2} \|\mu_\phi(x)\|^2\right]
\end{align}
This loss is intuitive: The mean $\mu_\phi(x)$ is penalized for being different from zero and the variance penalized for being different from $1$. As a total loss for the VAE, we obtain 
\begin{equation}
\begin{aligned}
&\mathcal{L}_{\text{VAE}}(\phi,\theta)\\
&=\mathcal{L}_{\text{VAE-Recon}}(\phi,\theta)+\beta\mathcal{L}_{\text{VAE-Prior}}(\phi)\\
&=\EE_{x \sim \pdata(x),z\sim q_\phi(z|x)} \left[\underbrace{\frac{1}{2\sigma^2_\theta(z)}\|x-\mu_\theta(z)\|^2}_{\text{recon. error}}+\underbrace{\frac{d}{2}\log \sigma_{\theta}^2(z)
}_{\text{decoder confidence}}+\underbrace{\frac{\beta}{2}
\mathcal{K}\left(\sigma_{\phi}^2(x)\right)}_{\text{make latent variance}=1}+ \underbrace{\frac{\beta}{2} \|\mu_\phi(x)\|^2}_{\text{make latent mean}=0}
\right]
\end{aligned}
\label{eq:total_vae_loss}
\end{equation}
The four terms of the above loss function are very intuitive: The first term is simply a reconstruction error. The second error describes the decoder's uncertainty: smaller variance makes the decoder more ``confident'' but also penalizes reconstruction errors more strongly. Further, we want to make the latent variance $1$ and the mean to be $0$ - to enforce that the distribution in latent is close to being Gaussian.

\paragraph{Training a VAE.} It remains to discuss how we would minimize the VAE loss $\mathcal{L}_{\text{VAE}}(\phi,\theta)$. The problem with the loss is that so far, the distribution we take the expected value over ($q_\phi(z|x)$) still depends on the parameter $\phi$. However, we can apply the so-called \themebf{reparameterization trick} to rewrite it.  Specifically, for 
\begin{align*}
q_\phi(z|x)=\mathcal{N}(z;\mu_\phi(x),\sigma_\phi^2(x)I_k)
\end{align*}
we can obtain samples via
\begin{align*}
\epsilon\sim\mathcal{N}(0,I_k),\quad z=\mu_\phi(x)+\sigma_\phi(x)\epsilon \quad \Rightarrow\quad z\sim q_\phi(\cdot|x)
\end{align*}
Note that in this equation, the only source of noise/stochasticity is from $\epsilon$ whose distribution is independent of $\phi$. Therefore, we can rewrite the loss as:
\begin{align*}
\mathcal{L}_{\text{VAE}}(\phi,\theta)=&\EE_{x \sim \pdata(x),\epsilon\sim\mathcal{N}(0,I_k)} \left[\frac{1}{2\sigma^2_\theta(z)}\|x-\mu_\theta(\mu_\phi(x)+\sigma_\phi(x)\epsilon)\|^2+\frac{d}{2}\log \sigma_{\theta}^2(z)
+\frac{\beta}{2}
\mathcal{K}\left(\sigma_{\phi}^2(x)\right)+ \frac{\beta}{2} \|\mu_\phi(x)\|^2
\right]
\end{align*}
After reparameterization, the randomness comes only from $\epsilon\sim\mathcal{N}(0,I_k)$, whose distribution does not depend on $\phi$. Therefore, we can minimize this loss with the standard tools of deep learning. To simplify things even further, we can set $\sigma_{\theta}^2(z)=\sigma^2$ constant everywhere again and obtain:
\begin{align*}
\mathcal{L}_{\text{VAE}}(\phi,\theta)=&\EE_{x \sim \pdata(x),\epsilon\sim\mathcal{N}(0,I_k)} \left[\frac{1}{2\sigma^2}\|x-\mu_\theta(\mu_\phi(x)+\sigma_\phi(x)\epsilon)\|^2
+\frac{\beta}{2}
\mathcal{K}\left(\sigma_{\phi}^2(x)\right)+ \frac{\beta}{2} \|\mu_\phi(x)\|^2
\right]
\end{align*}
In \cref{alg:vae_training}, we summarize the training procedure of the VAE.
\begin{algorithm}[ht]
\caption{$\beta$-VAE Training Procedure (Gaussian decoder with fixed variance $p_\theta(x|z)=\mathcal{N}(x;\mu_\theta(z),\tilde{\sigma}^2 I_d)$)}
\label{alg:vae_training}
\begin{algorithmic}[1]
\REQUIRE Dataset of samples $x\sim \pdata$, encoder networks $(\mu_\phi(x), \log\sigma_\phi^2(x))$, decoder network $\mu_\theta(z)$, latent dim $k$, constants $\beta\ge 0$, $\sigma^2>0$
\FOR{each mini-batch $\{x_i\}_{i=1}^B$}
    \STATE Encode each $x_i$: $\mu_i \gets \mu_\phi(x_i)$,\;\; $\log\sigma_i^2 \gets \log\sigma_\phi^2(x_i)$
    \STATE Sample noise $\epsilon_i \sim \mathcal{N}(0,I_k)$
    \STATE Reparameterize: $z_i \gets \mu_i + \sigma_i \odot \epsilon_i$ \hfill (where $\sigma_i=\exp(\tfrac12 \log\sigma_i^2)$)
    \STATE Decode mean: $\hat{x}_i \gets \mu_\theta(z_i)$
    \STATE Reconstruction loss:
    \[
        \mathcal{L}_{\text{recon}} \gets \frac{1}{B}\sum_{i=1}^B \frac{1}{2\tilde{\sigma}^2}\,\|x_i-\hat{x}_i\|^2
    \]
    \STATE KL loss to the prior $\prior(z)=\mathcal{N}(0,I_k)$:
    \[
        \mathcal{L}_{\text{KL}} \gets \frac{1}{B}\sum_{i=1}^B \frac12 \sum_{j=1}^k \left(\mu_{i,j}^2 + \sigma_{i,j}^2 - \log \sigma_{i,j}^2 - 1\right)
    \]
    \STATE Total loss: $\mathcal{L}\gets \mathcal{L}_{\text{recon}} + \beta\,\mathcal{L}_{\text{KL}}$
    \STATE Update $(\phi,\theta)\gets \text{grad\_update}(\mathcal{L})$
\ENDFOR
\end{algorithmic}
\end{algorithm}

% \begin{algorithm}[h]
% \caption{$\beta$-VAE Training Procedure (Gaussian decoder with fixed variance $p_\theta(x|z)=\mathcal{N}(x;\mu_\theta(z),\sigma^2 I_d)$)}
% \label{alg:vae_training}
% \begin{algorithmic}[1]
% \REQUIRE Dataset of samples $x\sim \pdata$, encoder networks $(\mu_\phi(x), \log\sigma_\phi^2(x))$, decoder network $\mu_\theta(z)$, latent dim $k$, constants $\beta\ge 0$, $\sigma^2>0$
% \FOR{each mini-batch $\{x_i\}_{i=1}^B$}
%     \STATE Encode each $x_i$: $\mu_i \gets \mu_\phi(x_i)$,\;\; $\log\sigma_i^2 \gets \log\sigma_\phi^2(x_i)$
%     \STATE Sample noise $\epsilon_i \sim \mathcal{N}(0,I_k)$
%     \STATE Reparameterize: $z_i \gets \mu_i + \sigma_i \odot \epsilon_i$ \hfill (where $\sigma_i=\exp(\tfrac12 \log\sigma_i^2)$)
%     \STATE Decode mean: $\hat{x}_i \gets \mu_\theta(z_i)$
%     \STATE Reconstruction loss:
%     \[
%         \mathcal{L}_{\text{recon}} \gets \frac{1}{B}\sum_{i=1}^B \frac{1}{2\sigma^2}\,\|x_i-\hat{x}_i\|^2
%     \]
%     \STATE KL loss to the prior $\prior(z)=\mathcal{N}(0,I_k)$:
%     \[
%         \mathcal{L}_{\text{KL}} \gets \frac{1}{B}\sum_{i=1}^B \frac12 \sum_{j=1}^k \left(\mu_{i,j}^2 + \sigma_{i,j}^2 - \log \sigma_{i,j}^2 - 1\right)
%     \]
%     \STATE Total loss: $\mathcal{L}\gets \mathcal{L}_{\text{recon}} + \beta\,\mathcal{L}_{\text{KL}}$
%     \STATE Update $(\phi,\theta)\gets \text{grad\_update}(\mathcal{L})$
% \ENDFOR
% \end{algorithmic}
% \end{algorithm}

\paragraph{Practical remarks.} The construction we developed here show the principles of autoencoder design. Of course, in practice, people might add more loss terms or other constraints. Therefore, we finally add a practical remarks about autoencoders:
\begin{enumerate}
\item \textbf{Choosing $\beta$ (and KL warm-up).}
Large $\beta$ enforces latents closer to the prior but can hurt reconstructions and may trigger \emph{posterior collapse} (the encoder ignores $x$ and outputs $q_\phi(z|x)\approx \mathcal{N}(0,I_k)$).
A common stabilization is \emph{KL warm-up}: start with $\beta=0$ and gradually increase it to a target value over the first epochs. However, in all modern autoencoders, the $\beta$ value is very small, i.e. $\beta<<1$.
\item \textbf{Decoder variance.}
Learning a Gaussian decoder variance $\sigma_\theta^2$ can be numerically delicate and may lead to degenerate solutions unless regularized.
For stability, many implementations fix $p_\theta(x|z)=\mathcal{N}(x;\mu_\theta(z),\sigma^2 I_d)$ with constant $\sigma^2$, which makes the reconstruction term proportional to mean-squared error (up to constants).
\item \textbf{Reconstruction losses beyond pixel MSE.}
For images, a pixelwise Gaussian likelihood (mean squared error) often yields overly smooth reconstructions.
In practice, people add \emph{perceptual losses} (feature-space losses using a pretrained network) to improve sharpness and semantic fidelity.
\item \textbf{Adversarial and hybrid objectives.}
To further improve visual realism, one can combine the VAE objective with an \emph{adversarial loss} (VAE-GAN style), using a discriminator on decoded samples.
This typically sharpens outputs but introduces additional optimization instability and extra hyperparameters.
\end{enumerate}

\begin{remarkbox}[Working in Latent Space]
To train a latent generative model, we simply follow the existing training recipe, but work directly in the \themebf{latent space}. At training time, we draw samples from $q_\phi(z|x)$ with $x \sim \pdata$, and at inference time, we sample $z$ from the latent diffusion or flow model, and then decode using $x = \mu_\text{mean}(z)$ (note that we take the mean rather than a random sample to avoid noise-induced artifacts). Intuitively, a well-trained autoencoder can be thought of as filtering out high-frequency or otherwise semantically meaningless details, allowing the generative model to ``focus'' on important, perceptually relevant features \cite{latent_diffusion}. At the time of the writing of this document, nearly all state-of-the-art approaches to image and video generation follow the so-called \themebf{latent diffusion} paradigm involving training a flow or diffusion model within the latent space of an autoencoder \citep{latent_diffusion,vahdat2021score}. However, it is important to note: one also needs to train the autoencoder before training the diffusion models. Crucially, performance now depends also on how good the autoencoder compresses images into latent space and recovers aesthetically pleasing images.
\end{remarkbox}
We provide additional discussion on VAEs in \cref{sec:vae_appdx}.

\subsection{Case Study: Stable Diffusion 3 and Meta Movie Gen}
\label{sec:large_scale_models}
We conclude this section by briefly examining two large-scale generative models: \themeit{Stable Diffusion 3} for image generation and Meta's \themeit{Movie Gen Video} for video generation \citep{sd3, moviegen}. As you will see, these models use the techniques we have described in this work along with additional architectural enhancements to both scale and accommodate richly structured conditioning modalities, such as text-based input.

\subsubsection{Stable Diffusion 3}

Stable Diffusion is a series of state-of-the-art image generation models. These models were among the first to use large-scale latent diffusion models for image generation. If you have not done so, we highly recommend testing it for yourself online (\url{https://stability.ai/news/stable-diffusion-3}).\\

%\paragraph{Training Objective.} 
Stable Diffusion 3 uses the same conditional flow matching objective that we study in this work (see \cref{alg:training_score_matching_gaussian_paths}).\footnote{In their work, they use a different convention to condition on the noise. But this is only notation and the algorithm is the same.} As outlined in their paper, they extensively tested various flow and diffusion alternatives and found flow matching to perform best. For training, it uses classifier-free guidance training (with dropping class labels) as outlined above. Further, Stable Diffusion 3 follows the approach outlined in \cref{sec:image_architecture} by training within the latent space of a pre-trained autoencoder. Training a good autoencoder was a big contribution of the first stable diffusion papers.\\

\begin{figure}[!t]
    \centering
    \includegraphics[width=0.9\textwidth]{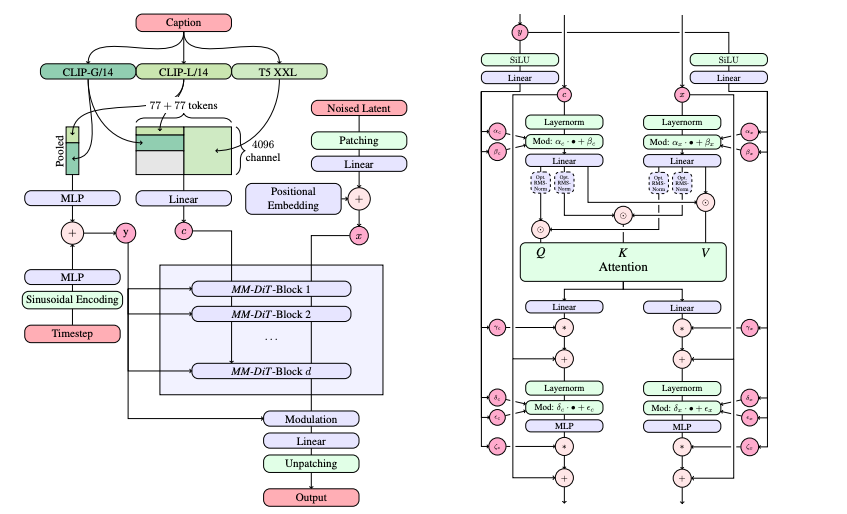}
    \caption{The architecture of the multi-modal diffusion transformer (MM-DiT) proposed in \citep{sd3}. Figure also taken from \citep{sd3}.}
    \label{fig:mmdit}
\end{figure}

To enhance text conditioning, Stable Diffusion 3 makes use of both 3 different types of text embeddings (including CLIP embeddings as well as the sequential outputs produced by a pretrained instance of the encoder of Google's T5-XXL \cite{t5}, and similar to approaches taken in \cite{balaji, saharia}). Whereas CLIP embeddings provide a coarse, overarching embedding of the input text, the T5 embeddings provide a more granular level of context, allowing for the possibility of the model attending to particular elements of the conditioning text. To accommodate these sequential context embeddings, the authors then propose to extend the diffusion transformer to attend not just to patches of the image, but to the text embeddings as well, thereby extending the conditioning capacity from the class-based scheme originally proposed for DiT to sequential context embeddings. This proposed modified DiT is referred to as a \themebf{multi-modal DiT} (MM-DiT), and is depicted in \cref{fig:mmdit}. Their final, largest model has \textbf{8 billion parameters}. For sampling, they use $50$ steps (i.e. they have to evaluate the network $50$ times) using a Euler simulation scheme and a classifier-free guidance weight between $2.0$-$5.0$.
% \begin{equation}
% \label{eq:sd3_cfm}
% \begin{aligned}
%     \Lcond(\theta) &=  \mathbb{E}_{\square}[\|u_t^\theta(x_t) - \uref_t(x_t|\varepsilon)\|^2]\\
%     \square &= \varepsilon \sim \mathcal{N}(0, I_d), x_t\sim p_t(\cdot|\varepsilon)
% \end{aligned}
% \end{equation}
% where $p_0 = \mathcal{N}(0, I_d)$. 
% Note that \cref{eq:sd3_cfm} differs from \cref{eq:cfm} in that we condition on $\varepsilon \in p_0 = \mathcal{N}(0,I_d)$ rather than $z \in p_{\text{data}}$. The analysis is otherwise fundamentally the same. Here, the conditional probability path $p_t(\cdot | \varepsilon)$ is defined implicitly as the law of the random variable
% \begin{equation}
%     X_t \triangleq \alpha_tX_1 + \beta_t \varepsilon, \quad \quad X_1 \sim p_{\text{data}},
% \end{equation}
% where $\alpha_t,\beta_t \in \mathcal{C}^2([0,1])$ satisfy $\alpha_0 = \beta_1 = 0$ and $\alpha_1 = \beta_0 = 1$ \cite{albergo2023stochastic}. Then, \cref{eq:sd3_cfm} may be modified to the general conditional setting as in \cref{eq:cfg_guided_cfm}.

%\paragraph{Architecture.}

\subsubsection{Meta Movie Gen Video}
Next, we discuss Meta's video generator, \themeit{Movie Gen Video} (\url{https://ai.meta.com/research/movie-gen/}). As the data are not images but \themeit{videos},  the data $x$ lie in the space $\mathbb{R}^{T \times C \times H \times W}$ where $T$ represents the new \themebf{temporal} dimension (i.e. the number of frames). As we shall see, many of the design choices made in this video setting can be seen as adapting existing techniques (e.g., autoencoders, diffusion transformers, etc.) from the image setting to handle this extra temporal dimension.\\

%\footnote{The paper actually proposes a slightly different path given instead by the interpolant $X_t = tz + (1-(1-\sigma_{\text{min}})t)X_0$ where $\sigma_{\text{min}} = 10^{-5}$.}.
% \begin{equation}
% \label{eq:moviegen_cfm}
% \begin{aligned}
%     \Lcond(\theta) &=  \mathbb{E}_{\square}[\|u_t^\theta(x_t) - \uref_t(x_t|z)\|^2]\\
%     \square &= z \sim p_{\text{data}}, x_t \sim p_t(\cdot|z),
% \end{aligned}
% \end{equation}
% where the linear (CondOT) probability path $p_t(x_t|z)$ is given implicitly as the law of the interpolant
% \begin{equation}
%     X_t = tz + (1-t)X_0, \quad \quad X_0 \sim p_0 = \mathcal{N}(0, I_d)
% \end{equation}
%so that the conditional vector field $\uref_t(x_t|z)$ is given by $\frac{z - x_t}{1-t}$.
%\paragraph{Architecture.} 
Movie Gen Video utilizes the conditional flow matching objective with the same straight line schedulers $\alpha_t=t,\sigma_{t}=1-t$. Like Stable Diffusion 3, Movie Gen Video also operates in the latent space of frozen, pretrained autoencoder. Note that the autoencoder to reduce memory consumption is even more important for videos than for images - which is why most video generators right now are pretty limited in the length of the video they generate.
% For brevity, we focus on three specific architectural design choices: the autoencoder, the diffusion-transformer backbone, and the conditioning mechanism. Like Stable Diffusion 3, Movie Gen Video also operates in the latent space of frozen, pretrained autoencoder. 
Specifically, the authors propose to handle the added time dimension by introducing a \themebf{temporal autoencoder} (TAE) which maps a raw video $x_t' \in \mathbb{R}^{T' \times 3 \times H \times W}$ to a latent $x_t\in\mathbb{R}^{T \times C \times H \times W}$, with $\tfrac{T'}{T} = \tfrac{H'}{H} = \tfrac{W'}{W} = 8$ \cite{moviegen}. To accomodate long videos, a temporal tiling procedure is proposed by which the video is chopped up into pieces, each piece is encoder separately, and the latents are sticthed together \cite{moviegen}. The model itself - that is, $u_t^\theta(x_t)$ - is given by a DiT-like backbone in which $x_t$ is patchified along the time and space dimensions. The image patches are then passed through a transformer employing both self-attention among the image patches, and cross-attention with language model embeddings, similar to the MM-DiT employed by Stable Diffusion 3. For text conditioning, Movie Gen Video employs three types of text embeddings: UL2 embeddings, for granular, text-based reasoning \cite{ul2}, ByT5 embeddings, for attending to character-level details (for e.g., prompts explicitly requesting specific text to be present) \cite{byte5}, and MetaCLIP embeddings, trained in a shared text-image embedding space \cite{metaclip, moviegen}. Their final, largest model has \textbf{30 billion parameters}. For a significantly more detailed and expansive treatment, we encourage the reader to check out the Movie Gen technical report itself \citep{moviegen}.

% \newpage
% \subfile{subfiles/part_06_distillation_empty}
\newpage
\section{Discrete Diffusion Models: Building Language Models with Diffusion}
\label{sec:discrete_diffusion}
In previous sections, we explored flow and diffusion models as generative models over \emph{Euclidean space} $\mathbb{R}^d$ that allow us to generate data points represented by \emph{vectors} $z\in \mathbb{R}^d$. However, not all data is naturally modeled as a point in Euclidean space $\R^d$. Many data types, such as text or DNA, are more naturally viewed as elements of a \emph{discrete} state space $S$. Most importantly, language consists of a sequence of discrete tokens that we want to model. How could we apply flow and diffusion models to such data types? It turns out that the principles that we have learned in previous sections extend to such data types as well. The resulting models are called \themebf{discrete diffusion models} in the machine learning literature \citep{campbell2022continuous,gat2024discrete}. However, it is important to keep in mind that there is no mathematical diffusion process (SDEs don't exist in discrete state spaces). Instead of having ODEs/SDEs, we use \themebf{continuous-time Markov chains (CTMCs)}. In the following, we will explain CTMC models (see \cref{subsec:ctmc_model}) and how to learn them (see \cref{subsec:ctmcs}) allowing us to build large language models (LLMs) using the principles of flow and diffusion models.

\subsection{Continuous-Time Markov chain (CTMC) models}
\label{subsec:ctmc_model}
In this section, we explain continuous-time Markov chains (CTMCs). You can think of CTMCs as a discrete analogue of SDEs that we can use to build neural network models that generate discrete states. Further, we will introduce CTMC models, i.e. neural network models that allow to generate discrete sequences such as text using CTMCs.

\begin{wrapfigure}{r}{0.5\textwidth}
  %\begin{center}
\includegraphics[width=0.5\textwidth]{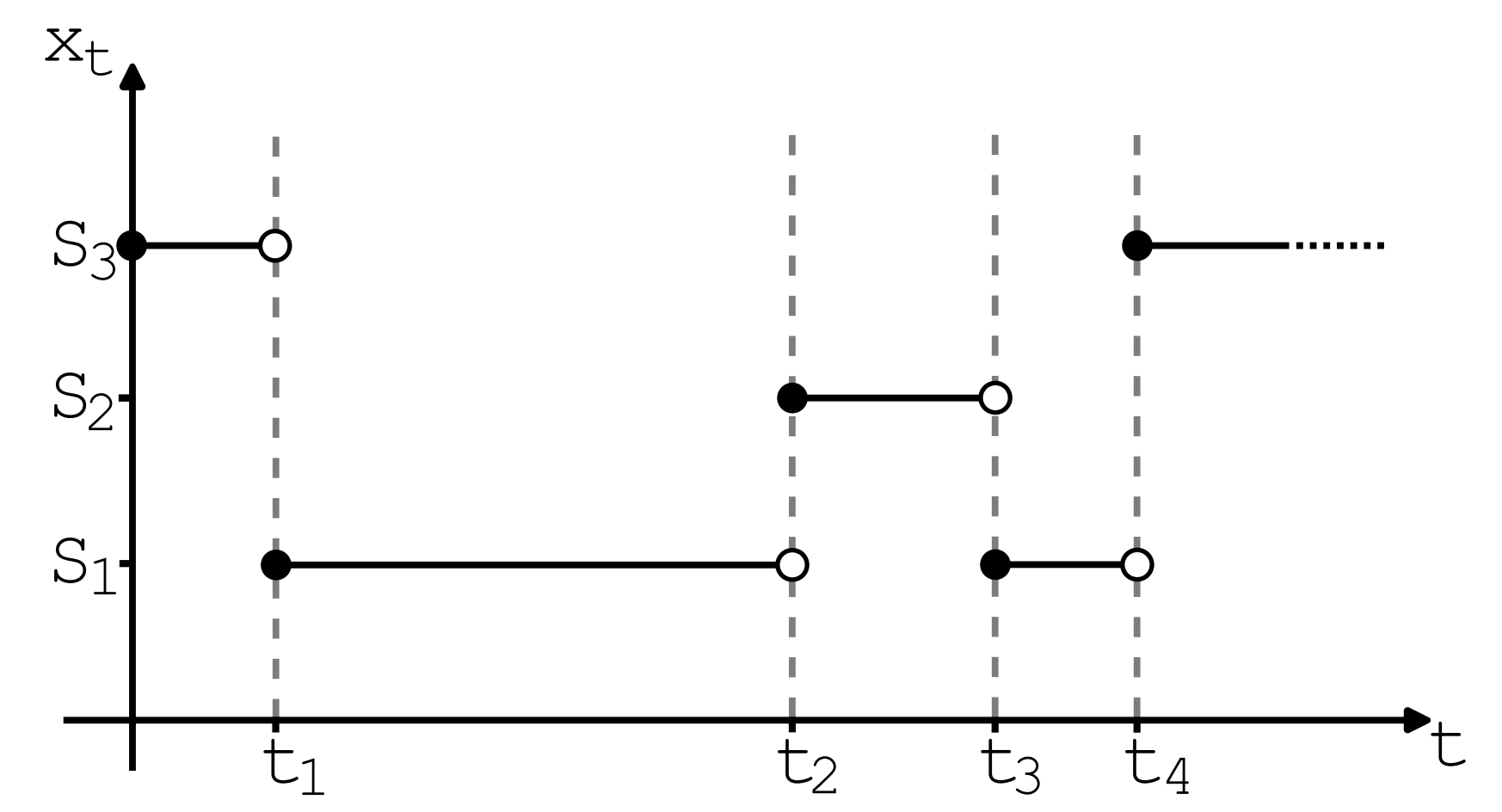}
  %\end{center}
\caption{\label{fig:CTMC_illustration}Illustration of a CTMC trajectory with state space $S=\{S_1,S_2,S_3\}$ (sequence length $d=1$). Figure adapted from \citep{campbell2022continuous}.}
\end{wrapfigure}

Let us begin by characterizing our \themebf{state space} $S$. Let $\mathcal{V}=\{v_{1},\cdots,v_{V}\}$ be our vocabulary. The state space is given by $S=\mathcal{V}^d$ where $d\in \mathbb{N}$ is sequence length and $V\in \mathbb{N}$ is the vocabulary size. For language, $\{v_{1},\cdots,v_{V}\}$ could enumerate our alphabet or a set of discrete tokens and $S$ would represent the set of sequences (or sentences) of length $d$. For DNA, $\{v_{1},\cdots,v_{V}\}$ could be all 4 DNA bases and $S$ all DNA sequences of length $d$.

Next, let $X_t$ be a stochastic process on $S$, i.e. a random trajectory $X:[0,1]\to S, t\mapsto X_{t}$ in $S$. We require $X_t$ to be a \themebf{Markov process}, i.e. a process that has no memory. Specifically, this means that the following condition holds
\begin{align*}
\underbrace{p(X_{t+h}|X_{t}, X_{t_1},\cdots, X_{t_k})}_{\text{prob. of future given present and past}}=\underbrace{p(X_{t+h}|X_{t})}_{\text{prob. of future given present}}\quad (\text{for all }0<h, 0\leq t_1<t_2<\cdots <t_k<t)
\end{align*}
In other words, the probabilities of future events only depend on the present - the past has no relevance for the future anymore. Note that ODE/SDEs - while not on discrete state spaces - are also Markov processes. Here, $X_t$ is on a discrete space and therefore is called a Markov \emph{chain}, specifically a \themebf{Continuous-time Markov chain (CTMC).} The quantity $p_{t+h|t}(X_{t+h}|X_{t})$ are the \themebf{transition probabilities} and they fully determine the CTMC together with the initial distribution $X_0\sim p_0$ of the Markov chain. Therefore, when we say CTMC, you can also just think of transition probabilities $p_{t+h|t}(X_{t+h}|X_{t})$.

Next, let us derive the analogue of a vector field in the discrete setting. As we are in a discrete setting, we can only jump (or switch) between states - we cannot go into a direction anymore like we did when specifying ODEs. Therefore, we define a \themebf{rate matrix} $Q_t(y|x)$ that effectively summarizes the rate of jumping (or switching) from state $x\in S$ to state $y\in S$. Formally, a rate matrix $Q_t$ is given by a bounded function (continuous in time) 
\begin{align}
\label{eq:rate_matrix}
    Q:S\times S\times[0,1] \to \mathbb{R},\quad 
    (x,y,t)\mapsto Q_t(y|x)
\end{align}
where $Q_t(y|x)$ describes the rate of switching from $x$ from $y$ such that 
\begin{align}
\label{eq:condition_1_ctmc}
    \text{(1) Outgoing rates are positives:  }Q_t(y|x)\geq& 0 \quad \text{whenever }x\neq y\\
\label{eq:condition_2_ctmc}
    \text{(2) Rate staying equals negative outgoing rate: }Q_t(x|x)=&-\sum\limits_{y\neq x}Q_t(y|x)\quad \text{ for all }x
\end{align}
The two conditions are intuitive: The first condition says that the rate of switching from $x$ to a different state $y\neq x$ can only be non-negative (not switching just corresponds to $0$ - so it does not make sense to have a rate that is smaller than $0$). The second condition says that the rate $Q_t(x|x)$ of staying at $x$ should cancel out with the rate of leaving $x$ - it is essentially a consistency condition saying that you have to either stay at $x$ or leave (there is no third option). Note that these conditions imply in particular that $Q_t(x|x)\leq 0$. Hence, $Q_t(y|x)$ is a matrix whose diagonal entries are all non-positive while all off-diagonal entries are non-negative.

We can now define the analogue of a differential equation, i.e. a condition on a CTMC to ``follow'' the rate matrix. The idea is basically that the distribution or evolution of $X$ should follow the rate matrix $Q_t$. In other words, we require that the transition probabilities fulfill
\begin{align}
\label{eq:ctmc_ode}
    \frac{\dd}{\dd h}p_{t+h|t}(X_{t+h}=y|X_{t}=x)_{|h=0}=Q_t(y|x)\quad\text{for all }x,y\in S, 0\leq t
\end{align}
The left-hand side is the infinitesimal rate of change of the probability of switching from $x$ to $y$. We impose the condition that these probabilities should change as specified by the rate matrix. Let's briefly check that it reasonable to request these conditions, i.e. we simply set $Q_t(y|x)$ as in \cref{eq:ctmc_ode}, would it be a valid rate matrix? For $h=0$, the probability of switching from $x$ to $y\neq x$ is zero (as no time has passed), i.e. $p_{t|t}(y|x)=0$ for all $y\neq x$. Therefore, we know that the derivative must be non-negative and $Q_t(y|x)\geq 0$ whenever $y\neq x$. This checks that the first condition in \cref{eq:condition_1_ctmc} holds. Further, we know that 
\begin{align*}
\sum\limits_{y\neq x} Q_t(y|x)=\sum\limits_{y\neq x}\frac{\dd}{\dd h}p(X_{t+h}=y|X_{t}=x)_{|h=0}=\frac{\dd}{\dd h}\sum\limits_{y\neq x}p(X_{t+h}=y|X_{t}=x)_{|h=0}=&\frac{\dd}{\dd h}(1-p(X_{t+h}=x|X_{t}=x))\\
=&-Q_t(x|x)
\end{align*}
where we used that probabilities sum to $1$. This shows \cref{eq:condition_2_ctmc}. This checks that every CTMC has at least one rate matrix satisfying \cref{eq:ctmc_ode}. But what if we go backwards - what if we specify $Q_t$, is there a corresponding CTMC and if so, is it unique? This is indeed the case.
\begin{theorem}[CTMC existence and uniqueness]
\label{thm:ctmc_existence_and_uniqueness}
For any rate matrix $Q_t$ (bounded and continuous in time $t$), there is a unique Markov chain $X_{t}$ (i.e. a unique set of transition probabilities $p_{t+h|t}(y|x)$) such that \cref{eq:ctmc_ode} holds.
\end{theorem}
For the interested reader, we provide a self-contained proof in \cref{appendix:existence_uniquenss_ctmcs}
. The key takeaway from this theorem is that for the purposes of machine learning, we can state a construct a rate matrix $Q_t$ (e.g. via  a neural network) and assume that there is a unique Markov chain that corresponds to $Q_t$.
\begin{examplebox}[Two-state CTMC with equal jump rates]
Let $S=\{a,b\}$ and consider a time-homogeneous CTMC $(X_t)_{t\ge 0}$ that
switches between both states at a constant rate $\lambda>0$:
\[
Q=
\begin{array}{c|cc}
 & a & b\\\hline
a & -\lambda & \lambda\\
b & \lambda & -\lambda
\end{array}.
\]
Then the transition probabilities over a time increment $h\ge 0$ are also constant in time $t$ and given by
\begin{align*}
\begin{pmatrix}
p(X_{t+h}=a|X_t=a) & p(X_{t+h}=a|X_t=b)\\
p(X_{t+h}=b|X_t=a) & p(X_{t+h}=b|X_t=b)
\end{pmatrix}
=
\frac12
\begin{pmatrix}
1+e^{-2\lambda h} & 1-e^{-2\lambda h}\\
1-e^{-2\lambda h} & 1+e^{-2\lambda h}
\end{pmatrix}.
\end{align*}
One can check by hand that \cref{eq:ctmc_ode} holds, i.e. these transition probabilities indeed are the correct ones for that rate matrix. In fact, these rates are very intuitive: The chain keeps flipping with an instantaneous rate
$\lambda$. The exponential term $e^{-2\lambda h}$
captures how the memory of the initial state decays. As infinite time passes, i.e. for $h\to\infty$, it holds that
\[
P(h)\to \begin{pmatrix}\frac12&\frac12\\\frac12&\frac12\end{pmatrix},
\]
so the chain forgets where it started and is in $a$ or $b$ with probability $1/2$. This convergence is faster the higher the rate $\lambda>0$ of switching.
\end{examplebox}

\textbf{Simulation of CTMC.} Next, let us think about how one would go about simulating a trajectory of a CTMC. Let $h>0$ be a step size and $\pinit$ be an initial distribution over $S$, e.g. $\pinit=\text{Unif}_{S}$ is the uniform distribution over $S$. Then we can simulate it iteratively by setting $X_0\sim \pinit$ and setting
\begin{align*}
X_{t+h}\sim &p_{t+h|t}(\cdot|X_t)
\end{align*}
Now, this would work if we knew $p_{t+h|t}(\cdot|X_t)$. However, for all but the simplest CTMCs, we typically do not know the transition kernel in closed form and only have access to the rate matrix $Q_t$. Still, by \cref{eq:ctmc_ode}:
\begin{align*}
  p_{t+h|t}(X_{t+h}=y|X_t=x)
=p_{t|t}(X_{t}=y|X_t=x)+hQ_t(y|x)+R_{t}(h)=1_{y=x}+hQ_t(y|x)+R_t(h)
\end{align*}
where $R_t(h)$ is an error term that we can neglect for small $h$. Therefore, for small $h$, we can set
\begin{align*}
p_{t+h|t}(X_{t+h}=y|X_t=x)\approx 1_{y=x}+hQ_t(y|x)=:\tilde{p}_{t+h|t}(y|x)
\end{align*}
One can check that $\tilde{p}_{t+h|t}(y|x)$ is indeed a valid probability distribution for small $h$ by the conditions we imposed on the rate matrix. Therefore, we can approximately sample the next point via
\begin{align}
\label{eq:ctmc_simulation}
X_{t+h}\sim \tilde{p}_{t+h|t}(\cdot|x)=(1_{y=x}+hQ_t(y|x))_{y\in S}
\end{align}
As the above is just a discrete distribution, we can sample from it easily via standard methods. This is a simple way to simulate a CTMC.

\paragraph{CTMC model.} Next, let us define how we can a parameterize a CTMC in a neural network. A \themebf{CTMC model} (or \themebf{discrete diffusion model}) is given by an initial distribution $\pinit$ over $S$ and a neural network $Q_t^\theta$ with parameters $\theta$ such that for every input $x\in S$ the model returns a single column of the rate matrix
\begin{align*}
    x\mapsto \{Q_t^\theta(y|x)\}_{y\in S}
\end{align*}
We want the model to return an entire column because we require it for simulation of the CTMC (\cref{eq:ctmc_simulation}), i.e. sampling the next state.

\begin{figure}[!t]
    \centering
\includegraphics[width=0.9\textwidth]{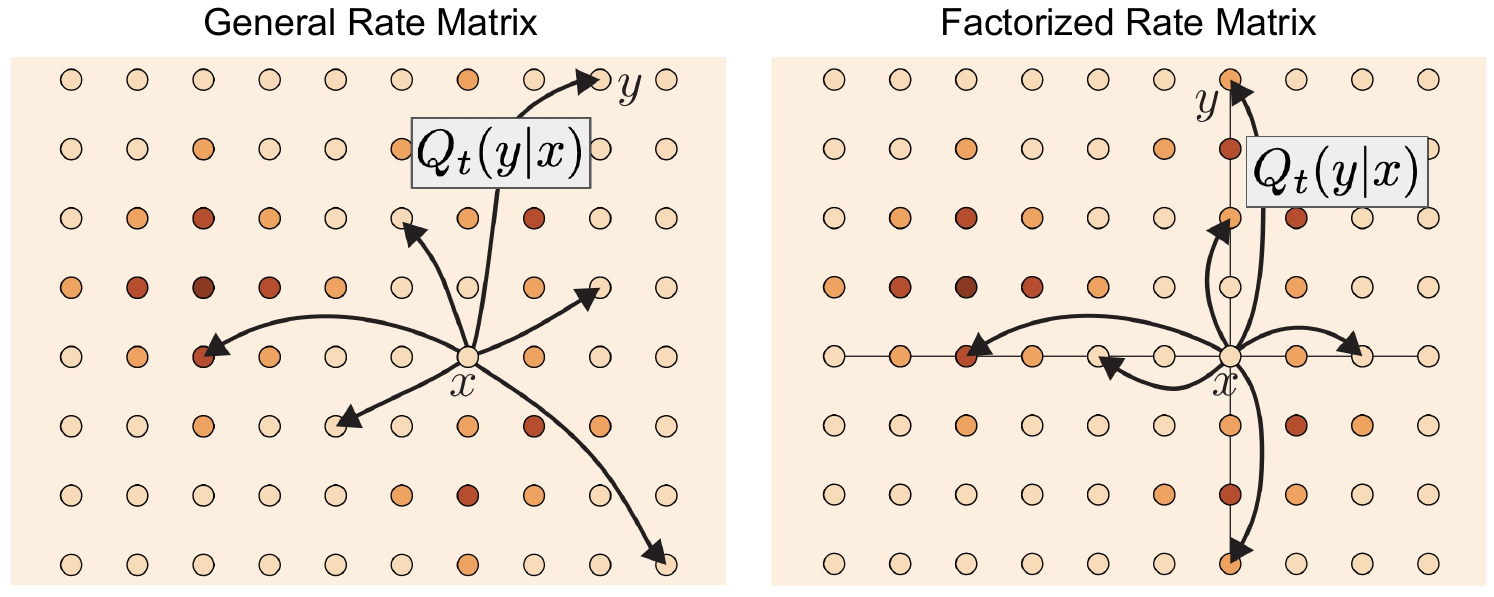}
    \caption{Illustration of a factorized CTMC model. Factorized CTMCs have only non-zero rates ($Q_t(y|x)\neq 0$) if the start and end point differ by only one dimension (here, $d=2$). Figure taken from \citep{lipman2024flow}.}
\label{fig:factorized_ctmc}
\end{figure}

One complication with the above model is that the space $S$ can be \emph{very} large. In particular, $|S|=V^d$ where $V$ is our vocabulary size and $d$ is the sequence length. This exponential growth makes it basically impossible to store an entire column of the rate matrix in memory - $\{Q_t^\theta(y|x)\}_{y\in S}$ could never be represented in a computer. Therefore, we have to constrain the model.  Specifically, almost all CTMC models are \emph{factorized} (see \cref{fig:factorized_ctmc}), which is effectively a sparsity constraint. Specifically, a \themebf{factorized CTMC model} is given by a CTMC model $Q_t^\theta$ such that for all $y=(y_1,\cdots, y_d),x=(x_1,\cdots,x_d)\in S=\mathcal{V}^d$ it holds
\begin{align*}
Q_t^\theta(y|x)=0\quad \text{whenever }y_i\neq x_i\text{ for more than one position }i
\end{align*}
We call all $y$ that differ from $x$ in at most one token the \themebf{neighbors} $N(x)$ of $x$. We can write such a factorized CTMC model as
\begin{align*}  x\mapsto \{Q_t^\theta(y|x)\}_{y\in N(x)} 
=&
\begin{pmatrix}
Q_t^\theta(v_1,1|x) & \cdots Q_t^\theta(v_{V},1|x)\\
\cdots\\
Q_t^\theta(v_1,d|x) & \cdots Q_t^\theta(v_{V},d|x)\\
\end{pmatrix}
\end{align*}
where $Q_t(y|x)=Q_t^\theta(v_i,j|x)$ now gives the rate of going from $x=(x_1,\cdots,x_d)$ to the neighbor of $x$ that we obtain swapping out the $j$-th element with $v_i$, i.e. $y=(x_1,\cdots,x_{j-1},v_{i},x_{j+1},\cdots,x_{d})$. Each row corresponds to a rate matrix per position $i=1,\cdots,d$, i.e. we require 
\begin{align*}
    Q_t^\theta(v,i|x)\geq 0\text{ if }v\neq x_i,\quad Q_t(x_i,i|x)=-\sum\limits_{v\neq x_i}Q_t^\theta(v,i|x)
\end{align*}
We can enforce these conditions on the output of a neural network easily, e.g. one can use a  transformer model on sequence length $d$ with output dimension $V$. Note also that the factorized rate matrix makes the output shape $d\times V$ - this size increases linearly in the dimension (as opposed to exponentially). 

% Finally, we note that then the incoming rate at $x$ is given by the sum of all incoming rates for each position:
% \begin{align*}
%     Q_t^\theta(x|x)=-\sum\limits_{i=1}^{d}\sum\limits_{v\neq x_i}Q_t^\theta(v,i|x)
% \end{align*}

\paragraph{Simulating a CTMC model.} To sample from a CTMC model, we sample $X_{0}\sim \pinit$ and perform an iteration where we sample the next state according to \cref{eq:ctmc_simulation}. We present an algorithm in \cref{alg:sampling_ctmc_model}. As shown there, for factorized CTMC models, one can use a parallel per-token Euler approximation, where each token is updated independently during a small step $h>0$. This approximation agrees with the full CTMC Euler step up to first order in $h$, but allows for a $O(h^2)$ probability of simultaneous updates to multiple tokens.
\begin{algorithm}[H]
\caption{Sampling from a Factorized CTMC Model}
\label{alg:sampling_ctmc_model}
\begin{algorithmic}[1]
\REQUIRE Rate network $Q_t^\theta$ (factorized), initial distribution $\pinit$, number of steps $n$
\STATE Set $t \gets 0$, step size $h \gets \frac{1}{n}$
\STATE Draw a sample $X_0 \sim \pinit$, where $X_0=(X_0^{(1)},\dots,X_0^{(d)})\in\mathcal{V}^d$
\FOR{$i=1,\dots,n$}
    \STATE Compute factorized jump rates $\{q_{j}(v)\}_{j=1..d,\ v\in\mathcal{V}} \gets Q_t^\theta(\cdot \mid X_t)$
    \FOR{$j=1,\dots,d$ \textbf{(in parallel)}}
        \STATE $x \gets X_t^{(j)}$ \COMMENT{current token at position $j$}
        \STATE Define the per-position Euler transition probabilities $\tilde p_{j,t}(\cdot \mid X_t^{(j)}=x)$ by
        \[
        \tilde p_{j,t}(v\mid x) \;=\;
        \begin{cases}
        h\, q_{j}(v), & v\neq x,\\[4pt]
        1 - h\sum\limits_{v'\in \mathcal{V}\setminus\{x\}} q_{j}(v'), & v=x.
        \end{cases}
        \]
        \STATE Sample $X_{t+h}^{(j)} \sim \textsc{Categorical}\!\left(\{\tilde p_{j,t}(v\mid x)\}_{v\in\mathcal{V}}\right)$
    \ENDFOR
    \STATE Set $t \gets t + h$
\ENDFOR
\RETURN $X_1$
\end{algorithmic}
\end{algorithm}

\subsection{Training CTMC models}
\label{subsec:ctmcs}

We next discuss how to learn CTMC models. The principles are the same as for flow matching: (1) We construct a probability path interpolating between noise and data. (2) We derive a conditional rate matrix and marginal rate matrix. (3) We learn the marginal rate matrix in a simulation-free manner. We will explain this recipe now step-by-step.

In this section, the \themebf{data distribution} $\pdata$ is a distribution over $S$ characterized by a probability mass function. Namely, $\pdata:S\to\mathbb{R}_{\geq 0}, z\mapsto \pdata(z)$ with $\sum_{z\in S}\pdata(z)=1$. We do not know $\pdata$ but we access to samples $z\sim \pdata$ during training in form of a data set. For example, all texts on the world wide web. Our goal is to learn to generate samples $z\sim \pdata$. Our goal is to train the CTMC model $Q_t^\theta$ such that 
\begin{align*}
X_0\sim \pinit, \quad X_t\text{ CTMC of }Q_t^\theta\quad \Rightarrow \quad X_{1}\sim \pdata
\end{align*}
So as you might realize, this is no different from the Euclidean case $\mathbb{R}^d$ (see \cref{sec:odes_sdes,sec:flow_matching}), just that we use a CTMC model instead of a flow/diffusion model.

\subsubsection{Conditional and Marginal Probability Path}
We define $\delta_{z}(x)$ to be function such that $\delta_{z}(x)=0$ if $x\neq z$ and $\delta_{z}(x)=1$ if $x=z$. A (discrete) \themebf{conditional probability path} is given by set of  distributions $p_t(x|z)$ for $x,z\in S$ and $0\leq t\leq 1$ such that 
\begin{align*}
p_0(\cdot|z)=\pinit, \quad p_1(\cdot|z)=\delta_{z}
\end{align*}
So similar to the Euclidean case, a discrete conditional probability path interpolates between a distribution that is independent of $z$ to a distribution that has all mass on $z$. A (discrete) \themebf{marginal probability path} is then given by
\begin{align*}
p_t(x)=\sum\limits_{z\in S}p_t(x|z)\pdata(z)
\end{align*}
One can easily check that the marginal probability path interpolates ``noise'' and data:
\begin{align}
\label{eq:discrete_prob_path_interpolation}
    p_0=\pinit, \quad p_1=\pdata
\end{align}
\begin{examplebox}[Factorized mixture path (independent noising per token)]
\label{example:factorized_mixture_path}
Let $S=\mathcal{V}^d$ and let $\pinit(x)=\prod_{j=1}^d \pinit^{(j)}(x_j)$ be a factorized initial distribution.
Fix a \textbf{scheduler} $0\le \kappa_t\le 1$ such that $\kappa_{0}=0,\kappa_{1}=1$ with $\frac{\dd}{\dd t}\dot{\kappa}_t\geq 0$. Define the conditional path by
\begin{align*}
p_t(x|z)
&=\prod_{j=1}^d\Big[(1-\kappa_t)\,\pinit^{(j)}(x_j)+\kappa_t\,\delta_{z_j}(x_j)\Big].
\end{align*}
Equivalently, one can sample $x\sim p_t(\cdot\mid z)$ by drawing i.i.d.\ masks $m_j=0,1$ and noise $\xi_j\sim \pinit^{(j)}$, then setting
\begin{align*}
m_j&\sim\mathrm{Bernoulli}(\kappa_t), \quad \xi_j\sim \pinit^{(j)}\\
x_j &= m_j\, z_j + (1-m_j)\,\xi_j,\qquad j=1,\dots,d\\
x&=(x_1,\cdots,x_d)
\end{align*}
We call the above the \textbf{factorized mixture path}. The above procedure effectively ``destroys'' the $j$-th token independently for each position in the sequence with a probability $1-\kappa_{t}$, i.e. for $t=0$ $1-\kappa_{t}=1$ and all information is destroyed and for $t=1$ it holds that $1-\kappa_{t}=0$ and no information is destroyed. Note that this is similar to the Gaussian probability path \cref{example:gaussian_path} in the sense that information is destroyed progressively with a speed determined by a scheduler $\kappa_{t}$. However, it is also different from the Gaussian probability path as the factorized mixture path does \emph{not} move/transports probability mass (there is no direction as we are in discrete space) - it simply fades in one distribution and fades out another.
\end{examplebox}

\begin{figure}[H]
  \begin{center}
\includegraphics[width=\textwidth]{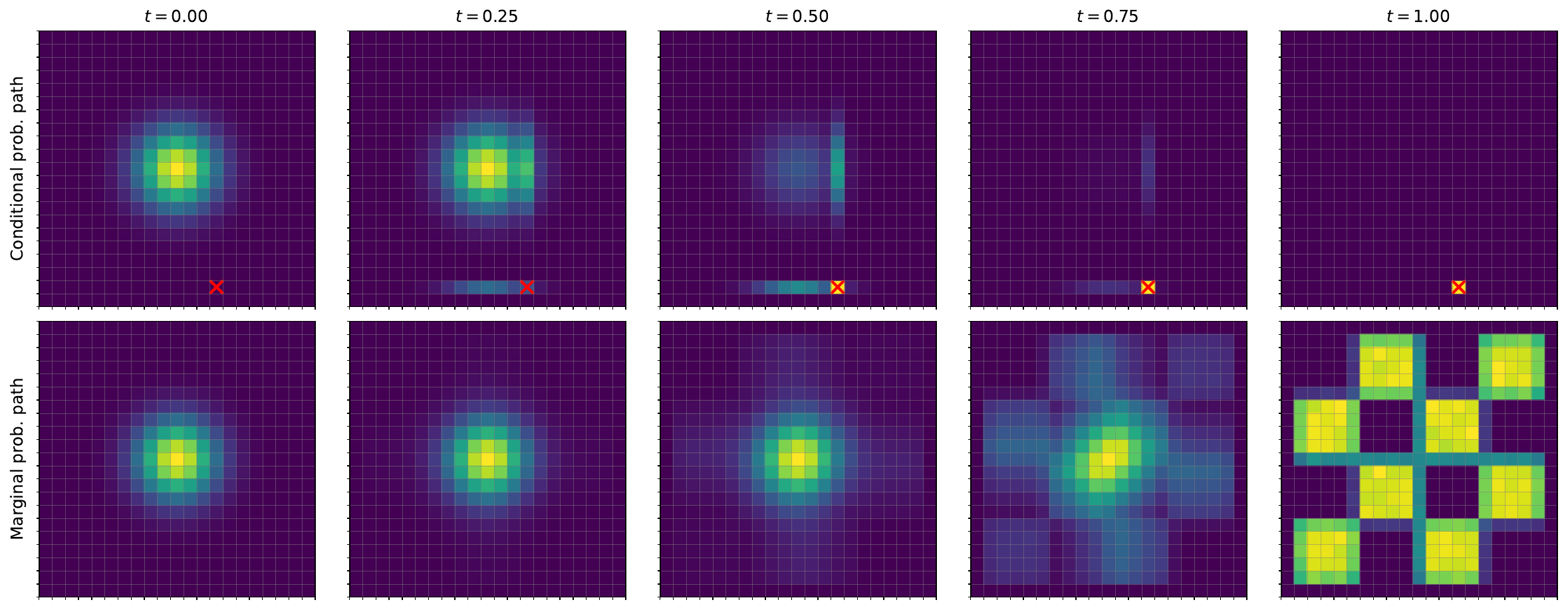}
\end{center}
\caption{\label{fig:discrete_prob_path_illustration} Illustration of a discrete probability path for $d=2$. Top row: Conditional probability path interpolating between initial distribution and Dirac distribution. Bottom row: Interpolation between initial distribution and data distribution (here, chess board pattern). Note the similarity and differences to \cref{fig:cond_marginal_path_histograms}: Here, the probability path is ``teleported'' (we downweigh the initial distribution and upweight the terminal distribution).}
\end{figure}

\subsubsection{Conditional and Marginal Rate Matrix}

As a next step, we will now construct the training target of discrete flow matching. First, we construct a conditional rate matrix - the analogue to the conditional vector field for flow matching. Let $Q_t^z(y|x)$ be a rate matrix for every data point $z\in S$. Then we call it a \themebf{conditional rate matrix} if 
\begin{align*}
    X_0\sim \pinit,\quad  X_t\text{ CTMC of }Q_t^z\quad \Rightarrow \quad X_{t}\sim p_t(\cdot|z)
\end{align*}
In other words, the conditional rate matrix is such that its CTMC ``follows'' the conditional probability path. The conditional rate matrix serves as a building block to construct the marginal rate matrix that follows the marginal probability path:
\begin{theorem}[Discrete marginalization trick]
\label{thm:discrete_marginalization_trick}
The \themebf{marginal rate matrix} defined by
\begin{align}
\label{eq:discrete_marginal_rate_matrix}
Q_t(y|x)=\sum\limits_{z\in S}Q_t^z(y|x)\frac{p_t(x|z)\pdata(z)}{p_t(x)}=\sum\limits_{z\in S}Q_t^z(y|x)p_{1|t}(z|x)\quad \text{where }p_{1|t}(z|x):=\frac{p_t(x|z)\pdata(z)}{p_t(x)}
\end{align}
is a valid rate matrix and fulfills the following condition:
\begin{align*}
     X_0\sim \pinit,\quad  X_t\text{ CTMC of }Q_t\quad \Rightarrow \quad X_{t}\sim p_t
\end{align*}
In particular, $X_1\sim \pdata$ by \cref{eq:discrete_prob_path_interpolation}, i.e. the CTMC of the marginal rate matrix converts noise to data.
\end{theorem}
To prove this statement, we need a fundamental equation for CTMCs, the so-called \emph{Kolmogorov Forward equation}:
\begin{proposition}[Kolmogorov Forward Equation]
\label{prop:kfe_ctmc}
Let $p_t$ be a set of distributions on $S$ for every $0\leq t\leq 1$. Further, let $X_t$ be a CTMC with matrix $Q_t$ and initial distribution $p_0$. Then $X_t\sim p_t$ for all $0\leq t\leq 1$ if and only if the \themebf{Kolmogorov Forward Equation (KFE)} holds:
\begin{align*}
    \frac{\dd}{\dd t}p_t(x)=\sum\limits_{y\in S}Q_t(x|y)p_t(y)
\end{align*}
\end{proposition}
\begin{proof}[Proof of KFE]
To show that the KFE is necessary, assume that $p_t(x)$ are the true marginals of the CTMC, i.e. $X_t\sim p_t$ for every $0\leq t\leq 1$. Then we can compute:
\begin{align*}
\frac{\dd}{\dd t}p_t(x)&\overset{(i)}{=}\frac{\dd}{\dd h}_{|h=0}p_{t+h}(x)\\
&\overset{(ii)}{=}\frac{\dd}{\dd h}_{|h=0}\sum\limits_{y}p_{t+h|t}(x|y)p_t(y)\\
&\overset{(iii)}{=}\sum\limits_{y}\frac{\dd}{\dd h}_{|h=0}p_{t+h|t}(x|y)p_t(y)\\
&\overset{(iv)}{=}\sum\limits_{y}Q_t(x|y)p_t(y)
\end{align*}
where in $(i)$ we simple use a time offset, in $(ii)$ we use the definition of the transition probabilities, in $(iii)$ we swap sum and derivative, and in $(iv)$ we use the definition of the rate matrix (see \cref{eq:ctmc_ode}).

Next, to show that the KFE is sufficient, we can rewrite the KFE in matrix form:
\begin{align*}
\frac{\dd}{\dd t}p_t = Q_tp_t
\end{align*}
where in this equation we consider $p_t=(p_t(x))_{x\in S}$ as a vector and $Q_t=(Q_t(y|x))_{x,y\in S}$ as a matrix. Note that the above is a linear ODE over vector space $\mathbb{R}^S$. Its initial condition is fixed by $p_0$ as stated in the theorem. Therefore, if any other set of marginals $q_t$ fulfills this equation, we know that by the uniqueness of ODEs (see \cref{thm:ode_existence_and_uniqueness}) that we can conclude that $q_t=p_t$. This shows that the KFE is also sufficient.
\end{proof}

\begin{proof}[Proof of \cref{thm:discrete_marginalization_trick}]
Using the KFE, it remains to show that marginal rate matrix defined as in the theorem (see \cref{eq:discrete_marginal_rate_matrix}) fulfills the KFE:
\begin{align*}
\frac{\dd}{\dd t}p_t(x)\overset{(i)}{=}&\frac{\dd}{\dd t}\sum\limits_{z\in S}p_t(x|z)\pdata(z)\\
\overset{(ii)}{=}&\sum\limits_{z\in S}\frac{\dd}{\dd t}p_t(x|z)\pdata(z)\\
\overset{(iii)}{=}&\sum\limits_{z\in S}\left[\sum\limits_{y\in S}Q_t^z(x|y)p_t(y|z)\right]\pdata(z)\\
\overset{(iv)}{=}&\sum\limits_{y\in S}p_t(y)\left[\sum\limits_{z\in S}Q_t^z(x|y)\frac{p_t(y|z)\pdata(z)}{p_t(y)}\right]\\
\overset{(v)}{=}&\sum\limits_{y\in S}p_t(y)Q_t(x|y)
\end{align*}
where $(i)$ follows by the definition of the marginal probability path, in $(ii)$ we swap the sum and the derivative, in $(iii)$ we use the KFE on the conditional rate matrix, in $(iv)$ we multiply and divide by $p_t(y)$, and in $(v)$ we use the definition of the marginal rate matrix $Q_t(y|x)$. This shows that the KFE is fulfilled. The statement follows by \cref{prop:kfe_ctmc}.
\end{proof}
Let us now derive a concrete example of a conditional rate matrix for the factorized mixture path.
\begin{examplebox}[Conditional rate matrix for factorized mixture path]
\label{example:cond_rate_matrix_mixture_path}
Set $\frac{\dd}{\dd t}\kappa_t=\dot{\kappa}_t$. The factorized mixture path has a factorized conditional rate matrix given by
\begin{align*}
Q_t^z(y|x)&=(Q_t^z(v_i,j|x_j))_{v_i,j}\\
Q_t^z(v_i,j|x_j)&=\frac{\dot{\kappa}_t}{1-\kappa_t}(\delta_{z_j}(v_i)-\delta_{x_j}(v_i))\\
=&\frac{\dot{\kappa}_t}{1-\kappa_t}\begin{cases}
    0 & \text{if }x_j=z_j\\
    1 & \text{ if }v_i=z_j, x_j\neq z_j\\
    0 & \text{ if }v_i\neq z_j, x_j\neq z_j\\
    -1 & \text{ if }v_i=x_j, x_j\neq z_j
\end{cases}
\end{align*}
Note that this is a very simple rate matrix: It only allows for jumps to $z^j$ - i.e. if any token $j$ is updated, it must jump to the token value of the terminal data point $z=(z_1,\cdots,z_d)$ - and it only jumps to $z^j$ if we are not yet there.
\begin{proof}
We note that the factorized mixture path completely factorizes into independent components and so does the suggested conditional rate matrix. Therefore, we can without loss of generality assume that $d=1$. So we just do the calculation per dimension. Then, we can derive:
\begin{align*}
\frac{\dd}{\dd t}p_t(x|z)
\overset{(i)}{=}&\frac{\dd}{\dd t}\left[(1-\kappa_t)\pinit(x)+\kappa_t\delta_{z}(x)\right]\\
\overset{(ii)}{=}&\dot{\kappa}_t\delta_{z}(x)-\dot{\kappa}_t\pinit(x)\\
\overset{(iii)}{=}&\frac{\dot{\kappa}_t}{1-\kappa_t}(\delta_{z}(x)-[(1-\kappa_t)\pinit(x)+\kappa_t\delta_{z}(x)])\\
\overset{(iv)}{=}&\frac{\dot{\kappa}_t}{1-\kappa_t}(\delta_{z}(x)-p_t(x|z))\\
\overset{(v)}{=}&\frac{\dot{\kappa}_t}{1-\kappa_t}\delta_{z}(x)\left(
1-p_t(x|z)
\right)+\frac{\dot{\kappa}_t}{1-\kappa_t}(\delta_{z}(x)-1)p_t(x|z)\\
\overset{(vi)}{=}&\sum\limits_{y\neq x}\frac{\dot{\kappa}_t}{1-\kappa_t}\delta_{z}(x)p_t(y|z)
+\frac{\dot{\kappa}_t}{1-\kappa_t}(\delta_{z}(x)-1)p_t(x|z)\\
\overset{(vii)}{=}&\sum\limits_{y\neq x}Q_t^z(x|y)p_t(y|z)+Q_t^z(x|x)p_t(x|z)\\
\overset{(viii)}{=}&\sum\limits_{y\in S}Q_t^z(x|y)p_t(y|z)
\end{align*}
where $(i)$ uses the definition of the factorized mixture path for $d=1$, $(ii)$ is obtained by taking derivatives and setting $\frac{\dd}{\dd t}\kappa_t=\dot{\kappa}_t$, $(iii)$ follows by simple algebra, $(iv)$ by the definition of the factorized mixture path, $(v)$ by simple algebra, $(vi)$ follows by the definition the fact that $\sum_{y\in S}p_t(y|z)=1$, $(vii)$ by the definition of the rate matrix, and $(viii)$ by simple algebra. The above shows that the KFE is fulfilled and therefore the statement follows.
\end{proof}
\end{examplebox}

\subsubsection{Learning the Marginal Rate Matrix}
In this section, we derive the fundamental algorithm for training CTMC models. By \cref{thm:discrete_marginalization_trick}, training a  CTMC model $Q_t^\theta(y|x)$ can be achieved by learning the marginal rate matrix. 

In this section, we now restrict ourselves to the factorized mixture path (see \cref{example:factorized_mixture_path}) as this is the path most discrete diffusion/flow matching models use so far. In this case, the marginal rate matrix has a very intuitive shape:
\begin{theorem}[Marginalization trick for factorized mixture path]
The marginal rate matrix of the factorized mixture path is factorized and has the form
\begin{align*}
Q_t(v_i,j|x)&=\frac{\dot{\kappa}_t}{1-\kappa_t}(p_{1|t}(z_j=v_i|x)-\delta_{x_j}(v_i))
\end{align*}
where $p_{1|t}(z_j=v_i|x)$ is the conditional probability of the $j$-th position ($j$-th token in the sequence) being equal to $v_i$ given the full noisy sequence $x$.
\end{theorem}
\begin{proof}
The marginal rate matrix is given by
\begin{align}
Q_t(y|x)&=\sum\limits_{z\in S}Q_t^z(y|x)p_{1|t}(z|x)
\end{align}
Now, whenever $y$ and $x$ are not neighbors (differ by more than one token), $Q_t^z(y|x)=0$ for every $z$. Therefore, also $Q_t(y|x)=0$ in this case. This shows that marginal rate matrix factorizes as well. It then holds that 
\begin{align}
Q_t(v_i,j|x)&=\sum\limits_{z\in S}Q_t^z(v_i,j|x)p_{1|t}(z|x)\\
&\overset{(i)}{=}\sum\limits_{z\in S}\frac{\dot{\kappa}_t}{1-\kappa_t}(\delta_{z_j}(v_i)-\delta_{x_j}(v_i))p_{1|t}(z|x)\\
&\overset{(ii)}{=}\frac{\dot{\kappa}_t}{1-\kappa_t}\left(\sum\limits_{z\in S}\delta_{z_j}(v_i)p_{1|t}(z|x)-\delta_{x_j}(v_i)\right)\\
&\overset{(iii)}{=}\frac{\dot{\kappa}_t}{1-\kappa_t}\left(p_{1|t}(z_j=v_i|x)-\delta_{x_j}(v_i)\right)
\end{align}
where $(i)$ follows by the formula for the conditional rate matrix (see \cref{example:cond_rate_matrix_mixture_path}), $(ii)$ follows by the fact that $\sum_{z\in S}p_{1|t}(z|x)=1$, and $(iii)$ follows by marginalization. This finishes the proof.
\end{proof}
The previous theorem is remarkable: The marginal rate matrix is effectively a reparameterization of the probabilities $p_{1|t}(z_j=v_i|x)$. This is effectively nothing else than learning a classifier for each token position $j=1,\dots,d$. In other words, we can simply define a \themebf{denoising probabilities network} as
\begin{align*}
p_{1|t}^\theta:\underbrace{x}_{\text{network input}}\mapsto\underbrace{(p_{1|t}^\theta(z_j=v_i|x))_{j=1,\cdots,d, v_i\in \mathcal{V}}}_{\text{network output}}
\end{align*}
Note that the network output has shape $d\times V$. One can obtain probabilities per token position via  simple softmax layer. The network itself can be a standard sequence-to-sequence network, e.g. a transformer works (see \cref{subsec:transformers}).

\begin{figure}[H]
  \begin{center}
\includegraphics[width=\textwidth]{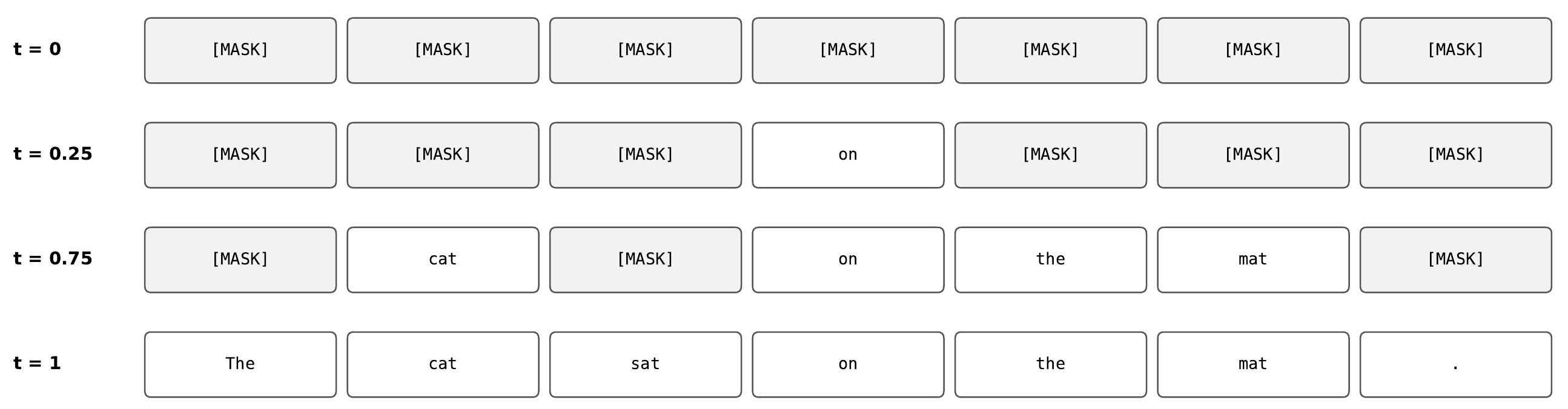}
\end{center}
\caption{\label{fig:MDLM_illustration} Illustration of the trajectory of a Masked Diffusion Language Model.}
\end{figure}

As this is simply a classifier per position $j$, we can train such a network via the cross-entropy loss per $j=1,\cdots,d$. This leads to the \themebf{Discrete Flow Matching loss} given by
\begin{align*}
\mathcal{L}_{\text{DFM}}(\theta)=\mathbb{E}_{z\sim \pdata, t\sim\text{Unif}_{[0,1]}, x\sim p_t(\cdot|z)}\left[\sum\limits_{j=1}^{d}-\log p_{1|t}^\theta(z_j|x)\right]
\end{align*}
This is remarkable: To train a generative model, all we need to do is to train a classifier model per position $j$. In the same way as continuous flow matching reduced to simple regression (see \cref{sec:flow_matching}), discrete flow matching and discrete diffusion models reduce  to simple classification training. In \cref{alg:dfm_training_single}, we summarize the training algorithm. Post-training, we can sample via \cref{alg:sampling_ctmc_model}.
\begin{examplebox}[Masked Diffusion Language Model]
A specific case of the above method is \themebf{masked diffusion language models (MDLMs)}. The idea of MDLMs is that we can extend the vocabulary of tokens $\mathcal{V}=\{v_1,\cdots,v_{V}\}$ with a new token $\text{[mask]}$ that indicates that this token is missing (or was masked). Specifically, we set $\mathcal{V}=\{v_1,\cdots,v_{V},\text{[mask]}\}$ and the initial point is simply $\text{[mask]}^d$, i.e. the sequence that is all-masked. Formally, this means setting $\pinit=\delta_{\text{[mask]}^d}$ in the above framework. The sampling procedure is illustrated in \cref{fig:MDLM_illustration}.
\end{examplebox}
\begin{algorithm}[h]
\caption{Training factorized CTMC  Model (Discrete Diffusion)}
\label{alg:dfm_training_single}
\begin{algorithmic}[1]
\REQUIRE Dataset of sequences $z\sim \pdata$ with $z=(z_1,\dots,z_d)\in\mathcal{V}^d$;\\
initial (noise) token marginals $\pinit^{(j)}$ on $\mathcal{V}$; schedule $\kappa_t\in[0,1]$; \\posterior network $f_\theta$ returning per-position logits over $\mathcal{V}$; optimizer \textsc{Opt}
\FOR{each training iteration}
    \STATE Sample a data point $z \sim \pdata$
    \STATE Sample time $t \sim \mathrm{Unif}[0,1]$ and compute $\kappa \gets \kappa_t$
    \STATE Sample a noisy state $x \sim p_t(\cdot\mid z)$ (factorized mixture path):
    \FOR{$j=1,\dots,d$ \textbf{(in parallel)}}
        \STATE Sample mask $m_j\sim\mathrm{Bernoulli}(\kappa)$
        \STATE Sample noise token $\xi_j\sim \pinit^{(j)}$
        \STATE Set $x_j \gets m_j\, z_j + (1-m_j)\,\xi_j$
    \ENDFOR
    \STATE $x \gets (x_1,\dots,x_d)$
    \STATE Predict terminal-token posteriors via logits from the network:
    \[
      \ell_j(\cdot)\ \gets\ f_\theta(x,t)_j
      \qquad\Rightarrow\qquad
      p^\theta_{1|t}(v\mid x)_j \;=\; \mathrm{Softmax}\!\big(\ell_j\big)(v)
    \]
    \STATE Discrete Flow Matching loss (token-wise NLL of $z$):
    \[
      \mathcal{L}_{\text{DFM}}(\theta)
      \;\gets\;
      \sum_{j=1}^d \Big[-\log p^\theta_{1|t}(z_j\mid x)_j\Big]
    \]
    \STATE Update parameters: $\theta \gets \textsc{Opt.step}\big(\nabla_\theta \mathcal{L}_{\text{DFM}}(\theta)\big)$
\ENDFOR
\end{algorithmic}
\end{algorithm}

This completes now a full pipeline of training and sampling CTMC models that allows us to generate discrete sequences such as text. Current state-of-the-art discrete diffusion models \citep{bie2025llada2} use the recipe described in this work, with neural networks (usually transformers) trained on web-scale data.

\begin{remarkbox}[Generator Matching]
You may wonder why the principles of flow/diffusion models could be translated so seamlessly to discrete state spaces. As it turns out, the principles of flow matching are not unique to flows or even CTMCs. Rather, these are general learning principles for constructing generative models with \emph{Markov processes}. This idea leads to the \themebf{Generator Matching} framework \citep{holderrieth2024generator}, a framework that extends and unifies both discrete and continuous flow and diffusion models into one. A generator is a generalization of a vector field $u_t$ and a rate matrix $Q_t$. Markov processes and generators can be built for any data modality and state spaces. For example, you can build models for smooth manifolds \citep{chen2023flow, de2022riemannian} (e.g. geometric data), mixed state spaces (e.g. joint text and image generation) \citep{campbell2024generative}, and other Markov processes such as jump processes \citep{holderrieth2024generator, campbell2023trans}.
\end{remarkbox}

% Optional (for sampling later): convert predicted posterior to a factorized marginal generator:
% \[
% Q_t^\theta(v,j \mid x)
% \;=\;
% \frac{\dot\kappa_t}{1_

% \newpage 
% \subfile{subfiles/acknowledgements.tex}
\printbibliography[
  heading=bibnumbered,
  title={References}
]

\appendix
\newpage
\section{A Reminder on Probability Theory}
\label{appendix:prob_theory_reminder}
We present a brief overview of basic concepts from probability theory. This section was partially taken from \citep{lipman2024flow}.
\subsection{Random vectors}

Consider data in the $d$-dimensional Euclidean space $x=(x^1,\ldots,x^d)\in \Real^d$ with the standard Euclidean inner product $\ip{x,y}=\sum_{i=1}^d x^i y^i$ and norm $\norm{x}=\sqrt{\ip{x,x}}$.
We will consider random variables (RVs) $X\in\R^d$ with continuous probability density function (PDF), defined as a \emph{continuous} function $p_X:\Real^d\too \Real_{\geq 0}$ providing event $A$ with probability
\begin{equation}
    \sP(X\in A) = \int_A p_X(x) \dd x,
\end{equation}
where $\int p_X(x)\dd x = 1$.
By convention, we omit the integration interval when integrating over the whole space ($\int \equiv \int_{\Real^d}$).
To keep notation concise, we will refer to the PDF $p_{X_t}$ of RV $X_t$ as simply $p_t$.
We will use the notation $X \sim p$ or $X \sim p(X)$ to indicate that $X$ is distributed according to $p$.
One common PDF in generative modeling is the $d$-dimensional isotropic Gaussian:
\begin{equation}\label{e:gaussian}
  \gN(x;\mu,\sigma^2 I) = (2\pi\sigma^2)^{-\frac{d}{2}}\exp\left(-\frac{\norm{x-\mu}_2^2}{2\sigma^2}\right),
\end{equation}
where $\mu\in \Real^d$ and $\sigma \in \Real_{>0}$ stand for the mean and the standard deviation of the distribution, respectively.

The expectation of a RV is the constant vector closest to $X$ in the least-squares sense:
\begin{equation}\label{e:E}
    \E\brac{X}=\argmin_{z\in\Real^d} \int \norm{x-z}^2 p_X(x)\dd x = \int x p_X(x)\dd x. % \qquad \metatriangleright\, \highlight{\text{expectation of a random variable}} 
\end{equation}
One useful tool to compute the expectation of \emph{functions of RVs} is the \emph{law of the unconscious statistician}:
%
% which is essentially a change of variables in integration theorem, reads:
%
\begin{equation}\label{e:law_of_uncon_stat}
    \E \brac{f(X)} = \int f(x) p_X(x) \dd x. % \qquad \metatriangleright\,\highlight{\text{Law of the Unconscious Statistician}}
\end{equation}
When necessary, we will indicate the random variables under expectation as $\E_{X} f(X)$.

%

% \yl{weak convergence of random variables $p_t\too p$ as $t\too 1$ is defined by $\E f(X_t) \too \E f(X)$ (standard convergence of scalar series) as $t\too 1$ for all continuous bounded $f$. }

\subsection{Conditional densities and expectations} \label{sec:conditional_densities_and_expectations}
\begin{wrapfigure}{r}{0.25\textwidth}
  \vspace{-40pt}
  \begin{center}
\includegraphics[width=0.25\textwidth]{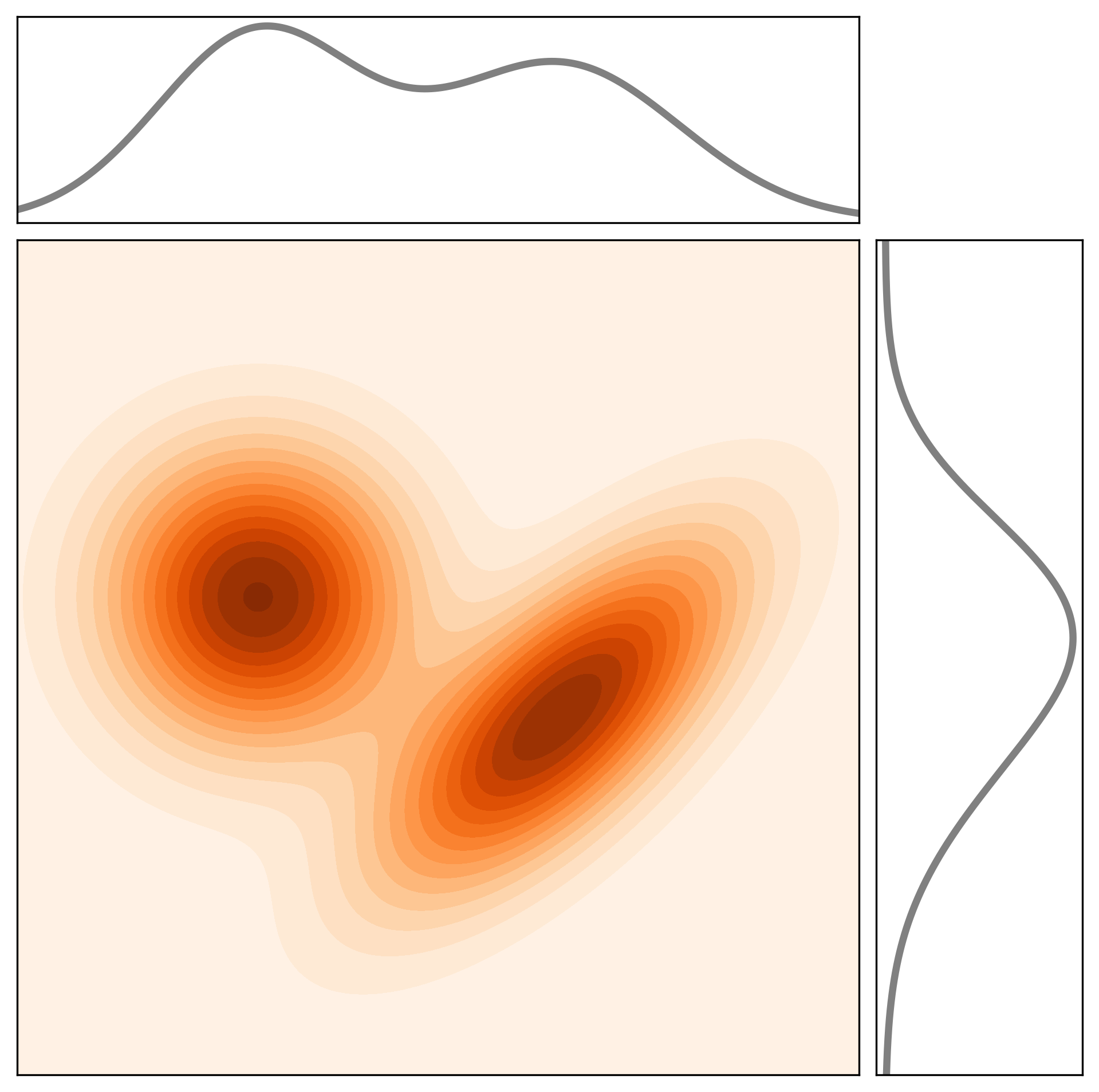}
  \end{center}
  \caption{Joint PDF $p_{X,Y}$ (in shades) and its marginals $p_X$ and $p_Y$ (in black lines). Figure from \citep{lipman2024flow}}.
  \label{fig:joint}
\end{wrapfigure}

Given two random variables $X,Y\in \Real^d$, their joint PDF $p_{X,Y}(x,y)$ has marginals 
\begin{equation}\label{e:marginals}
 \int p_{X,Y}(x,y)\dd y = p_X(x) \text{ and } \int p_{X,Y}(x,y)\dd x = p_Y(y). 
\end{equation}
See \cref{fig:joint} for an illustration of the joint PDF of two RVs in $\Real$ ($d=1$). 
%
% $X\sim p$ and $Y\sim q$, their \emph{joint distribution} $(X,Y)\sim \pi(X,Y)$ is a probability density over $\Real^d\times\Real^d$, with the marginals $\int \pi(x,y) \dd x=p(x)$ and $\int \pi(x,y)\dd x=q(y)$.
%
The \emph{conditional} PDF $p_{X|Y}$ describes the PDF of the random variable $X$ when conditioned on an event $Y=y$ with density $p_Y(y)>0$:
\begin{equation}\label{e:conditional}
    p_{X|Y}(x|y)\defe\frac{p_{X,Y}(x,y)}{p_Y(y)},
\end{equation} 
and similarly for the conditional PDF $p_{Y|X}$. Bayes' rule expresses the conditional PDF $p_{Y|X}$ with $p_{X|Y}$ by
\begin{equation}
    p_{Y|X}(y|x) = \frac{p_{X|Y}(x|y)p_Y(y)}{p_X(x)},
\end{equation}
for $p_X(x)>0$. 

The \emph{conditional expectation} $\E\brac{X | Y}$ is the best approximating \emph{function} $g_\star(Y)$ to $X$ in the least-squares sense: 
\begin{align}\nonumber
    g_\star &\defe  \argmin_{g:\Real^d\too\Real^d}\E\brac{\norm{X-g(Y)}^2} =  \argmin_{g:\Real^d\too\Real^d}\int \norm{x-g(y)}^2 p_{X,Y}(x,y)\dd x \dd y \\ \label{e:cond_E_g_star}
    &= \argmin_{g:\Real^d\too\Real^d} \int\brac{\textcolor{black}{\int \norm{x-g(y)}^2p_{X|Y}(x|y)\dd x}} p_Y(y)\dd y.
\end{align}
For $y\in \Real^d$ such that $p_Y(y)>0$ the conditional expectation function is therefore
\begin{equation}\label{e:cond_E_func}
    \E\brac{X|Y=y} \defe g_\star(y) = \int x p_{X|Y}(x|y) \dd x, %\qquad \metatriangleright\,\highlight{\text{conditional expectation (function $\Real^d\too\Real^d$)}}
\end{equation}
where the second equality follows from taking the minimizer of the inner brackets in \cref{e:cond_E_g_star} for $Y=y$, similarly to \cref{e:E}. 
Composing $g_\star$ with the random variable $Y$, we get 
\begin{equation}
    \E\brac{X|Y} \defe g_\star(Y), % \qquad \metatriangleright\,\highlight{\text{conditional expectation (RV $\Omega\too \Real^d$)} }
\end{equation}
which is a random variable in $\Real^d$.
Rather confusingly, both $\E\brac{X|Y=y}$ and $\E\brac{X|Y}$ are often called \emph{conditional expectation}, but these are different objects.
In particular, $\E\brac{X|Y=y}$ is a function $\Real^d\too\Real^d$, while $\E\brac{X|Y}$ is a random variable assuming values in $\Real^d$.
To disambiguate these two terms, our discussions will employ the notations introduced here.

The \emph{tower property} is an useful property that helps simplify derivations involving conditional expectations of two RVs $X$ and $Y$:
\begin{equation}\label{e:tower}
    \E\brac{\E\brac{X|Y}} = \E\brac{X} %\qquad \metatriangleright\,\highlight{\text{tower property}}
\end{equation}
Because $\E\brac{X|Y}$ is a RV, itself a function of the RV $Y$, the outer expectation computes the expectation of $\E\brac{X|Y}$.
The tower property can be verified by using some of the definitions above: 
\begin{align*}
    \E\brac{\E\brac{X|Y}} &= \int \parr{\int x p_{X|Y}(x|y) \dd x} p_Y(y) \dd y \\
    &\overset{\eqref{e:conditional}}{=} \int \int x p_{X,Y}(x,y) \dd x\dd y \\
    &\overset{\eqref{e:marginals}}{=} \int x p_X(x)\dd x= \E \brac{X}.
\end{align*}

Finally, consider a helpful property involving two RVs $f(X, Y)$ and $Y$, where $X$ and $Y$ are two arbitrary RVs.
Then, by using the Law of the Unconscious Statistician with \eqref{e:cond_E_func}, we obtain the identity
%
% %
% \begin{align*}
%     g_\star &=  \argmin_{g:\Real^d\too\Real^d} \E \brac{\norm{f(X,Y)-g(Y)}^2} \\
%     &\overset{\eqref{e:law_of_uncon_stat}}{=} \argmin_{g:\Real^d\too\Real^d} \int \brac{ \int \norm{f(x,y)-g(y)}^2 p_{X|Y}(x|y)\dd x } p_Y(y) \dd y,
% \end{align*}
% %
% which translates into the conditional expectation identity:
%
\begin{equation}\label{e:f_x_y_cond_y}
    \E\brac{f(X,Y)|Y=y} = \int f(x,y) p_{X|Y}(x|y) \dd x.
\end{equation}
\newpage
\section{A Proof of the Fokker-Planck equation}
\label{subsec:proof_fokker_planck}
In this section, we give here a self-contained proof of the Fokker-Planck equation which includes the continuity equation as a special case (\cref{thm:continuity_equation}). We stress that \textbf{this section is not necessary to understand the remainder of this document} and is mathematically more advanced. If you desire to understand where the Fokker-Planck equation comes from, then this section is for you.\\

\begin{theorem}[Fokker-Planck Equation]
\label{thm:fokker_planck_restated}
Let $p_t$ be a probability path with $p_0=\pinit$ and let us consider the SDE
\begin{align*}
    X_0\sim \pinit, \quad \dd X_t = u_t(X_t)\dd t + \sigma_t\dd W_t.
\end{align*}
Then $X_t$ has distribution $p_t$ for all $0\leq t\leq 1$ if and only if the \themebf{Fokker-Planck equation} holds:
    \begin{align}
    \label{e:fokker_planck_restated}
    \partial_t p_t(x) = -\divv (p_t u_t)(x)+\frac{\sigma_t^2}{2}\Delta p_t (x)\quad \text{ for all }x\in\R^d, 0\leq t\leq 1,
    \end{align}     
\end{theorem}

We start by showing that the Fokker-Planck is a necessary condition, i.e. if $X_t\sim p_t$, then the Fokker-Planck equation is fulfilled. The trick for the proof is to use \themebf{test functions} $f$, i.e. functions $f:\R^d\to\R$ that are infinitely differentiable ("smooth") and are only non-zero within a bounded domain (compact support). We use the fact that for arbitrary integrable functions $g_1,g_2:\mathbb{R}^d\to\mathbb{R}$ it holds that
\begin{align}
\label{eq:test_function_usecase}
g_1(x) = g_2(x)\text{ for all }x\in\mathbb{R}^d\quad \Leftrightarrow \quad \int f(x) g_1(x)\dd x = \int f(x) g_2(x)\dd x\text{ for all test functions }f
\end{align}
In other words, we can express the pointwise equality as equality of taking integrals. The useful thing about test functions is that they are smooth, i.e. we can take gradients and higher-order derivatives. In particular, we can use \themebf{integration by parts} for arbitrary test functions $f_1,f_2$:
\begin{align}
    \int f_1(x) \frac{\partial}{\partial x_i}f_2(x)\dd x = - \int f_2(x) \frac{\partial}{\partial x_i}f_1(x)\dd x
\end{align}
under the condition that $f_1,f_2$ and their product $f_1\cdot f_2$ is integrable. By using this together with the definition of the divergence and Laplacian (see \cref{eq:divergence_laplacian_definition}), we get the identities:
\begin{align}
\label{eq:integration_by_parts_divergence}
    \int \nabla f_1^T(x) f_2(x)\dd x =& -\int f_1(x)\divv(f_2)(x)\dd x \quad (f_1:\R^d\to\R,f_2:\R^d\to\R^d)\\
\label{eq:integration_by_parts_laplacian}
    \int f_1(x) \Delta f_2(x) \dd x =& \int f_2(x) \Delta f_1(x) \dd x\quad (f_1:\R^d\to\R,f_2:\R^d\to\R)
\end{align}

Now let's proceed to the proof. We use the stochastic update of SDE trajectories as in \cref{e:infinitesimal_updates_sdes}:
\begin{align}
    X_{t+h} =& X_{t}+hu_t(X_t) + \sigma_t(W_{t+h}-W_{t})+hR_t(h)\\
    \label{eq:approximate_condition}
    \approx &X_{t}+hu_t(X_t) + \sigma_t(W_{t+h}-W_{t})
\end{align}
where for now we simply ignore the error term $R_t(h)$ for readability as we will take $h\to 0$ anyway. We can then make the following calculation:
\begin{align*}
f(X_{t+h})-f(X_t)\overset{\eqref{eq:approximate_condition}}{=}&f(X_{t}+hu_t(X_t) + \sigma_t(W_{t+h}-W_{t}))-f(X_t)\\
\overset{(i)}{=}&\nabla f(X_t)^T\left(hu_t(X_t) + \sigma_t(W_{t+h}-W_{t}))\right)\\&+\frac{1}{2}\left(hu_t(X_t) + \sigma_t(W_{t+h}-W_{t}))\right)^T\nabla^2 f(X_t)\left(hu_t(X_t) + \sigma_t(W_{t+h}-W_{t}))\right)\\
\overset{(ii)}{=}&h\nabla f(X_t)^Tu_t(X_t) + \sigma_t\nabla f(X_t)^T(W_{t+h}-W_{t})\\&+\frac{1}{2}h^2u_t(X_t)^T\nabla^2 f(X_t)u_t(X_t)+ h\sigma_tu_t(X_t)^T\nabla^2 f(X_t)(W_{t+h}-W_{t})+\\&+\frac{1}{2}\sigma_t^2(W_{t+h}-W_t)^T\nabla^2 f(X_t)(W_{t+h}-W_t)
\end{align*}
where in (i) we used a 2nd Taylor approximation of $f$ around $X_t$ and in (ii) we used the fact that the Hessian $\nabla^2 f$ is a symmetric matrix. Note that $\mathbb{E}[W_{t+h}-W_t|X_t]=0$ and $W_{t+h}-W_{t}|X_t\sim\mathcal{N}(0,hI_d)$. Therefore
\begin{align*}
    &\mathbb{E}[f(X_{t+h})-f(X_t)|X_t]\\
=&h\nabla f(X_t)^Tu_t(X_t)+\frac{1}{2}h^2u_t(X_t)^T\nabla^2 f(X_t)u_t(X_t)+\frac{h}{2}\sigma_t^2\mathbb{E}_{\epsilon_t\sim\mathcal{N}(0,I_d)}[\epsilon_t^T\nabla^2 f(X_t)\epsilon_t]\\
\overset{(i)}{=}&h\nabla f(X_t)^Tu_t(X_t)+\frac{1}{2}h^2u_t(X_t)^T\nabla^2 f(X_t)u_t(X_t)+\frac{h}{2}\sigma_t^2\text{trace}(\nabla^2 f(X_t))\\
\overset{(ii)}{=}&h\nabla f(X_t)^Tu_t(X_t)+\frac{1}{2}h^2u_t(X_t)^T\nabla^2 f(X_t)u_t(X_t)+\frac{h}{2}\sigma_t^2\Delta f(X_t)
\end{align*}
where in $(i)$ we used the fact that $\mathbb{E}_{\epsilon_t\sim\mathcal{N}(0,I_d)}[\epsilon_t^T A\epsilon_t]=\text{trace}(A)$ and in $(ii)$ we used the definition of the Laplacian and the Hessian matrix. With this, we get that 
\begin{align*}
&\partial_t \mathbb{E}[f(X_t)]\\
=&\lim\limits_{h\to 0}
\frac{1}{h}\mathbb{E}[f(X_{t+h})-f(X_t)]\\
=&\lim\limits_{h\to 0}
\frac{1}{h}\mathbb{E}[\mathbb{E}[f(X_{t+h})-f(X_t)|X_t]]\\
=&\mathbb{E}[\lim\limits_{h\to 0}\frac{1}{h}\left(
h\nabla f(X_t)^Tu_t(X_t)+\frac{1}{2}h^2u_t(X_t)^T\nabla^2 f(X_t)u_t(X_t)+\frac{h}{2}\sigma_t^2\Delta f(X_t)
\right)]\\
=&\mathbb{E}[\nabla f(X_t)^Tu_t(X_t)+\frac{1}{2}\sigma_t^2\Delta f(X_t)]\\
\overset{(i)}{=}&\int \nabla f(x)^Tu_t(x)p_t(x)\dd x+\int \frac{1}{2}\sigma_t^2\Delta f(x)p_t(x)\dd x\\
\overset{(ii)}{=}&-\int f(x)\divv(u_t p_t)(x)\dd x+\int \frac{1}{2}\sigma_t^2 f(x)\Delta p_t(x)\dd x\\
=&\int f(x)\left(-\divv(u_t p_t)(x)+\frac{1}{2}\sigma_t^2\Delta p_t(x)\right)\dd x
\end{align*}
where in (i) we used the assumption that $p_t$ as the distribution of $X_t$ and in (ii) we used \cref{eq:integration_by_parts_divergence} and \cref{eq:integration_by_parts_laplacian}. Note that to use this, we require integrability of the product $p_t(x)u_t(x)$, i.e. such that 
\begin{align*}
\int p_t(x)\|u_t(x)\|\dd x <\infty
\end{align*}
Note that this condition almost always holds in machine learning (bounded data and functions because of numerical precision limits). Therefore, it holds that
\begin{align}
\partial_t\mathbb{E}[f(X_t)] =& \int f(x)\left(-\divv (p_t u_t)(x)+\frac{\sigma_t^2}{2}\Delta p_t (x)\right)\dd x\quad (\text{for all }f\text{ and }0\leq t\leq 1)\\
\overset{(i)}{\Leftrightarrow}\quad \partial_t\int f(x) p_t(x)\dd x =& \int f(x)\left(-\divv (p_tu_t)(x)+\frac{\sigma_t^2}{2}\Delta p_t (x)\right)\dd x\quad (\text{for all }f\text{ and }0\leq t\leq 1)\\
\overset{(ii)}{\Leftrightarrow}\quad \int f(x)\partial_t p_t(x)\dd x =& \int f(x)\left(-\divv (p_t u_t)(x)+\frac{\sigma_t^2}{2}\Delta p_t (x)\right)\dd x\quad (\text{for all }f\text{ and }0\leq t\leq 1)\\
\overset{(iii)}{\Leftrightarrow}\quad \partial_t p_t(x) =& -\divv (p_t u_t)(x)+\frac{\sigma_t^2}{2}\Delta p_t(x)\quad (\text{for all }x\in\R^d, 0\leq t\leq 1)
\end{align} 
where in (i) we used the assumption that $X_t\sim p_t$, in (ii) we swapped the derivative with the integral and (iii) we used \cref{eq:test_function_usecase}
. This completes the proof that the Fokker-Planck equation is a necessary condition.\\

Finally, we explain why it is also a sufficient condition. The Fokker-Planck equation is a partial differential equation (PDE). More specifically, it is a so-called \emph{parabolic partial differential equation}. Similar to \cref{thm:ode_existence_and_uniqueness}, such differential equations have a unique solution given fixed initial conditions (see e.g. \citep[Chapter 7]{evans2022partial}). Now, if \cref{e:fokker_planck_restated} holds for $p_t$, we just shown above that it must also hold for true distribution $q_t$ of $X_t$ (i.e. $X_t\sim q_t$) - in other words, both $p_t$ and $q_t$ are solutions to the parabolic PDE. Further, we know that the initial conditions are the same, i.e. $p_0=q_0=\pinit$ by construction of an interpolating probability path. Hence, by uniqueness of the solution of the differential equation, we know that $p_t=q_t$ for all $0\leq t\leq 1$ - which means $X_t\sim q_t=p_t$ and which is what we wanted to show.
\section{Existence and Uniqueness of Continuous-time Markov chains}
\label{appendix:existence_uniquenss_ctmcs}
We prove \cref{thm:ctmc_existence_and_uniqueness} in this section.
\begin{proof}
\textbf{Uniqueness:} We need to show that there can be only one transition kernel $p_{t'|t}(X_{t'}=y|X_{t}=x)$ that satisfies \cref{eq:ctmc_ode}. As a first step, we realize that \cref{eq:ctmc_ode} implies that
\begin{align}
    &\frac{\dd}{\dd t'}p_{t'|t}(X_{t'}=y|X_{t}=x)\\
=&\frac{\dd}{\dd h}p_{t'+h|t}(X_{t'+h}=y|X_{t}=x)_{|h=0}\\
=&\frac{\dd}{\dd h}\left[\sum\limits_{z\in S}p_{t'+h|t'}(X_{t'+h}=y|X_{t'}=z)p_{t'|t}(X_{t'}=z|X_{t}=x)\right]_{|h=0}\\
=&\sum\limits_{z\in S}Q_{t'}(y|z)p_{t'|t}(X_{t'}=z|X_{t}=x)
\end{align}
For fixed $x,t$, one can consider $t'\mapsto p_{t'|t}(X_{t'}=y|X_t=x)$ as a vector-valued function and the above is a linear ODE of that function (the Kolmgorov forward equation, in fact, see \cref{prop:kfe_ctmc}) with a known initial condition, i.e. $p_{t|t}(X_{t}=y|X_{t}=x)=\delta_{y}(x)$. As we know, every linear ODE has a unique solution (see \cref{thm:ode_existence_and_uniqueness}), therefore $p_{t'|t}(X_{t'}=y|X_{t}=x)$ must also be unique.

\textbf{Existence: }Conversely, any linear ODE has a solution, i.e. we know that for every $x,t$ there is a $p_{t'|t}(X_{t'}=y|X_{t}=x)$ such that 
\begin{align}
\label{eq:initial_conditional_ctmc_ode}
    p_{t|t}(X_{t}=y|X_{t}=x)&=\delta_{y}(x)\\
\frac{\dd}{\dd t'}p_{t'|t}(X_{t'}=y|X_{t}=x)&=\sum\limits_{z\in S}Q_{t'}(y|z)p_{t'|t}(X_{t'}=z|X_t=x)
\end{align}
For $t'=t$, this implies \cref{eq:ctmc_ode} in particular. It remains to show that $p_{t'|t}(X_{t'}=y|X_{t}=x)$ is a valid transition kernel in this case, i.e. the following 3 properties must hold:
\begin{align}
    \sum\limits_{y\in S}p_{t'|t}(X_{t'}=y|X_{t}=x)=&1\\
    p_{t'|t}(X_{t'}=y|X_{t}=x)\geq& 0\\
    \sum\limits_{z\in S}p_{t_{2}|t_{1}}(X_{t_2}=y|X_{t_1}=z)p_{t_1|t_0}(X_{t_1}=z|X_{t_0}=x)=&p_{t_2|t_0}(y|x)
\end{align}
To the first property, one can observe that it holds for $t'=t$ by \cref{eq:initial_conditional_ctmc_ode} and that 
\begin{align}
&\frac{\dd}{\dd t'}\sum\limits_{y\in S}p_{t'|t}(X_{t'}=y|X_{t}=x)\\
=&\sum\limits_{y\in S}\frac{\dd}{\dd t'}p_{t'|t}(X_{t'}=y|X_{t}=x)\\
=&\sum\limits_{z\in S}\left[\sum\limits_{y\in S}Q_{t'}(y|z)\right]p_{t'|t}(X_{t'}=z|X_t=x)\\
=&0
\end{align}
where we used the fact that the columns of rate matrices sum to $0$. To show the second property, note that it holds at time $t'=t$. Further, whenever $p_{t'|t}(X_{t'}=y|X_{t}=x)=0$, it must hold that 
\begin{align*}
\frac{\dd}{\dd t'}p_{t'|t}(X_{t'}=y|X_{t}=x)&=\sum\limits_{z\neq y}\underbrace{Q_{t'}(y|z)}_{\geq 0}p_{t'|t}(X_{t'}=z|X_t=x)\\
&\geq 0
\end{align*}
Therefore, whenever $p_{t'|t}(X_{t'}=y|X_{t}=x)=0$, it can only increase. Therefore, $p_{t'|t}(X_{t'}=y|X_{t}=x)$ will never be negative.\\

To show the third property, define $q_{t_2|t_0}(y|x)$ to be 
\begin{align*}
q_{t_2|t_0}(y|x)=&\sum\limits_{z\in S}p_{t_{2}|t_{1}}(X_{t_2}=y|X_{t_1}=z)p_{t_1|t_0}(X_{t_1}=z|X_{t_0}=x)
\end{align*}
Then we know that 
\begin{align*}
    q_{t_2=t_1|t_0}(y|x)=&\sum\limits_{z\in S}\delta_{y}(z)p_{t_1|t_0}(X_{t_1}=z|X_{t_0}=x)=p_{t_1|t_0}(X_{t_1}=y|X_{t_0}=x)
\end{align*}
and
\begin{align*}
\frac{\dd}{\dd t_2}q_{t_2|t_0}(y|x)=&\sum\limits_{z\in S}\frac{\dd}{\dd t_2}p_{t_{2}|t_{1}}(X_{t_2}=y|X_{t_1}=z)p_{t_1|t_0}(X_{t_1}=z|X_{t_0}=x)\\
=&\sum\limits_{z\in S}\sum\limits_{\tilde{z}\in S}Q_{t_2}(y|\tilde{z})p_{t_2|t_1}(X_{t_2}=\tilde{z}|X_{t_1}=z)p_{t_1|t_0}(X_{t_1}=z|X_{t_0}=x)\\
=&\sum\limits_{\tilde{z}\in S}Q_{t_2}(y|\tilde{z})\left[\sum\limits_{z\in S}p_{t_2|t_1}(X_{t_2}=\tilde{z}|X_{t_1}=z)p_{t_1|t_0}(X_{t_1}=z|X_{t_0}=x)\right]\\
=&\sum\limits_{\tilde{z}\in S}Q_{t_2}(y|\tilde{z})q_{t_2|t_0}(\tilde{z}|x)
\end{align*}
This shows that $p_{t_2|t_0}(z|x)$ and $q_{t_2|t_0}(z|x)$ fulfill the same ODE. Hence, it must hold 
\begin{align*}
\sum\limits_{z\in S}p_{t_{2}|t_{1}}(X_{t_2}=y|X_{t_1}=z)p_{t_1|t_0}(X_{t_1}=z|X_{t_0}=x)=&q_{t_2|t_0}(y|x)=p_{t_2|t_0}(y|x)
\end{align*}
This shows the third property. So $p_{t'|t}(y|x)$ is indeed the transition kernel satisfying \cref{eq:ctmc_ode}. This finishes the proof.
\end{proof}

% \newpage
% \subfile{subfiles/appendix_neural_odes}
\newpage
\section{Additional Perspectives on VAEs}
\label{sec:vae_appdx}
In this section, we expand on the treatment of VAEs presented in the main text and provide a variational derivation of the total VAE loss from \cref{eq:total_vae_loss}. As a first step, notice that both the encoder and decoder give rise to a joint distribution over both $x$ and the latent $z$, viz.,
\begin{align*}
q_\phi(x,z) &= \pdata(x)q_\phi(\cdot |x) \quad& (\text{encoder joint}) \\
p_\theta(x,z) &= p_\theta(x|z)\prior(z) \quad& (\text{decoder joint})
\end{align*}
We might therefore conceptualize training the VAE as learning $\phi$ and $\theta$ so that the encoder and decoder joint distributions are reasonably similar. We can do this via the KL-divergence of the joint latent and data distribution:
\begin{equation}
\begin{aligned}
    \dkl{q_\phi(x,z)}{p_\theta(x,z)} &= \dkl{\pdata(x)q_\phi(z\mid x)}{p_\theta(x\mid z)\prior(z)}\\
    &= \EE_{\blacksquare} \left[\log \left(\frac{\pdata(x)q_\phi(z\mid x)}{p_\theta(x\mid z)\prior(z)}\right)\right]\\
    &= \textcolor{BrickRed}{\EE_{\blacksquare} \left[\log \pdata(x)\right]} + \textcolor{RoyalBlue}{\EE_{\blacksquare} \left[\log \left(\frac{q_\phi(z\mid x)}{\prior(z)}\right)\right]} - \textcolor{ForestGreen}{\EE_{\blacksquare} \left[\log p_\theta(x\mid z)\right]}\\
    \blacksquare &= x \sim \pdata(x)\, z\sim q_\phi(z|x).
\end{aligned}
\label{eq:lvae_init}
\end{equation}
Let us now examine each of the three remaining terms in turn. First, we find that
\begin{equation}
    \textcolor{BrickRed}{\EE_{\blacksquare} \left[\log \pdata(x)\right] =  \textcolor{BrickRed}{\EE_{x\sim \pdata(x)} \left[\log \pdata(x)\right]} = C},
\end{equation}
for some constant $C$ independent of $\phi$ and $\theta$. Next, we find that
\begin{equation}
    \textcolor{RoyalBlue}{\EE_{\blacksquare} \left[\log \left(\frac{q_\phi(z\mid x)}{\prior(z)}\right)\right] = \mathbb{E}_{x \sim \pdata(x)}\left[\dkl{q_\phi(z\mid x)}{\prior(z)}\right]}
\end{equation}
encourages $q_\phi(z\mid x)$ to resemble the prior $\prior(z)$. Finally, we find that
\begin{equation}
    -\textcolor{ForestGreen}{\EE_{x \sim \pdata(x)\, z\sim q_\phi(z|x)} \left[\log p_\theta(x\mid z)\right]}
\end{equation}
corresponds the average negative log-likelihood, and thus serves as to minimize the reconstruction loss. Ignoring the constant term, we combine the \textcolor{RoyalBlue}{prior penalty} and \textcolor{ForestGreen}{reconstruction} terms to obtain that the VAE loss is actually simply the KL-divergence in joint data and latent space:
\begin{align}
    \label{eq:l_vae}
    \mathcal{L}_{\text{VAE}}(\phi, \theta) &= \underbrace{\textcolor{RoyalBlue}{\mathbb{E}_{x \sim \pdata(x)}\left[\dkl{q_\phi(z\mid x)}{\prior(z)}\right]}}_{\text{prior enforcement loss}} - \underbrace{\textcolor{ForestGreen}{\EE_{x \sim \pdata(x)\, z\sim q_\phi(z|x)} \left[\log p_\theta(x\mid z)\right]}}_{\text{reconstruction loss}}\\
    &=\dkl{q_\phi(x,z)}{p_\theta(x,z)}+\text{const}
\end{align}
Therefore, we can interpret the VAE as a KL-divergence in the joint space of latents and images.

\paragraph{VAEs as generative models.} We now explain how one could interpret VAEs as generative models. We could generate a sample by setting $z\sim \prior=\mathcal{N}(0,I_k)$ and sampling $x\sim p_\theta(\cdot|z)$ from the decoder. The resulting distribution that we would get is given by:
\begin{equation*}
    p_\theta(x) = \int_z p_\theta(x|z) \prior(z)\dd z
\end{equation*}
We now want to demonstrate that the VAE learns to approximately sample from $p_\theta$. To show this, we need the following result:
\begin{proposition}[Chain rule]
Let $q(x,z), p(x,z)$ be distributions over two variables $x\in\mathbb{R}^{l_1},z\in\mathbb{R}^{l_2}$. Then, it holds that:
\begin{align*}
\dkl{q(z,x)}{p(z,x)}= \dkl{q(x)}{p(x)}+\mathbb{E}_{x\sim q}\left[\dkl{q(z|x)}{p(z|x)}\right].
\end{align*}
In particular, as the second summand is non-negative due to \cref{eq:kl_non_negative}, we obtain the data-processing inequality
\begin{align}
\label{eq:kl_upper_bound}
\dkl{q(x)}{p(x)}\leq \dkl{q(z,x)}{p(z,x)}.
\end{align}
\label{prop:chain_rule}
\end{proposition}
\begin{proof}
\begin{align*}
   \dkl{q(z,x)}{p(z,x)}&=\mathbb{E}_{q}\left[
   \log \frac{q(z,x)}{p(z,x)}
   \right]\\
&=\mathbb{E}_{(x,z) \sim q}\left[
   \log \frac{q(z|x)}{p(z|x)}\frac{q(x)}{p(x)}
   \right]\\
&=\mathbb{E}_{(x,z) \sim q}\left[
   \log \frac{q(z|x)}{p(z|x)}\right]+\mathbb{E}_{x\sim q} \left[\log \frac{q(x)}{p(x)}
   \right]\\
&=\dkl{q(x)}{p(x)}+\mathbb{E}_{x\sim q}\left[\dkl{q(z|x)}{p(z|x)}\right]
\end{align*}
where we have repeatedly applied the definition of KL divergence.
\end{proof}
By \cref{prop:chain_rule}, we can now show that
\begin{equation}
    \mathcal{L}_{\text{VAE}}(\phi, \theta) = \dkl{q_\phi(x,z)}{p_\theta(x,z)} +\text{const}\ge \dkl{\pdata(x)}{p_\theta(x)} +\text{const}
    \label{eq:amort_zero}
\end{equation}
where we used the fact the $x$-marginal of $q_\phi(x,z)$ is $\pdata$. In other words, the VAE loss minimizes an upper bound on the KL-divergence between the data distribution $\pdata$ and the distribution generated by the VAE. Hence, we can look at VAEs as generative models in their own right. In the same way, we can show that 
\begin{equation}
    \mathcal{L}_{\text{VAE}}(\phi, \theta) = \dkl{q_\phi(x,z)}{p_\theta(x,z)} +\text{const}\ge \dkl{q_\phi(z)}{\prior(z)} +\text{const}
    \label{eq:amort_one}
\end{equation}
In other words, the VAE objective minimize an upper bound to the KL-divergence between latent distribution and the prior.
\paragraph{Why not stop at VAEs?}
Per the discussion above, VAEs can be realized as generative models in their own right, with the encoder simply existing to facilitate the training of a complementary decoder which transforms a Gaussian into the desired data distribution. Samples could then be obtained by sampling $z \sim \prior$ and then $x \sim p_\theta(x|z)$. Why then, we do insist on training a separate generative model within the learned latent space? The answer has to do with the so-called \textbf{amortization gap} between the left and right hand sides of both \cref{eq:amort_zero} and \cref{eq:amort_one}, corresponding precisely to the gap in the information processing inequality. This gap is zero if and only if $q_\phi(z|x)=p_\theta(z|x)$, in which case the encoder represents the true posterior. Thus, while e.g., $\dkl{q_\phi(x,z)}{p_\theta(x,z)}$ is minimized implies $\dkl{q_\phi(z)}{\prior(z)}$ is minimized (see \cref{eq:amort_one}, a decrease in the former does not necessarily imply an equal decrease in the latter. Consequently, at the end of training, it is simultaneously true that both $\dkl{q_\phi(x,z)}{p_\theta(x,z)}$ and the amortization gap  
\begin{equation}
    \dkl{q_\phi(x,z)}{p_\theta(x,z)} - \dkl{q_\phi(z)}{\prior(z)}
\end{equation}
 are not completely minimized, so that $q_\phi(z) \neq \prior(z)$. Finally, observe that during training, the decoder learns to reconstruct from $q_\phi(z)$ rather than $\prior(z)$, so that switching to reconstructing from $\prior(z)$ during inference would amount to going out of distribution from training. In practice however, this mismatch is a feature rather than a bug. Practice has shown flow and diffusion models to be more capable models in general than the convolutional stacks used to implement the VAE decoder, so that it makes sense to farm off some of the generative complexity to the latent generative model. We return to this line of discussion later on in the discussion. Additionally, and beyond the scope of these notes, variational formulations of diffusion and flow models realize these modeling families as VAEs in their own right.
\paragraph{The evidence lower bound.}
    Properly rearranged, the terms within \cref{eq:lvae_init} can present various complementary perspectives. One is the so-called \themebf{evidence lower bound}, which we extract as follows. Observe that for fixed $x$
    \begin{equation}
    \begin{aligned}
        \EE_{z\sim q_\phi(z|x)} \left[\log \left(\frac{q_\phi(z\mid x)}{p_\theta(x\mid z)\prior(z)}\right)\right]
        &= \EE_{z\sim q_\phi(z|x)} \left[\log \left(\frac{q_\phi(z\mid x)}{p_\theta(z\mid x)}\right)\right] - \log p_\theta(x)\\
        &= \dkl{q_\phi(z\mid x)}{p_\theta(z\mid x)} - \log p_\theta(x)
    \end{aligned}
    \label{eq:elbo_one}
    \end{equation}
    where the first equality is obtained from 
    \begin{equation*}
        p_\theta(z\mid x) = \frac{p_\theta(x\mid z)\prior(z)}{p_\theta(x)}.
    \end{equation*}
    We may thus rearrange \cref{eq:elbo_one} to obtain
    \begin{equation}
        \EE_{z\sim q_\phi(z|x)} \left[\log \left(\frac{p_\theta(x\mid z)\prior(z)}{q_\phi(z\mid x)}\right)\right] + \dkl{q_\phi(z\mid x)}{p_\theta(z\mid x)} = \log p_\theta(x),
    \end{equation}
    from which it follows that 
    \begin{equation}
        \underbrace{\EE_{z\sim q_\phi(z|x)} \left[\log \left(\frac{p_\theta(x\mid z)\prior(z)}{q_\phi(z\mid x)}\right)\right]}_{\triangleq \,\text{ELBO}(x;\phi, \theta)} \le \underbrace{\log p_\theta(x)}_{\text{evidence}}.
    \end{equation}
    The left-hand side is therefore commonly referred to as the evidence lower bound, or ELBO. We may now rewrite $\mathcal{L}_{\text{VAE}}$ from \cref{eq:l_vae} in terms of the ELBO via
    \begin{equation}
        \begin{aligned}
            \mathcal{L}_{\text{VAE}}
            &= \dkl{q_\phi(x,z)}{p_\theta(x,z)}+\text{const}\\
            &= \mathbb{E}_{x \sim \pdata} \mathbb{E}_{z \sim q_\phi(z|x)} \left[\log \left(\frac{\pdata(x)q_\phi(z\mid x)}{p_\theta(x\mid z)\prior(z)}\right)\right] + \text{const}\\
            &= \mathbb{E}_{x \sim \pdata} \left[\log \pdata(x) - \text{ELBO}(x; \phi, \theta)\right] + \text{const}\\
            &= -\mathbb{E}_{x \sim \pdata} \left[\text{ELBO}(x; \phi, \theta)\right] \underbrace{- H(\pdata) + \text{const}}_{\text{const}}\\
            &= -\mathbb{E}_{x \sim \pdata} \left[\text{ELBO}(x; \phi, \theta)\right] + \text{const}
        \end{aligned}
    \end{equation}
    so that the original VAE objective can be seen as simply trying to maximize the expected ELBO. Finally, let's consider what occurs in the limit that we train our VAE perfectly.
    
\begin{remarkbox}[What Happens When $q_\phi(x, z) \approx p_\theta(x, z)$?]
    First, note that the sampling distribution used to train our latent generative model is given by the marginal
    \begin{align*}
        q_{\phi}(z) = \int_x q_\phi(z | x) \pdata(x)\, \d x.
    \end{align*}
    If $q_\phi(x, z) = p_\theta(x, z)$, then in particular
    \begin{align*}
            q_{\phi}(z) = p_\theta(z) = \prior(z).
    \end{align*}
    Thus, $q_\phi(x, z) \approx p_\theta(x, z)$ \textbf{implies regularization of the latent sampling distribution}. Second, $q_\phi(x, z) \approx p_\theta(x, z)$ implies that the variational approximation $p_\theta(x \mid z) \approx q_\phi(x \mid z)$ is good, and in turn \textbf{implies low reconstruction error}.
\end{remarkbox}

\begin{remarkbox}[What's Variational About VAEs?]
    Why can't we simply take $q_\phi(\cdot \mid x) = p_\theta(\cdot \mid x)$, thereby guaranteeing $q_\phi(x, z) = p_\theta(x, z) = 0$? The reason is that while we know the \themebf{likelihood} $p_\theta(x \mid z)$, the \themebf{posterior} 
    \begin{align*}
        p_\theta(z \mid x) = \tfrac{p_\theta(x \mid z)\prior(z)}{p_\theta(x)}
    \end{align*} 
    is generally intractable, as we lack access to the likelihood $p_\theta(x)$. The presence of \textit{variational} in VAE is thus due to the fact that $q_\phi(\cdot \mid x)$ serves as a substitute, or \themebf{variational approximation}, of the intractable posterior $p_\theta(\cdot \mid x)$.
\end{remarkbox}

\begin{figure}[!t]
\centering
\begin{subfigure}[c]{.49\textwidth}
  \centering
  \includegraphics[width=0.95\linewidth]{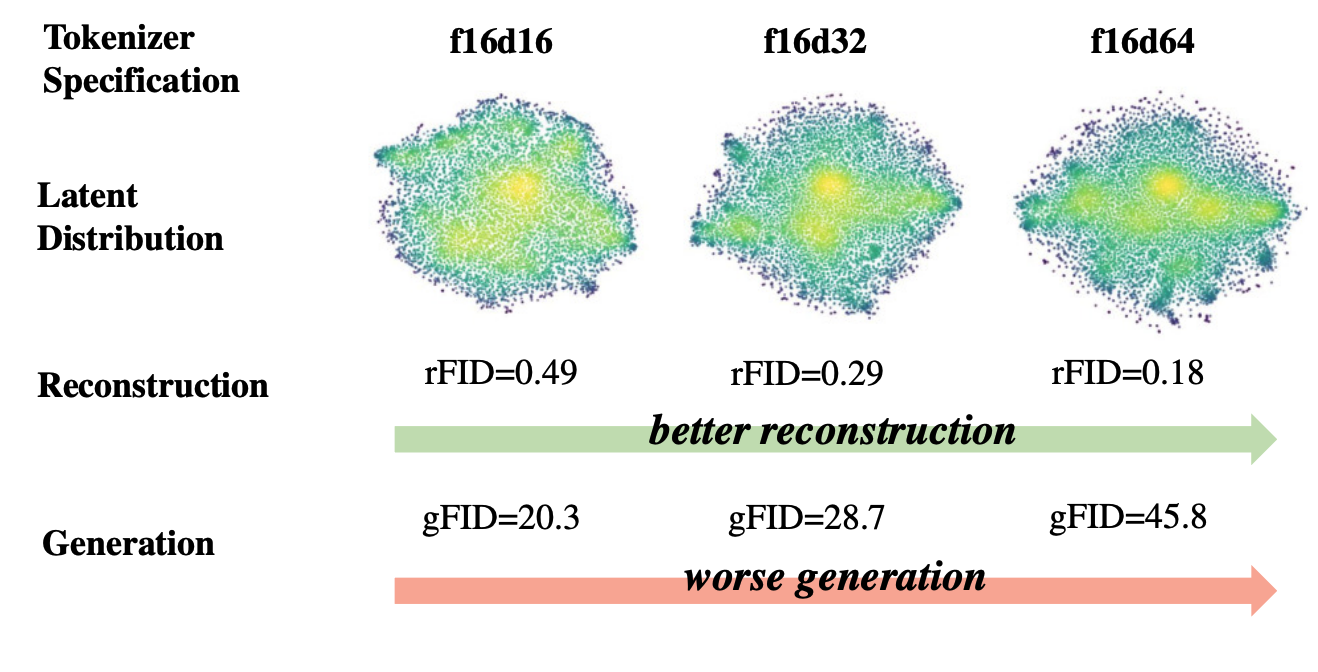}
\end{subfigure}
\begin{subfigure}[c]{.49\textwidth}
  \centering
  \includegraphics[width=0.95\linewidth]{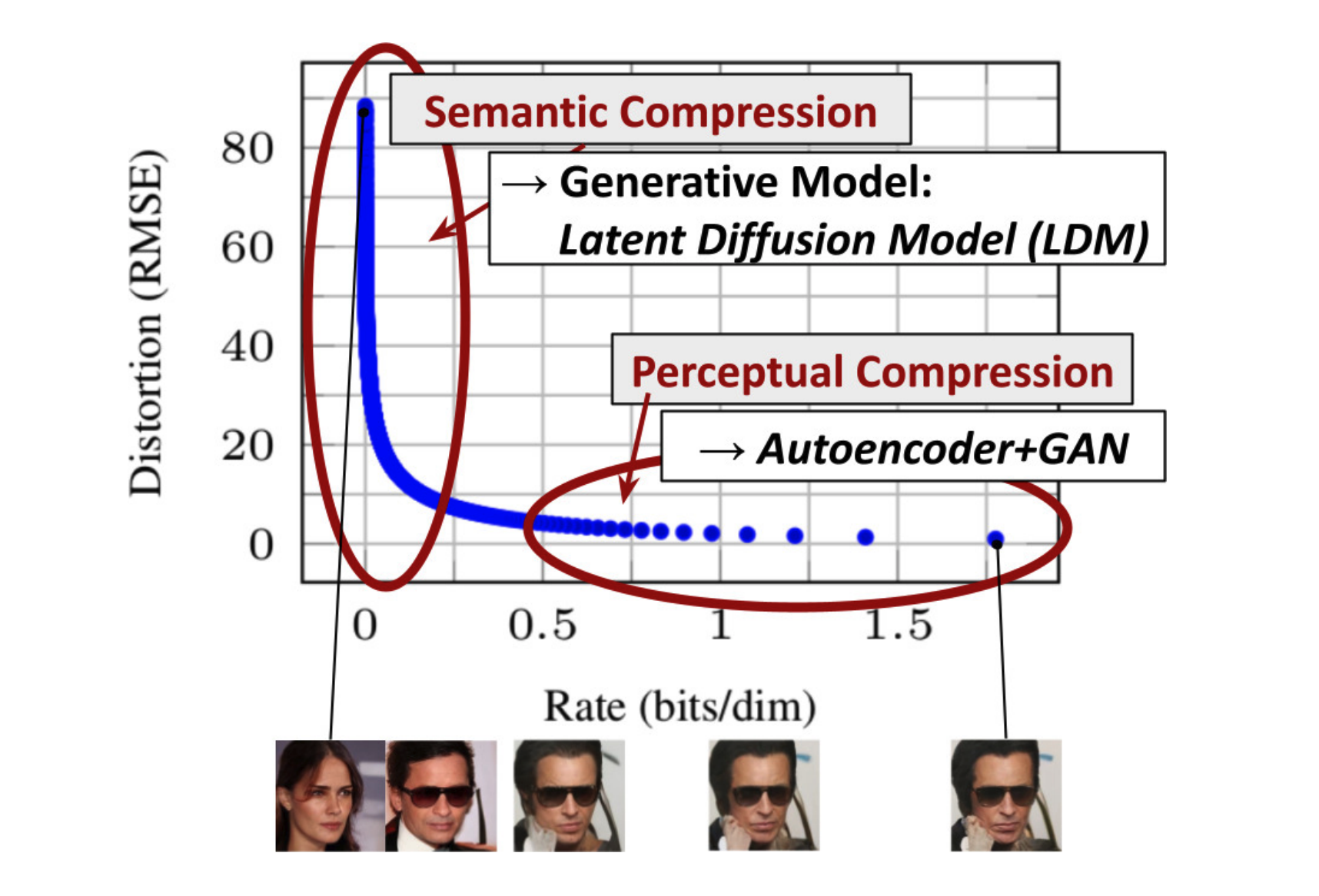}
\end{subfigure}%
\caption{Right: The tradeoff between between gFID and rFID, figure taken from \citep{rfid_gfid}. Here, $f$ denotes the downsampling factor, and $d$ denotes the latent channel dimension. Right: Distortion (reconstruction quality) vs rate, taken from \citep{ho2020denoising, rombach2022high}. We remark that this particular curve was generated using a DDPM (itself a type of VAE). While certain technical subtleties in the distortion and rate computations may differ from the imprecise definition presented in this text, the overall intuition remains the same.}
\label{fig:autoencoder}
\end{figure}

\paragraph{Reconstruction vs Generation.} Given an encoder $q_\phi(z | x)$, decoder $p_\theta(x | z)$, and latent generative model $r_\psi$ trained to sample from $q_\phi(z)$, we may consider the following two generative models
\begin{align*}
r_{\psi, \theta}^{\text{recon}}(x_{\text{out}})
&= \int_{z, x_{\text{in}}}
    p_\theta(x_{\text{out}} \mid z)\,
    q_\phi(z \mid x_{\text{data}})\,
    \pdata(x_{\text{data}})
    \d z
    \d x_{\text{in}}
& (\text{reconstruction sampler}) \\
r_{\psi, \phi}^{\text{gen}}(x_{\text{out}}) &= \int_{z_{\text{gen}}} p_\theta(x_{\text{out}} | z_{\text{gen}})r_\psi(z_{\text{gen}})\, \d z_{\text{gen}} \quad& (\text{generative sampler})
\end{align*}
In other words, the reconstruction sampler starts at $x_{\text{data}} \in \pdata$ encodes to $z$, and decodes to $x_{\text{out}}$, while the generative sampler starts from $z_{\text{gen}} \in r_{\psi}$ from the generative model, and then passes through the decoder. By computing the Fréchet inception distance of the two respective samplers' distributions to $\pdata$, we obtain the \themebf{reconstruction-FID} (rFID) and \themebf{generative-FID} (gFID). One might also consider measuring the quality of the reconstruction sampler via the average \themebf{distortion} (root mean square error of reconstruction), although such a metric would not make sense for the generative sampler. As it turn out, there is a natural tension between the quality of the reconstruction sampler, and the quality of the generative sampler. Low rFID (a high quality reconstruction sampler) generally indicates low information loss in the latent, so that the latent distribution $q_\phi(z)$ largely resembles $\pdata$, and so that the task of learning the latent generative model is likely more difficult, raising gFID. Conversely, high rFID generally indicates high information loss, and an easier latent distribution $q_\phi(z)$ to learn, thereby lowering gFID. This phenomena is visualized in \cref{fig:autoencoder}.

\paragraph{The Division of Labor.} The reconstruction-generative sampler tradeoff forces us to consider how information loss should be divided up between the autoencoder and the latent generative model $r_\psi$. Intuitively, $r_{\psi}$, via some learned vector field $u_t^\psi(z_t)$, transports a standard Gaussian to $q_\phi(z) \approx \prior$, after which the decoder $p_\theta(x|z)$ transports $q_\phi(z)$ to $\pdata$. Let us now (imprecisely) define the \themebf{rate} as the degree to which the latent distribution $q_\phi(z)$ matches the matches the $\prior(z)$, and by extension, the degree to which the task of generation is farmed off to the latent generative model.\footnote{We defer a more technical discussion of the rate to the next subsection.} This division of labor can be visualized by plotting the Pareto frontier between rate and distortion, as shown in \cref{fig:autoencoder}. In particular, when the rate is high, the distortion is low, and vice versa, offering a second perspective on the preceding discussion of reconstruction versus generation sampler quality. We culminate our discussion in the following insight.

\begin{intuitionbox}[The Division of Labor]
    The key insight from \cref{fig:autoencoder} is that an optimal division of labor exists at the ``knee'' of the Pareto frontier, at which point we obtain low rate (high compression!) without high distortion. In other words, such a point corresponds to a level of compression which simultaneously reduces the difficulty of training the underlying generative model while preserving reasonable reconstruction quality.
\end{intuitionbox}

% \begin{remarkbox}[Flow-Based Decoders]
    
% \end{remarkbox}

% \begin{remarkbox}[Quantization]
    
% \end{remarkbox}
\section{A Guide to the  Diffusion Model Literature}
\label{subsec:guide_to_flow_matching_literature}

There is a whole family of models around diffusion models and flow matching in the literature. When you read these papers, you will likely find a different (but equivalent) way of presenting the material from this class. This makes it sometimes a little confusing to read these papers. For this reason, we want to give a brief overview over various frameworks and their differences and put them also in their historical context. \textbf{This is not necessary to understand the remainder of this document} but rather intended to be a support for you in case you read the literature.

\paragraph{Discrete time vs. continuous time.} The first denoising diffusion model papers \citep{sohl2015deep, song2019generative, ho2020denoising} did not use SDEs but constructed Markov chains in \themebf{discrete time}, i.e. with time steps $t=0,1,2,3,\dots$. To this date, you will find a lot of works in the literature working with this discrete-time formulation. While this construction is appealing due to its simplicity, the disadvantage of the time-discrete approach is that it forces you to choose a time discretization before training. Further, the loss function needs to be approximated via an \themebf{evidence lower bound (ELBO)} - which is, as the name suggests, only a lower bound to the loss we actually want to minimize. Later, \citet{song2020score} showed that these constructions were essentially an approximation of a time-continuous SDEs. Further, the ELBO loss becomes tight (i.e. it is not a lower bound anymore) in the continuous time case (e.g. note that \cref{thm:fm_loss} and \cref{thm:dsm_loss} are equalities and not lower bounds - this would be different in the discrete time case). This made the SDE construction popular because it was considered mathematically "cleaner" and that one could control the simulation error via ODE/SDE samplers post training. It is important to note however that both models employ the same loss and are \textit{not} fundamentally different.

\paragraph{"Forward process" vs probability paths.} The first wave of denoising diffusion models \citep{sohl2015deep, song2019generative, ho2020denoising, song2020score} did not use the term \themebf{probability path} but constructed a noising procedure of a data point $z\in\mathbb{R}^d$ via a so-called \themebf{forward process}. This is an SDE of the form
\begin{align}
\label{eq:forward_process}
    \bar{X}_0=z,\quad \dd \bar{X}_t = \uforw_t(\bar{X}_t)\dd t + \sigforw_t \dd \bar{W}_t
\end{align}
The idea is that after drawing a data point $z\sim \pdata$ one simulates the forward process and thereby corrupts or "noises" the data. The forward process is designed such that for $t\to \infty$ its distribution converges to a Gaussian $\mathcal{N}(0,I_d)$. In other words, for $T\gg 0$ it holds that $\bar{X}_{T}\sim\mathcal{N}(0,I_d)$ approximately. Note that this essentially corresponds to a probability path: the conditional distribution of $\bar{X}_t$ given $\bar{X}_0=z$ is a conditional probability path $\bar{p}_t(\cdot|z)$ and the distribution of $\bar{X}_t$ marginalized over $z\sim \pdata$ corresponds to a marginal probability path $\bar{p}_t$.\footnote{Note however that they use an \themebf{inverted time convention}: $\bar{p}_0(\cdot|z)=\pdata$ here.} However, note that with this construction, we need to know the distribution of $X_t|X_0=z$ in closed form in order to train our models to avoid simulating the SDE. This essentially restrict the vector field $\uforw_t$ to ones such that we know the distribution $\bar{X}_t|\bar{X}_0=z$ in closed form. Therefore, throughout the diffusion model literature, vector fields in forward processes are always of the affine form, i.e. $\uforw_t(x)=a_t x$ for some continuous function $a_t$. For this choice, we can use known formulas of the conditional distribution \citep{sarkka2019applied,song2021sde, karras2022elucidating}:
\begin{align*}
\bar{X}_t|\bar{X}_0=z\sim\mathcal{N}\left(\alpha_t z,\beta_t^2 I\right),\quad\alpha_t=\exp\parr{\int\limits_{0}^{t}a_r\dd r},\quad\beta_t^2=\alpha_t^2\int\limits_{0}^{t}\frac{(\sigforw_r)^2}{\alpha^2_r}dr
\end{align*}
Note that these are simply Gaussian probability paths. Therefore, one can say that \textbf{a forward process is a specific way of constructing a (Gaussian) probability path.} The term probability path was introduced by flow matching \citep{lipman2022flow} to both simplify the construction and make it more general at the same time: First, the "forward process" of diffusion models is never actually simulated (only samples from $\bar{p}_t(\cdot|z)$ are drawn during training). Second, a forward process only converges for $t\to\infty$ (i.e. we will never arrive at $\pinit$ in finite time). Therefore, we choose to use probability paths in this document.

\paragraph{Time-Reversals vs Solving the Fokker-Planck equation.}

The original description of diffusion models did not construct the training target $\uref_t$ or $\nabla\log p_t$ via the Fokker-Planck equation (or Continuity equation) but via a \themebf{time-reversal} of the forward process \citep{anderson1982reverse}. A time-reversal $(X_t)_{0\leq t\leq T}$ is an SDE with the same distribution over trajectories inverted in time, i.e.
\begin{align}
\mathbb{P}[\bar{X}_{t_1}\in A_1,\dots,\bar{X}_{t_n}\in A_n]=\mathbb{P}[X_{T-t_1}\in A_1,\dots,X_{T-t_n}\in A_n]\\
    \text{ for all }0\leq t_1,\dots, t_n\leq T, \text{ and } A_1,\dots,A_n\subset S
\end{align}
As shown in \citet{anderson1982reverse}, one can obtain a time-reversal satisfying the above condition by the SDE:
\begin{align*}
    \dd X_t =& \left[-u_t(X_t)+\sigma_t^2\nabla\log p_t(X_t)\right]\dd t+ \sigma_{t}\dd W_t,\quad u_t(x)=\uforw_{T-t}(x),\sigma_t=\bar{\sigma}_{T-t}
\end{align*}
As $u_t(X_t)=a_tX_t$, the above corresponds to a specific instance of training target we derived in \cref{prop:conversion_formula_gaussian_prob_path} (this is not immediately trivial as different time conventions are used. See e.g. \citep{lipman2024flow} for a derivation). However, for the purposes of generative modeling, we often only use the final point $X_1$ of the Markov process (e.g., as a generated image) and discard earlier time points. Therefore, whether a Markov process is a ``true'' time-reversal or follows along a probability path does not matter for many applications. Therefore, using a time-reversal is not necessary and often leads to suboptimal results, e.g. the probability flow ODE is often better \citep{karras2022elucidating, ma2024sit}. All ways of sampling from a diffusion models that are different from the time-reversal rely again on using the Fokker-Planck equation. We hope that this illustrates why nowadays many people construct the training targets directly via the Fokker-Planck equation - as pioneered by \citep{lipman2022flow,liu2022flow,albergo2023stochastic} and done in this class.

\paragraph{Flow Matching \citep{lipman2022flow} and Stochastic Interpolants \citep{albergo2023stochastic}.}  The framework that we present is most closely related to the frameworks of flow matching and \themebf{stochastic interpolants (SIs)}. As we learnt, flow matching restricts itself to flows. In fact, one of the key innovations of flow matching was to show that one does not need a construction via a forward process and SDEs but flow models alone can be trained in a scalable manner. Due to this restriction, you should keep in mind that sampling from a flow matching model will be  deterministic (only the initial $X_0\sim \pinit$ will be random). Stochastic interpolants included both the pure flow and the SDE extension via "Langevin dynamics" that we use here (see \cref{thm:langevin_trick}). Stochastic interpolants get their name from a \themebf{interpolant function} $I(t,x,z)$ intended to interpolate between two distributions. In the terminology we use here, this corresponds to a different yet (mainly) equivalent way of constructing a conditional and marginal probability path. The advantage of flow matching and stochastic interpolants over diffusion models is both their simplicity and their generality: their training framework is very simple but at the same time they allow you to go from an arbitrary distribution $\pinit$ to an arbitrary distribution $\pdata$ - while denoising diffusion models only work for Gaussian initial distributions and  Gaussian probability path. This opens up new possibilities for generative modeling that we will touch upon briefly later in this class.

\begin{summarybox}[Alternative Diffusion Formulations]
Alternative formulations for diffusion models that are popular in the literature often involve some combination of the following elements:
\begin{enumerate}
    \item \textbf{Discrete-time: }Approximations of SDEs via discrete-time Markov chains are often used.
    \item \textbf{Inverted time convention: }It is popular to use an inverted time convention where $t=0$ corresponds to $\pdata$ (as opposed to here where $t=0$ corresponds to $\pinit$).
    \item \textbf{Forward process: }Forward processes (or noising processes) are ways of constructing (Gaussian) probability paths.
    \item \textbf{Training target via time-reversal:} A training target can also be  constructed via the time-reversal of SDEs. This is a specific instance of the construction presented here (with an inverted time convention).
\end{enumerate}

\end{summarybox}
% \subfile{subfiles/appendix_neural_odes.tex}
\end{document}